\documentclass[12pt,a4paper]{report}

\usepackage{titlesec}
\titleformat{\chapter}[display]   
{\normalfont\huge\bfseries}{\chaptertitlename\ \thechapter}{20pt}{\Huge}   
\titlespacing*{\chapter}{0pt}{-50pt}{40pt}

\usepackage[mt,fs,english]{packages/ethasl}

\usepackage[utf8]{inputenc}
\usepackage[OT1]{fontenc}

\usepackage{a4}

\usepackage{fancyhdr}

\usepackage{textcomp}\usepackage{gensymb}

\usepackage[english]{isodate}

\usepackage{multicol}

\usepackage{graphicx}

\usepackage{subcaption}

\usepackage[numbers]{natbib}
\setcitestyle{authoryear, open=[,close=]}

\usepackage{booktabs}
\usepackage{array}
\usepackage{multirow}

\usepackage{color}
\usepackage{colortbl}
\definecolor{black}{rgb}{0,0,0}
\definecolor{white}{rgb}{1,1,1}
\definecolor{darkred}{rgb}{0.5,0,0}
\definecolor{darkgreen}{rgb}{0,0.5,0}
\definecolor{darkblue}{rgb}{0,0,0.5}

\usepackage{amsmath}
\usepackage{amssymb}
\usepackage{amsthm}

\usepackage{nicefrac}

\usepackage{upgreek}

\usepackage{isomath}

\usepackage{units}

\usepackage{rotating}

\usepackage{parskip}

\usepackage{pdfpages}
\includepdfset{pages={-}, frame=true, pagecommand={\thispagestyle{fancy}}}

\PassOptionsToPackage{hyphens}{url}\usepackage{hyperref}

\usepackage[longend]{algorithm2e}
\usepackage{textgreek}
\usepackage{tcolorbox}

\usepackage{cleveref}

\theoremstyle{definition}
\newtheorem{definition}{Definition}[section]
\crefname{definition}{Definition}{Definitions}

\usepackage{multirow}
\usepackage{adjustbox}

\begin{document}

\title{Teacher-student curriculum learning for reinforcement learning }


\supervisionA{Prof. Dr. Manfred Vogel}

\projectYear{\the\year} 
\subtitle{Yanick Schraner}

\maketitle
\pagestyle{plain}
\pagenumbering{roman}

\pagestyle{empty}


\section*{Declaration of Originality}

\vspace{1cm}

I hereby declare that the written work I have submitted entitled

\vspace{0.5cm}

\textbf{Teacher-student curriculum learning for reinforcement learning}

\vspace{0.5cm}

is original work which I alone have authored and which is written in my own words.

\vspace{1cm}

\textbf{Author}

\vspace{.5cm}

\begin{tabular}{ p{1.5cm} p{1.5cm} }
  Yanick & Schraner \\
\end{tabular}

\vspace{.5cm}

\textbf{Supervisor}

\vspace{.5cm}

\begin{tabular}{ p{1.5cm} p{1.5cm} }
  Manfred & Vogel \\
\end{tabular}

\vspace{1cm}

With the signature I declare that I have been informed regarding normal academic citation rules.
The citation conventions usual to the discipline in question here have been respected.

\vspace{4cm}

\begin{tabular}{ p{5cm} p{1cm} p{5cm} }
  \cline{1-1} \cline{3-3}
  Place and date & & Signature \\
\end{tabular}


\setcounter{tocdepth}{2}

\tableofcontents
\clearpage

\chapter*{Abstract}
\addcontentsline{toc}{chapter}{Abstract}

Reinforcement learning (rl) is a popular paradigm for sequential decision making problems.
The past decade's advances in rl have led to breakthroughs in many challenging domains such as video games, board games, robotics, and chip design.
The sample inefficiency of deep reinforcement learning methods is a significant obstacle when applying rl to real-world problems.
Transfer learning has been applied to reinforcement learning such that the knowledge gained in one task can be applied when training in a new task.
Curriculum learning is concerned with sequencing tasks or data samples such that knowledge can be transferred between those tasks to learn a target task that would otherwise be too difficult to solve.
Designing a curriculum that improves sample efficiency is a complex problem.
In this thesis, we propose a teacher-student curriculum learning setting where we simultaneously train a teacher that selects tasks for the student while the student learns how to solve the selected task.
Our method is independent of human domain knowledge and manual curriculum design.
We evaluated our methods on two reinforcement learning benchmarks: grid world and the challenging Google Football environment.
With our method, we can improve the sample efficiency and generality of the student compared to tabula-rasa reinforcement learning.
\clearpage
\chapter*{Acknowledgements}
\addcontentsline{toc}{chapter}{Acknowledgements}

I want to thank everyone who supported me throughout this thesis.
I am thankful for the moral support, open door for my thoughts and concerns, and advice during my thesis work.

Firstly, I would like to extend my special thanks to my advisor Prof. Dr. Manfred Vogel, who supported me in my thesis and throughout my master's studies.
I am thankful for his criticism, humor, and his guidance during this time.

I would also like to thank Christian Scheller and Lukas Neukom for their proofreading and valuable comments on this thesis.
\clearpage
\chapter*{Symbols}
\label{sec:symbols}
\addcontentsline{toc}{chapter}{Symbols}

\section*{Symbols}
The following list describes important notations used in this work.

\begin{tabbing}
 \hspace*{2cm} \= \kill
  $\mathbb{E}[X]$   \> expectation of a random variable X \\[0.5ex]
  $\alpha, \beta$       \> step-size parameters \\[0.5ex] 
  $\gamma$   \> discount-rate parameter \\[0.5ex]
  $s, s' \in \mathcal{S}$   \> states, $s'$ is subsequent step of $s$ \\[0.5ex]
  $a \in \mathcal{A}$       \> an action \\[0.5ex]
  $r \in \mathcal{R}$       \> a reward \\[0.5ex]
  $\mathcal{S}$       \> set of all states \\[0.5ex]
  $\mathcal{A}(s)$       \> set of all actions available in state $s$ \\[0.5ex] 
  $\mathcal{R}$     \> set of all possible rewards, a finite subset of $\mathbb{R}$ \\[0.5ex]
  $t$       \> discrete time step \\[0.5ex]
  $T$       \> final time step of an episode\\[0.5ex]
  $a_t$       \> action at time $t$ \\[0.5ex]
  $s_t$       \> state at time $t$ \\[0.5ex]
  $r_t$       \> reward at time $t$ \\[0.5ex]
  $\pi$       \> policy \\[0.5ex]
  $\pi(s)$      \> action taken in state $s$ under \textit{deterministic} policy $\pi$ \\[0.5ex]
  $\pi(a|s)$       \> probability of taking action $a$ in state $s$ under \textit{stochastic} policy $\pi$ \\[0.5ex]
  $p(s'|s,a)$       \> probability of transition to state $s'$, from state $s$ taking action $a$ \\[0.5ex]
  $r(s,a)$       \> expected immediate reward from state $s$ after action $a$ \\[0.5ex]
  $V^\pi(s)$       \> state-value function for policy $\pi$\\[0.5ex]
  $V^*(s)$       \> optimal state-value function \\[0.5ex]
  $Q^\pi(s, a)$       \> action-value function under policy $\pi$  \\[0.5ex]
  $Q^*(s, a)$       \> optimal action-value function \\[0.5ex]
  $\boldsymbol{\theta},\boldsymbol{\theta_t}$       \> parameter vector of target policy \\[0.5ex]
  $\pi(a|s,\boldsymbol{\theta})$       \> probability of taking action $a$ in state $s$ given parameter vector $\boldsymbol{\theta}$ \\[0.5ex]
  $\pi_{\boldsymbol{\theta}}$       \> policy corresponding to parameter $\boldsymbol{\theta}$ \\[0.5ex]
  $\nabla\pi(a|s,\boldsymbol{\theta})$       \> column vector of partial derivatives of $\pi(a|s,\boldsymbol{\theta})$ with respect to $\boldsymbol{\theta}$ \\[0.5ex]
  $J(\boldsymbol{\theta})$       \> performance measure for the policy $\pi_{\boldsymbol{\theta}}$ \\[0.5ex]
  $\nabla J(\boldsymbol{\theta})$       \> column vector of partial derivatives of $J(\boldsymbol{\theta})$ with respect to $\boldsymbol{\theta}$ \\[0.5ex]
  $\mathcal{V}$ \> Set of all task used in curriculum learning \\[0.5ex]
\end{tabbing}


\section*{Acronyms and Abbreviations}
\begin{tabbing}
 \hspace*{1.6cm}  \= \kill
 ALP \> Absolute learning progress \cref{sec:teacherObservation} \\[0.5ex]
 EMA \> Exponential moving average \cref{sec:teacherObservation} \\[0.5ex]
 FS-EMA \> Fast and slow exponential moving average \cref{sec:teacherObservation} \\[0.5ex]
 LP \> Learning progress \cref{sec:teacherObservation} \\[0.5ex]
 PTR \> Previous task reward \cref{sec:teacherObservation} \\[0.5ex]
 RH \> Reward history \cref{sec:teacherObservation} \\[0.5ex]
 RL \> Reinforcement Learning \cref{sec:rl} \\[0.5ex]
 SMM \> Super Mini Map \cref{sec:GFenvironment} \\[0.5ex]
 MDP \> Markov decision problem \cref{sec:mdp} \\[0.5ex]
 POMDP \> Partially observable Markov decision problem \\[0.5ex]
 PPO \> Proximity Policy Optimization \cref{sec:ppo} \\[0.5ex]

\end{tabbing}
\clearpage

\pagestyle{fancy}
\pagenumbering{arabic}

\chapter{Introduction}
\label{sec:introduction}

Reinforcement learning is based on the idea of learning through interaction with an environment.
Learning through interaction is a natural way to gain new skills and a concept underlying nearly all theories of learning and intelligence.
Infants learn a lot about the cause and effect of their actions by observing the results of their actions.
We learn how to achieve specific goals solely by observing the joy of actions leading to a pleasant state in our environment (like grabbing our favorite toys) or getting harmed by other actions (like touching a hot plate).

Reinforcement learning (RL) builds upon this idea of learning through interaction.
As supervised learning (SL) and unsupervised learning (UL), reinforcement learning is a machine learning paradigm.
Due to its generality, researchers apply RL to a broad range of tasks.
The ongoing rise of deep learning enabled reinforcement learning on high-dimensional input and highly complex environments.
Successful applications are board games \citep{Silver2016,Silver2017,Silver1140}, video games \citep{Mnih2015, openai2019dota, Vinyals2019} and robotics \citep{openai2019solving}.

The use of reinforcement learning in a practical setting is often not realistic due to the sample inefficiency of RL.
RL agents usually require a large number of interactions with the environment to learn proper behavior.
For example, the training of OpenAI Five \citep{openai2019dota} took two months with 150 PFlops / day to surpass professional human-level performance.
Training Agent57 \citep{badia2020agent57} for the Atari benchmark required 90 billion environment frames to reach human performance.

There are multiple ways to improve the sample efficiency of reinforcement learning.
Imitation learning methods make use of an existing expert demonstration dataset to learn from expert priors.
The simplest form of imitation learning is behavior cloning, where a policy is learned in a supervised learning fashion \citep{10.5555/89851.89891}.
This method suffers from the distribution mismatch of the training data and the actual environment.
A slight divergence of trajectories present in the dataset during inference leads to situations unknown to the trained policy.
\citet{pmlr-v15-ross11a} use a human in the loop process to counter the distribution mismatch by continuously labeling the data for those unknown situations.
Another option is to use reinforcement learning to fine-tune the initial policy \citep{pmlr-v123-scheller20a, Vinyals2019}.
A different popular strategy to make RL more sample efficient is to exploit the hierarchical structure of the given problem \citep{10.5555/645527.657453, NIPS1992_714, 10.1016/S0004-3702(99)00052-1}.
One can imagine training a sub-policy for every individual task and a general policy that chooses one of those sub-policies.
Another way to improve sample efficiency is curriculum learning (CL), where the learning process is structured so that new concepts are learned in a sequence and an agent can leverage what he previously learned.
Let us consider maths as an analogy: We first have to learn basic arithmetic before we can do linear algebra.
The idea of using such a structured learning process to train artificial agents dates back to \citet{Elman1993-ELMLAD} for grammar learning.

In this work, we build upon the idea of curriculum learning to improve the sample efficiency of our RL approach.
There are many ways to define a curriculum.
We will specify those in \cref{sec:curriculumBackground}.
Our work focuses on how we can automatically define such a curriculum during training by phrasing the task sequencing problem as a meta Markov decision process.
We present a novel curriculum learning framework where no domain knowledge for the task sequencing is required.
This framework aims to increase the sample efficiency as well as the asymptotic reward compared to vanilla reinforcement learning and a set of heuristic-based baselines.

This work builds upon a previous project where we applied manually designed curriculum learning on a football simulation.
In this previous work, we were able to increase the sample efficiency of our algorithm and the asymptotic reward to 1 compared to vanilla reinforcement learning with a reward of -1.4 on the 11 vs. 11 hard environment. 
In \cref{sec:prevWork} we provide a brief recapitulation of the previous work and its findings.

\section{Related Work}
\label{sec:relatedWork}
Our work lies in the field of curriculum learning in reinforcement learning.
CL comprises three key elements: transfer learning, task sequencing, and task generation.
In this work, we focus on transfer learning and task sequencing.

\paragraph{Transfer learning}
is a field in machine learning which studies how knowledge gained by solving a source task can be transferred to a target problem such that the target task can be solved faster.
In reinforcement learning, transfer learning increases the sample efficiency or the performance on the target task.
Knowledge can be transferred by selecting samples collected on a source task and use them as input for batch reinforcement learning \citep{10.1145/1390156.1390225, NIPS2011_fe7ee8fc}.
Options \citep{Sutton99betweenmdps} or macro-actions can be extracted on a source task and included in the action space of the target task \citep{DBLP:conf/aaai/SoniS06, 10.5555/3157382.3157488}.
Models of the transition and reward function of a source task can be transferred to allow a hybrid approach of model-free and model-based RL on the target task \citep{FACHANTIDIS201323}.
The parameters of a learned policy or value function can also be used to initialize the policy or value function in the target task \citep{FERNANDEZ2010866, JMLR07-taylor, AAMAS05-transfer}.

Those methods make different assumptions about the source and target Markov Decision Processes (MDP).
The action or state space must be shared to allow the initialization of a policy or value function.
Alternatively, a task mapping for states and actions of the source task to their equivalents of the target task is needed in tabular reinforcement learning.
For some methods, we need access to a model of the source task, and not all methods allow the use of multiple source tasks.

In our work, we focus on transfer learning through policy transfer and reward shaping.

We refer to the surveys on transfer learning for reinforcement learning provided by \citet{JMLR:v10:taylor09a, Lazaric2012} for more insights.

\paragraph{Task sequencing} is concerned with how the tasks can be sequenced in order to provide a curriculum.
The sequencing of those tasks is crucial in CL.
If the tasks at the beginning of the sequence are too hard, then the agent may fail to learn and is unable to solve the following tasks.
The goal is to sequence the task into a curriculum which allows the agent to learn with a higher sample efficiency and final performance compared to an agent trained solely on the target ask.
A suitable task sequence depends on the set of tasks, the agent's characteristics, the target task, and is domain-specific.

To perform task sequencing, we must control the environment to a certain degree to create different tasks.
Depending on the level of control we have, we can apply different task sequencing methods.
In practice, the first approach to task sequencing is made manually.
We focus on how we can automatically generate a task sequence for our curriculum.
The simplest form is a single task curriculum where we reorder recorded experience without changing the MDP, referred to as sample sequencing.
We collect this experience into an experience buffer during agent-environment interactions.
This idea has its roots in supervised learning, where the training data is sampled in a specific order to speed up the training process \citep{bengio:2009}.
One example of sample sequencing in RL is Prioritized Experience Replay for DQN and its follow-up work \citep{Schaul2016, NIPS2017_453fadbd, 8278851, DBLP:journals/corr/abs-1801-00904}.

In multi-agent environments, we can create a curriculum by controlling the interaction of agents in the same environment.
This approach to task sequencing is called co-learning.
In its simplest form, one performs self-play where an agent competes against or acts cooperatively with an older version of itself.
The agent to train and his counterpart gets progressively better on the task and therefore creates an implicit curriculum.
This setup has proven to be successful in the well-known Go AI Alpha Go and its successors \citep{Silver2016, Silver1140}.
In \citet{Vinyals2019} the idea of self-play was extended to a league setting.
The goal of this league setup is that the agent faces increasingly stronger opponents and opponents trained to perform specific strategies to exploit weaknesses and prevent mode collapse.

In co-learning and sample sequencing, the target MDP is not changed, and no specific level of control over the environment is required.
If control over the environment is possible, one can change the target MDP by changing the initial state distribution or the reward function to create a suitable task sequence.
One example of this approach is the reverse curriculum generation, where a robot is learning to reach a goal from a set of starting positions increasingly far from the goal \citep{pmlr-v78-florensa17a}.

\citet{8850690} uses metaheuristic search methods such as beam search, tabu search, genetic algorithms, or ant colony search in order to solve the task sequencing problem.
In their follow-up paper, they compare those metaheuristic search methods to their Heuristic Task Sequencing for Cumulative Return (HTS-CR) algorithm \citep{ijcai2019-320}.

In our work, we treat the task sequencing as an MDP \citep{ijcai2017-353, 8827566, AAMAS19-Narvekar} where we use reinforcement learning in order to train a teacher agent to perform the task sequencing online.
In this setting, we formalize curriculum learning as an interaction of two MDPs.
The standard MDP modeling the learning agent interacting with the environment is referred to as student MDP.
A meta level MDP for the curriculum agent to perform the task sequencing referred to teacher MDP.

At the time of writing, no MDP-based task sequencing method where both student and teacher use neural network-based function approximators and where both are trained using reinforcement learning is known to the author.
Therefore the key focus is to propose such a framework and analyze the efficiency of such an approach.

\paragraph{Task generation} specifies how and when the source tasks for the curriculum are defined.
The quality of its source tasks heavily influences the quality of a curriculum.
The tasks can either be created beforehand or online during training.
The goal of task generation is to create a set of source tasks that allow a knowledge transfer through them such that it is easier to solve the target task.

In \citet{AAMAS16-Narvekar}, a method for creating a set of source tasks by specifying task descriptors, that are controlling the degrees of freedom of the task, is introduced.
Those task descriptors specify the environment like the environment size, action set, opponent type, initial states, et cetera.
In Powerplay \citep{10.3389/fpsyg.2013.00313} a framework where new tasks are generated online is introduced.
The system searches for new source tasks such that the agent becomes more and more general.
A fixed computation budget is applied to force the task generator to create new tasks that are only slightly more difficult than the previous ones.
\citet{jiang2020prioritized} introduced the idea of prioritized level replay.
In reinforcement learning environments, there is usually a level identifier.
This could be a level index or a random seed.
Usually, the level to use is sampled uniformly.
Which level is sampled can influence the difficulty of the task as well as the environment dynamics.
This diversity among levels implies that different levels hold different learning potentials for RL agents at any point in training.
Prioritized level replay introduces a new level sampling strategy to prioritize levels based on their learning potential creating an implicit curriculum.

\section{Previous Work}
\label{sec:prevWork}
In previous work, we tried to reproduce the work of \citet{kurach2019google} on the Google Research Football environment.
We also implemented and evaluated extensions to the Proximal Policy Optimization (PPO) algorithm such as Actor with Variance Estimated Critic (AVEC) \citep{fletberliac2020standard}.
Additionally, we carried out experiments with curriculum learning in the Google football environment.
Detailed information about this environment can be found in \cref{sec:GFenvironment}.
We evaluated two manually defined curricula as well as two automatically generated curricula, and a prioritized level replay curriculum.
We used the weights obtained by training on the source task as an initialization for the policy on the target task as a transfer method.

The results of our previous work are summarized in \cref{tab:prevwork_results} for a comparison in \cref{sec:resultsFootball}.
\textit{Scenarios curriculum} and \textit{11 vs 11 curriculum} are both manually defined curricula, the first one over both a set of football academy tasks (see \cref{tab:footballAcademy}) and the 11 vs 11 full game on difficulty easy, medium and hard.
The second curriculum only uses the 11 vs 11 full game environments.
The \textit{increasing curriculum} consistently increases the game difficulty parameter $\delta$ by $0.05$ starting from $0.01$ and finishing at $0.95$.
A $\delta$ value of $0.95$ is equivalent to the 11 vs 11 hard environment.
At the first task change, we increase $\delta$ from $0.01$ to $0.05$ instead.
The difficulty is increased automatically after 100 training iterations.
\textit{Smooth increasing curriculum} is similar to the \textit{increasing curriculum}.
Instead of adapting the difficulty depending on training iterations, we increase the difficulty as soon as the average return is greater than $0.9$.
Additionally, we increment $\delta$ by $0.001$, start with a value of $0.001$ and end with a $\delta$ value of $0.95$.

Although four out of five curricula improve our results, only the \textit{smooth increasing curriculum} can outperform the baseline significantly.
Using only the 11 vs. 11 tasks includes too hard tasks at the early stages of learning.
The \textit{smooth increasing curriculum} is superior over the \textit{increasing curriculum} as the task difficulty increases more evenly.
\textit{Scenarios curriculum} includes task switches with significant changes in their initial state distribution and observation space, making this type of curriculum worse than the increasing ones.

\begin{table}[htb]
    \centering
    \begin{tabular}{cc}
    \hline
        \textbf{Experiment} & \textbf{Return} \\
        \hline
        PPO & -1.4 \\
        Best \textit{scenarios curriculum} & -0.85 \\
        Best \textit{11 vs 11 curriculum} & -2.08 \\
        Best \textit{increasing curriculum} & -0.48 \\
        Best \textit{smooth increasing curriculum} & 1 \\
        Prioritized level replay & -1.05 \\
        \hline
        Baseline & -1.39
    \end{tabular}
    \caption{Comparison of our different curriculum learning approaches, the average return over 100 episodes on the 11 vs 11 hard environment is reported.}
    \label{tab:prevwork_results}
\end{table}

\section{Problem Statement}
\label{sec:problemStatement}
In this work, we are developing a teacher-student curriculum learning setup focusing on online task sequencing and transfer learning.
We manually do the task generation and focus on the task sequencing problem formulated as a curriculum MDP \citep{ijcai2017-353}.
Both the teacher and student are policy gradient reinforcement learning agents using neural networks as function approximators and are trained with PPO.
We aim to answer the following research questions:

\begin{itemize}
    \itemsep0em
    \item Is it possible to train both teacher and student with the curriculum MDP (CMDP) setting from scratch such that the sample efficiency or average reward on the target task increases compared to vanilla PPO on the target task?
    \item What is a suitable reward function for our teacher agent to increase the sample efficiency or average reward on the target task?
    \item How can we define observations in the CMDP such that the sample efficiency or average reward on the target task increases?
    \item What is a suitable transfer method for the specified setting?
    \item Are we able to improve the results of our previous work on the Google Research Football environment with this curriculum learning framework?
\end{itemize}

\clearpage
\chapter{Background}
\label{sec:background}

In this chapter, we introduce the core ideas and theoretical foundations used in this work.
We introduce the reinforcement learning paradigm focusing on policy gradient methods and highlight some core ideas important in this work.
We introduce the proximal policy optimization (PPO) algorithm and additional RL improvements used in this work.
Parts of this chapter were written in our previous work \citep{schraner_2020} and added to this thesis for a coherent document. 

\section{Reinforcement Learning}
\label{sec:rl}
Reinforcement learning is a popular framework suited to solve sequential decision-making processes.
An agent learns how to act in an environment by observing a numerical reward signal.
The agent has to learn a policy to predict an action based on the environment's state, such that the cumulative reward is maximized.
At the early stages of learning, RL is very similar to trial and error learning.
By making progress, the learner usually continuously observes new states of the environment and learns how to act in those new states without forgetting about the correct behavior in the early stages.
The only indication wether a single or a series of actions leads to preferable changes in the environment is the reward signal.
After every action, the agent receives a representation of the new environment state (observation) and a reward signal.
This signal can be either positive or negative, and it does not tell the agent exactly which series of actions lead to that signal.
The reward can be delayed, for example, students receive a negative reward signal while studying for their exams, but they receive a large positive reward signal after successfully passing the exam.
This way of learning makes RL a general framework, as stated by the reward hypothesis:
\begin{quote}
    That all of what we mean by goals and purposes can be well thought of as the maximization of the expected value of the cumulative sum of a received scalar signal (called reward) \citep{Sutton1998}.
\end{quote}
Biological systems inspire the use of a reward signal to learn: through experiencing pleasure or pain, we know what actions are good in an immediate sense.

In contrast to RL, supervised learning uses a training set of labeled examples. 
With supervised learning, we are trying to capture the knowledge represented in the training set.
The use of a training set proved to be very powerful to learn intelligent behavior, but it also makes supervised learning hard for interactive problems.
Often it is impractical or costly to obtain a useful set of examples of the desired behavior that is both correct and representative for all situations.
There is an intersection of RL and supervised learning called imitation learning.
In imitation learning, the agent receives an initial set of labeled demonstrations of actions in the environment and uses it in a supervised fashion.
The agent may be improved further with standard reinforcement learning.

With unsupervised learning, one can find structure hidden in a large amount of unlabeled data.
There is no reward signal to maximize in unsupervised learning, which differentiates it from RL.

\subsection{Finite Markov Decision Processes}
Markov decision processes (MDP) is a formalisation for sequential decision-making processes.
A Markov decision process is a 4-tuple $(\mathcal{S},\mathcal{A},p,r)$, where
\begin{itemize}
    \item $\mathcal{S}$ is a set of states
    \item $\mathcal{A}$ is a set of actions
    \item $p(s'|s,a) = P[s_{t+1} = s' | s_t=s, a_t = a]$ is the probability of transitioning from state $s$ to state $s'$ when taking action $a$.
    \item $r(s,a) = \mathbb{E}[r_{t+1} | s_t = s, a_t = a]$ is the expected reward received by taking action $a$ in state $s$.
\end{itemize}

MDPs are an idealized form of the reinforcement learning problem and are used to formulate mathematically precise theoretical statements \citep{Sutton1998}.

\begin{figure}[htb]
    \centering
    \includegraphics[width=8cm]{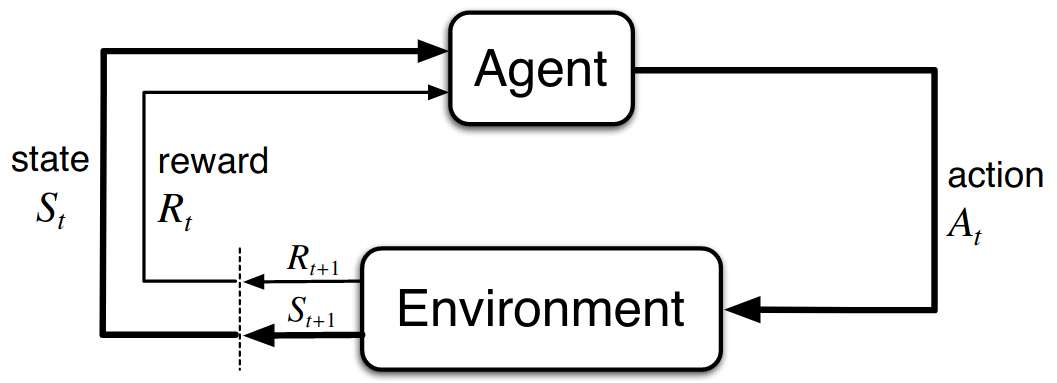}
    \caption{The agent-environment interaction in a Markov decision process \citep{Sutton1998}.}
    \label{fig:RLAgentEnvironmentMDP}
\end{figure}

The agent-environment interaction in a MDP is shown in \cref{fig:RLAgentEnvironmentMDP}.
The decision-maker is called an agent.
The agent interacts with the environment by following a policy denoted by $\pi$.
The policy maps from state $s$ to action $a \in \mathcal{A}$ and can either be deterministic $\pi(s) = a$ or stochastic $\pi(a|s) = P_\pi[a_t=a|s_t=s]$.
Everything outside the agent is the environment.
Taking an action in the environment leads to a change in the environment's state.
For example in the grid world environment, an action can be moving forward, picking up a key, or opening a door.
The environment receives the action of the agent and returns a reward $r$ and a representation of its new state $s \in \mathcal{S}$.
If the entire state of the environment is not observable, e.g., the agent has only a limited field of view, the agent receives an observation $o \in \mathcal{O}$ instead of a state.
The observation only partially describes the environment's state.
The environment can be deterministic or stochastic and may change itself without interactions from the agent.
Agent and environment are interacting in a sequence of discrete-time steps $t$.
$t$ starts from $0$ in the first time-step, and goes up to $T$, with $T$ being the last time-step.
At each time step, the agent receives a representation of the environment state $s_t$ and a reward $r_t$ based on which he selects and executes action $a_t$.
This leads to a sequence of interactions called trajectory: $s_0 a_0 r_1 s_1 a_1 r_2 s_2 a_2 r_3 \dots s_t a_t r_t$.
A trajectory has to be finite.

In a finite MDP, the sets of states, actions and rewards all have a finite number of elements.
The next state is only dependent on the preceding state-action pair.
Therefore we have a well defined discrete probability distributions for the next state dependent only on the preceding state-action pair \citep{Sutton1998}:
\begin{equation}
    \text{P}[s_{t+1} | s_t , a_t] = P[s_{t+1} | s_1, a_1, \dots, s_t, a_t]
\end{equation}
This is called the Markov property.
To fulfill the Markov property, the probabilities given by $p$ can only depend on the preceding state and action and completely characterize the dynamics of the environment.
Therefore, a single state must include information about every aspect of the past agent-environment interactions that make a difference for the future.

\subsection{Policy and Value Functions}
In reinforcement learning, agents select their actions according to their policy function.
A policy has to be time-independent, the trajectory up to the time step t does not influence the action probabilities.
The policy $\pi(a|s)$ outputs a probability distribution over all possible actions given a state:
\begin{equation}
    \pi(a|s) = P[a_t=a | s_t = s]
\end{equation}

The objective of this policy is to maximize the cumulative future reward, also called return $G_t$:
\begin{equation}
    G_t=r_{t+1}+\gamma r_{t+2} + \gamma^2 r_{t+3} + \cdots = \sum_{k=t}^{T-1}\gamma^k r_{t+k+1}
\end{equation}

\clearpage

The discount factor $\gamma \in [0,1]$ serves multiple purposes:
\begin{itemize}
    \item future rewards may have higher uncertainty
    \item future rewards do not provide immediate benefits.
    In some cases, immediate rewards are of more value, like in economics where we prefer the money now over later as we could invest it in maximizing future earnings.
    \item the discount factor provides mathematical convenience, as it solves problems with infinite MDPs or loops in the state transition graph \citep{Sutton1998}
\end{itemize}

The performance of a policy is measured by its state- and action-value functions.
Those value functions for the policy $\pi$ are the expectation of what return the agent receives by following policy $\pi$ from state  $s \in \mathcal{S}$.
This estimation is called state-value function $V(s)$ (how good is it to be in a given state) or action-value function $Q(s, a)$ (how good it is to perform a given action in a given state).
The state-value function is the expected return when following policy $\pi$ from state $s$:
\begin{equation}
    V^\pi(s) \doteq \mathbb{E}_\pi[G_t | s_t = s]
\end{equation}
The action-value function is the expected return when taking action $a$ in state $s$ and then following policy $\pi$:
\begin{equation}
    Q^\pi(s,a) \doteq \mathbb{E}_\pi[G_t | s_t = s, a_t = a]
\end{equation}

A policy $\pi$ is considered better than another policy $\pi'$ if for all states $s \in \mathcal{S}$ the expected return is larger:
\begin{equation}
    V^{\pi}(s) > V^{\pi'}(s), \forall s \in \mathcal{S}
\end{equation}

An optimal policy $\pi^*$ is a policy that is better or as good as any other policy in any state:
\begin{equation}
    V^{\pi^*}(s) \geq V^\pi(s), \forall s \in \mathcal{S} \land \forall \pi \in \Pi
\end{equation}
where $\Pi$ is the set of all possible policies.

The optimal state-value function $V^*$ is the maximum expected return over all policies when being in state s:
\begin{equation}
    V^*(s) = \text{max}_\pi V^\pi (s)
\end{equation}

The optimal action-value function $Q^*$ is the maximum expected return over all policies when being in state s and taking action a:
\begin{equation}
    Q^*(s, a) = \text{max}_\pi Q^\pi (s, a)
\end{equation}

Therefore, the optimal policy $\pi^*$ can be obtained by acting greedily according to the action that maximizes $Q^*(s,a)$:

\[
    \pi^*(a_t | s_t) =
\begin{cases}
    1 ,& \text{if } a_t = \text{argmax}_{a \in \mathcal{A}} Q^*(s_t,a)\\
    0 ,& \text{otherwise}
\end{cases}
\]

Standard old fashion approaches to find the optimal state-value function $V^*$ or optimal action-value function $Q^*$ are Dynamic Programming, Monte-Carlo Methods, or Temporal-Difference Learning.
We will not cover those methods in this report and continue with deep reinforcement learning and policy gradient methods.
See \citet{Sutton1998} for an introduction of those traditional methods.

\subsection{Deep Reinforcement Learning}
Tabular methods work well for problems with a small state and action space.
For such environments, it is easy to build a table with value or action-value estimates and act according to this table.
Those problems are not very interesting and far away from real-world applications.
With growing state or action spaces, the memory and computation needs are growing exponentially, with respect to the state or action space, for tabular methods.
It is also not feasible to visit every state to fill a table with value estimates.
With stochastic environments, the problem gets even worse, as we would have to visit every state multiple times to calculate statistics.
This raises the need for state approximation to generalize between similar states.
There are numerous ways to approximate states, but the advances in deep learning made neural networks the preferred option as a function approximator.
The work of \citet{Mnih2015} can be considered as the breakthrough of mixing deep learning with reinforcement learning and allowing a single algorithm to learn how to play Atari games. 
One challenge of deep learning with RL is that the optimization problem is non-stationary because the agent encounters new states in the environment during the ongoing learning progress.

\subsection{Policy Gradient Methods}
Instead of learning value functions and selecting actions based on those functions, we can directly learn a parameterized policy that selects actions without consulting a value function.
This approach is called policy gradient method.
In policy gradient methods the policy $\pi_\theta$ is parametrized with a parameter vector $\theta \in \mathbb{R}^d$.
When using a neural network as a function approximator for the optimal policy, the last layer of this network is usually a softmax layer in case of discrete actions.
The probability distribution will be close to a uniform distribution in the early learning stages, leading to natural exploration.
If the optimal policy is stochastic, then the softmax distribution will approximate the optimal stochastic policy.
If the optimal policy is deterministic, then softmax distribution degenerates to a nearly deterministic policy.
There is no need to specify that beforehand.
Another advantage of policy gradient methods is that we directly optimize for what we care about, which is the optimal policy.
It may be simpler to learn the policy directly than to estimate the state or state-action value.

Given a performance measure for $\pi_\theta$ in the form of a differentiable objective function $J(\theta)$, we can perform gradient \textit{ascent} to update the pararameters $\theta$:
\begin{equation}
    \theta \leftarrow \theta + \alpha \nabla_\theta J(\theta)
\end{equation}
The learning rate is specified by $\alpha$ and needs to be tuned.

The objective of reinforcement learning is to maximize the expected sum of total discounted rewards.
Policy gradient methods take this as an objective function to maximize:
\begin{equation}
    J(\theta) = \mathbb{E}_{\pi_\theta}[\sum_{t=0}^{T-1} \gamma^t r_t]
\end{equation}
The objective function in this form is usually not differentiable as it is dependent on the stationary state distribution of the environment and thus can not be used for gradient ascent.
By the policy gradient theorem \citep{Sutton1998} we can rewrite $J(\theta)$ to a differentiable form:
\begin{equation}
    \nabla J(\theta) = \mathbb{E}_{\pi_\theta}[\nabla_\theta \text{log}\pi_\theta(a_t|s_t) Q_{\pi_\theta}(s_t,a_t)]
\end{equation}

Actor-critic methods are policy gradient methods that are also learning a value function.
This value function is used as a critic.
The critic is used for bootstrapping as in temporal-difference learning \citep{Sutton1998}.
Using one or more estimated values, in this case, an estimate for the action-value, in the update step for the same kind of estimated value is called bootstrapping.
Actor-critic methods use a bootstrapped $n$-step return or directly estimate the action-value function.
The advantage of bootstrapping is variance reduction of the estimates and allowing us to take updates on partial episodes.
However, it comes at the cost of a bias towards the learned critic as it usually does not match the real action-value function.

In advantage actor-critic methods, the same critic used for bootstrapping is also used as a baseline function.
The baseline (estimated action-value) is subtracted from the sampled $n$-step return resulting in a term called advantage.
The advantage tells the agent how much better or worse his policy performs than currently estimated by the critic.
The actor updates the policy parameters $\theta$ for $\pi_\theta(a_t|s_t)$, in the direction suggested by the critic.
If the observed return is less than the expected return, we want to lower the probability of taking the given action; otherwise, we want to assign that action a higher probability.
Using a baseline leads to further variance reduction and thus increases the convergence speed.

\subsection{Proximal Policy Optimization}
\label{sec:ppo}
Proximal policy optimization (PPO) is a policy gradient method for reinforcement learning by \citet{DBLP:journals/corr/SchulmanWDRK17}.
Similar to trust region policy optimization (TRPO) \citep{pmlr-v37-schulman15} a constraint on the policy update is added such that parameter updates do not change the policy too much.
This improves training stability as it prevents large policy updates.
In TRPO an objective function is maximized, while a constraint on the size of policy updates is enforced.
The Kullback–Leibler (KL) divergence of the old policy $\pi_{\theta_\text{old}}$ and the policy after the updated $\pi_\theta$ is used for such a constraint:
\begin{equation}
    \text{maximize}_\theta \mathbb{E} [\dfrac{\pi_\theta (a_t | s_t)}{\pi_{\theta_\text{old}}(a_t | s_t)}A_t(s_t, a_t)]
\end{equation}
\begin{equation}
    \text{subject to } \mathbb{E} [\text{KL}[\pi_{\theta_\text{old}}(\cdot | s_t), \pi_\theta(\cdot | s_t)]] \leq \delta
\end{equation}
where $\theta_\text{old}$ is the vector of policy parameters before the update,  and $A$ is the advantage.

This constraint can be transformed to a penalty in order to solve an unconstrained optimization problem:
\begin{equation}
    \text{maximize}_\theta \mathbb{E} [\dfrac{\pi_\theta(a_t | s_t)}{\pi_{\theta_\text{old}}(a_t | s_t)}A_t - \beta \text{KL}[\pi_{\theta_\text{old}}(\cdot | s_t), \pi_\theta(\cdot | s_t)]]
\end{equation}
where $\beta$ is a scaling coefficient.
In TRPO, the hard constraint is used instead of a penalty because it is hard to choose a value of $\beta$ that performs well. Especially on problems where the characteristics change throughout learning, as is the case in RL.

PPO builds upon this idea of a trust region but implements a more straightforward constraint.
A clipped surrogate objective is used to archive this simplification.
The ratio between old an new policies is denoted by $\phi(\theta)$:
\begin{equation}
    \phi(\theta) = \dfrac{\pi_\theta(a_t | s_t)}{\pi_{\theta_\text{old}}(a_t | s_t)}
\end{equation}

The TRPO objective is, therefore:
\begin{equation}
    J^{TRPO}(\theta) = \mathbb{E}[\phi(\theta) A_t(s_t,a_t)]
\end{equation}

PPO now imposes the constraint by forcing $\phi(\theta)$ to stay within a trust region of $[1-\epsilon, 1 + \epsilon]$ where $\epsilon$ is a hyperparameter.
Simply clipping updates force the policy to stay in the trust region:
\begin{equation}
    J^\text{CLIP}(\theta) = \mathbb{E}[\text{min}(\phi(\theta)A, clip(\phi(\theta), 1 - \epsilon, 1 + \epsilon)A)]
\end{equation}
Therefore if the objective value is not in the trust region, the clipped value will be used.

When PPO is used in a network architecture with shared parameters for policy (actor) and value (critic) functions, an additional entropy term (blue) is introduced to encourage exploration.
Furthermore, the error term on the value estimation (red) is part of the PPO objective function.
\begin{equation}
    J^\text{CLIP}(\theta) = \mathbb{E}[J^\text{CLIP}(\theta) - \textcolor{red}{c_1(V_\theta(s_t) - V_\text{target}(s_t))^2} + \textcolor{blue}{c_2 H(s_t, \pi_\theta(\cdot, s_t))}]
\end{equation}
where both $c_1$ and $c_2$ are scaling parameters for those losses which need to be tuned.

PPO increases the sample efficiency of the RL algorithm empirically, and we hence use it in this work.
\citet{DBLP:journals/corr/abs-2009-10897} has shown that in some cases, PPO fails to converge to a bad local optimum:
\begin{itemize}
    \item Using standard PPO with a continuous action space, training becomes unstable when rewards vanish outside the trust region. This can happen due to a bad Gaussian policy update, where PPO fails to recover.
    \item On high-dimensional discrete action spaces, clipping might converge to suboptimal actions with standard softmax policy parametrization. This happens when the policy sees only bad actions (reward 0) and suboptimal actions without observing the optimal action. PPO then tends to increase the probability of suboptimal actions and not exploring new actions.
    \item PPO can converge to suboptimal actions if they are close to the initialization. 
\end{itemize}

\subsection{Challenges of Deep Reinforcement Learning}
There are a lot of unsolved challenges in deep reinforcement learning.
The most important two are the exploration vs. exploitation dilemma and the deadly triad.

The exploration vs. exploitation dilemma also exists in the real world.
We might have our favorite ski resort, where we go skiing every year.
However, there are a lot of other ski resorts which we have not visited yet.
It could be that one of the new resorts would please us more than our current favorite.
If we do not try out different resorts, we may never find the optimal one, but we risk skiing in a place with only easy slopes and bad restaurants if we try out new ones.
The best long-term strategy might involve short-term sacrifices to find the optimal ski resort.
The difficulty is to find the optimal ratio of exploring new places vs. exploiting the current best one.
This analogy can be adapted to reinforcement learning. We have to explore new actions to learn more about their effectiveness and maximize the expected return in the long run by exploiting new better actions found by exploring.
In stochastic tasks, this dilemma is even worse because we have to explore the same action multiple times due to the stochastic changes.
According to \citet{Sutton1998} this problem remains unresolved and is specific to RL as it does not occur in supervised or unsupervised learning.

The deadly triad is a definition by \citet{Sutton1998}.
It states that instability and even divergence while optimizing arises whenever we combine the following three elements:
\begin{itemize}
    \item \textbf{Function approximation} like neural networks.
    This element is needed to solve problems with large state and action spaces.
    \item \textbf{Bootstrapping} Using one or more estimated values in the update step for the same kind of estimated value \citep{Sutton1998}.
    Bootstrapping adds a bias towards our start estimation, but the updates result in high variance for long trajectories without bootstrapping.
    Without bootstrapping, we need more samples before our estimate converges, leading to a loss in data efficiency and a rise in computational cost.
    \item \textbf{Off-policy training} We call our training off-policy when we update another policy (target policy) than the one we followed to generate the training trajectories.
    We need to train multiple value functions and policies in parallel for some use-cases and are therefore off-policy.
    Many of the current state-of-the-art RL algorithms are off-policy algorithms.
    Being off-policy is also useful as it allows us to run a learned policy in parallel distributed on hundreds of actors and having a centralized learner who uses the generated trajectory to update a slightly off-policy policy.
    \end{itemize}

All of the three elements are very useful, and one does not want to give them up.
In practice, many RL architectures successfully use all three elements of the deadly triad, like DQN \citep{Mnih2015}.
In the example of DQN, they use many tricks to cope with the instability and prevent the estimates from divergence through training with experience replay and occasionally freeze the target network.

\clearpage

\section{Transfer Learning}
\label{sec:transferLearning}
In reinforcement learning, an agent usually starts with a random policy.
The agent then has to learn an optimal policy for the target task using no prior knowledge.
For challenging target tasks, for example, due to sparse reward signals or poor state representations, the agent might learn very slowly or entirely fails to learn at all.

Transfer learning is one area of research that tries to speed up the training of RL agents by transferring knowledge from one or more source task MDPs to a target task.
Instead of learning to solve the target task tabula rasa, the agent acquires knowledge on one or more source tasks.
The knowledge can be transferred in form of samples \citep{10.1145/1390156.1390225}, options \citep{DBLP:conf/aaai/SoniS06}, policies \citep{FERNANDEZ2010866}, models \citep{FACHANTIDIS201323} or value functions \citep{AAMAS05-transfer}.
In the case of policy and value function transfer, the parameters of a policy or value function obtained by training on one or multiple source tasks can be used to initialize the policy or value function of the agent.
Transferring the policy or value function leads to a bias in the action selection towards the experience acquired in the source task.

\subsection{Evaluation Metrics for Transfer Learning}
\label{sec:TLMetrics}
To quantify the benefits gained from transfer learning, we need meaningful metrics.
Typically we compare the learning curve on the target task for an agent after transfer with a tabula rasa agent.
We consider the following three metrics:
\begin{itemize}
    \item \textbf{Time to Threshold}: The time to threshold computes how much faster an agent with knowledge transfer reaches a reward threshold compared to a tabula rasa agent.
    The time can be measured by CPU / GPU instructions, wall clock time, episodes, or steps.
    \item \textbf{Jumpstart}: This measurement quantifies the initial performance boost we gain as a result of knowledge transfer.
    \item \textbf{Asymptotic Performance}: The asymptotic performance compares the final performance increase at the end of training.
\end{itemize}

\begin{figure}[!h]
    \centering
    \begin{subfigure}{.4\textwidth}
        \centering
        \includegraphics[width=.99\linewidth]{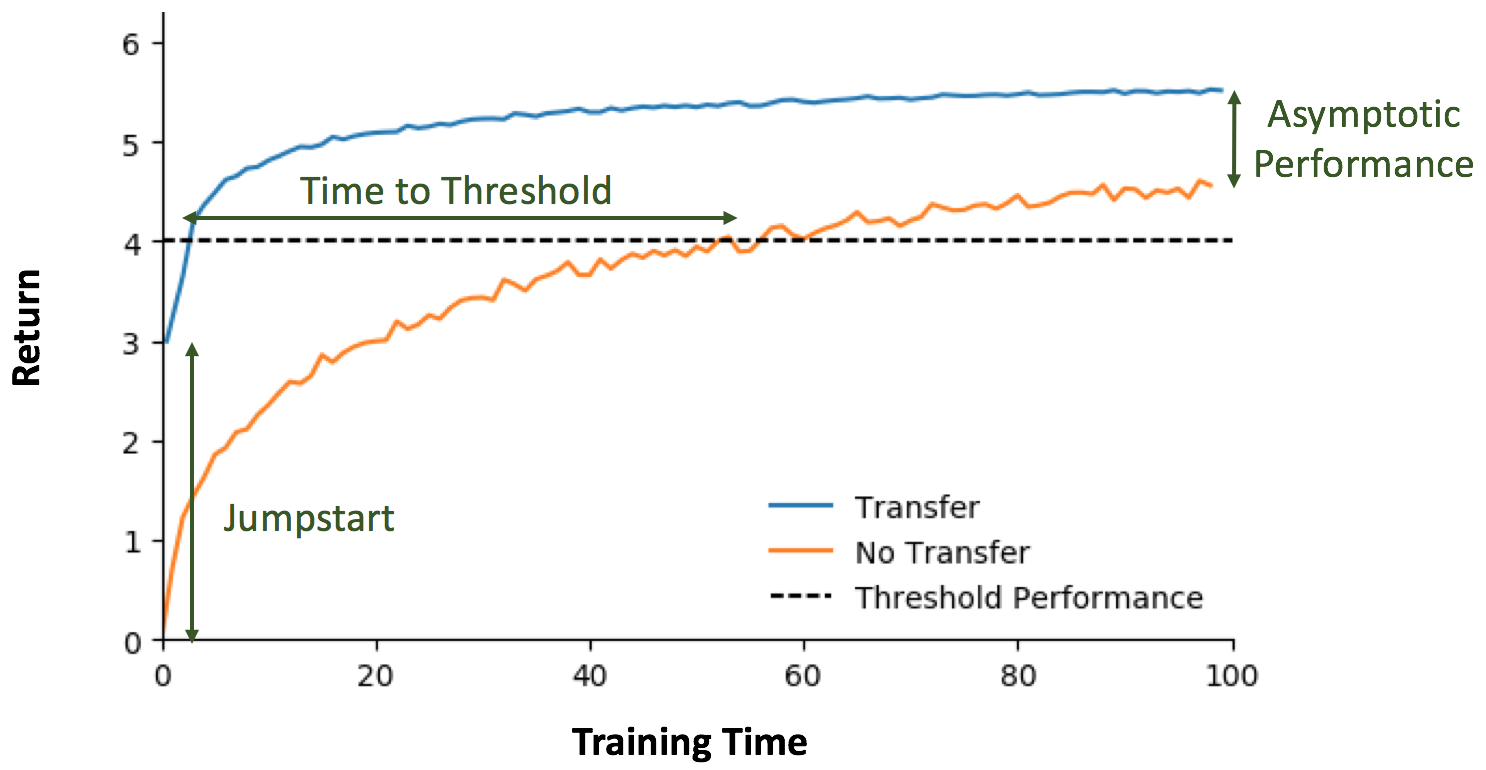}
        \caption{Weak transfer}
    \end{subfigure}%
    \begin{subfigure}{.4\textwidth}
        \centering
        \includegraphics[width=.99\linewidth]{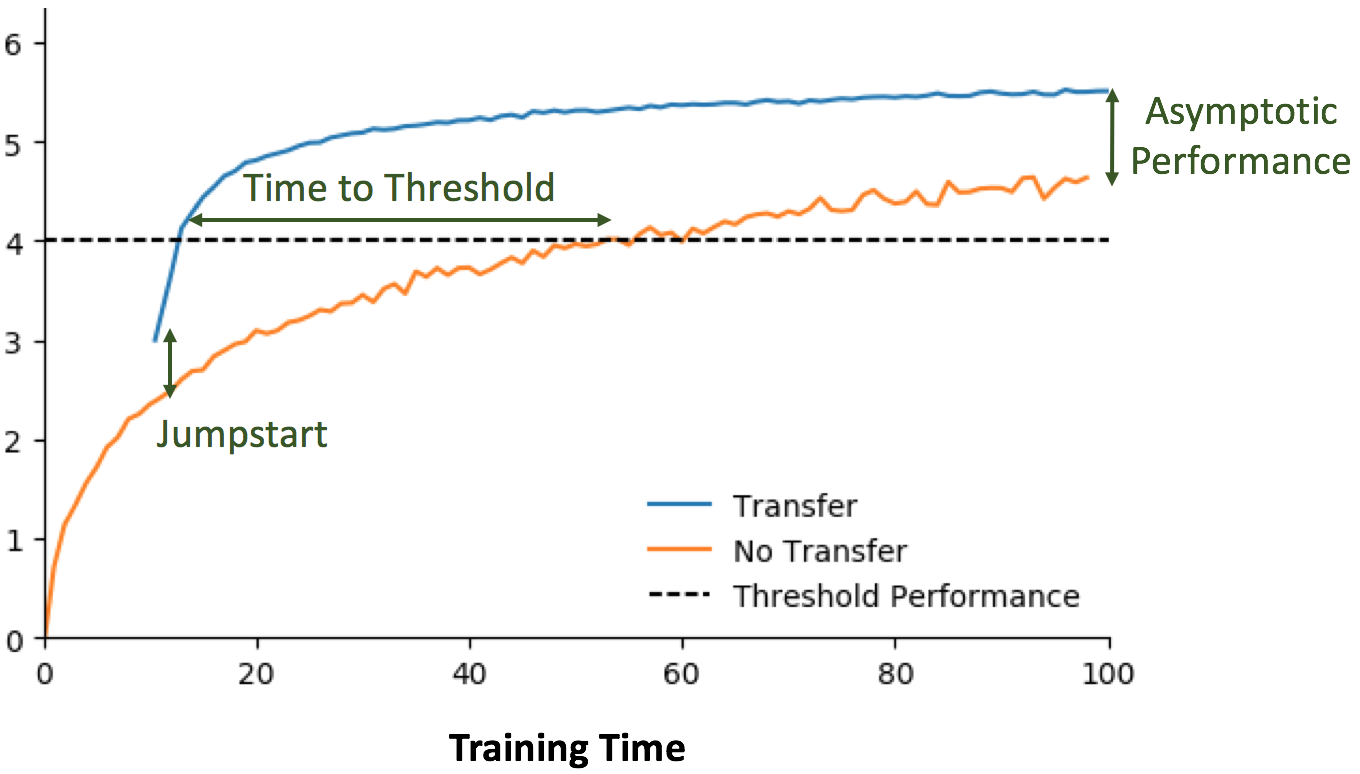}
        \caption{Strong transfer}
    \end{subfigure}
    \caption{Performance metrics for transfer learning using (a) weak transfer and (b) strong transfer (figure by \citet{DBLP:journals/jmlr/NarvekarPLSTS20}).}
    \label{fig:transferlearning_metrics}
\end{figure}

When comparing tabula rasa agents to agents that use transfer learning, we need to specify if we want to include the time spent learning the source tasks in our metrics.
We talk about \textit{weak transfer} when we do not include the costs of training in source tasks.
If we consider the time spent in the source tasks when calculation the metrics, we measure the \textit{strong transfer}.
In \cref{fig:transferlearning_metrics} we illustrate the three metrics once with a weak transfer and once with a strong transfer.
As the strong transfer includes costs for training in the source task, the transfer curve starts with a delayed training time in \cref{fig:transferlearning_metrics} (b). 

\clearpage

\section{Curriculum}
\label{sec:curriculumBackground}
We define a curriculum as a concept that organizes past experiences and schedules future experiences by training on tasks.
Every task is modeled as a Markov Decision Process.

In the following we use the curriculum definition of \citet{DBLP:journals/jmlr/NarvekarPLSTS20}:
\begin{definition}[Curriculum]\label{def:Curriculum}
Let $\mathcal{T}$ be a set of tasks, where $m_i = (\mathcal{S}_i,\mathcal{A}_i,p_i,r_i)$ is a task in $\mathcal{T}$ and $i$ is the task identifier.
Let $\mathcal{D}^\mathcal{T}$ be the set of all possible transition samples from tasks in $\mathcal{T}$: $\mathcal{D}^\mathcal{T} = \{(s,a,r,s') | \exists m_i \in \mathcal{T} \text{ s.t. } s \in \mathcal{S}_i, a \in \mathcal{A}_i, s' \sim p_i(\cdot|s, a), r \leftarrow r_i(s, a, s')\}$. 
A curriculum $C = (\mathcal{V}, \mathcal{E}, g, \mathcal{T})$ is a directed acyclic graph, where $\mathcal{V}$ is the set of vertices, $\mathcal{E} \subseteq \{(x, y) | (x, y) \in \mathcal{V} \times \mathcal{V} \wedge x \neq y\}$ is the set of directed edges, and $g: \mathcal{V} \rightarrow \mathcal{P}(\mathcal{D}^\mathcal{T})$ is a function that associates vertices to subsets of samples in $\mathcal{D}^\mathcal{T}$, where $\mathcal{P}(\mathcal{D}^\mathcal{T})$ is the power set of $\mathcal{D}^\mathcal{T}$. 
A directed edge $(v_j, v_k)$ in $C$ indicates that samples associated with $v_j \in \mathcal{V}$ should be trained on before samples associated with $v_k \in \mathcal{V}$. 
All paths terminate on a single sink node $v_t \in \mathcal{V}$.
\end{definition}

If the curriculum is created online, then the edges are added dynamically during the agent's training.
If the curriculum is created offline, then the graph is created beforehand.

One simplification to the curriculum definition is the single-task curriculum, where all samples stem from a single task. In a single-task curriculum, we rearrange the order in which we train on the experience samples as in prioritized experience replay \citep{Schaul2016}.

\begin{definition}[Single-task Curriculum]\label{def:SingleTaskCurriculum}
A single-task curriculum is a curriculum $C$ where the cardinality of the set of tasks considered for extraction samples $| \mathcal{T} | = 1$, and consists of only the target task $m_t$.
\end{definition}

A second simplification of the curriculum definition is the task-level curriculum.
In the task-level curriculum, we define a directed acyclic graph of intermediate tasks.
The main challenge here is how to order the intermediate tasks such that the agent can constantly learn to solve more complex tasks while preventing catastrophic forgetting on already solved tasks.
The mapping function $g$ determines the set of samples $\mathcal{D}^\mathcal{T}_i$ that are available at the next vertex.
There are multiple works on task-level curricula \citep{Svetlik_Leonetti_Sinapov_Shah_Walker_Stone_2017, 8827566}.

\begin{definition}[Task-level Curriculum]\label{def:TaskLevelCurriculum}
For each task $m_i \in \mathcal{T}$, let $\mathcal{D}^\mathcal{T}_i$ be the set of all samples associated with task $m_i: \mathcal{D}^\mathcal{T}_i = \{ (s, a, r, s') | s \in \mathcal{S}_i, a \in \mathcal{A}_i, s' \sim p_i(\cdot | s, a), r \leftarrow r_i(s,a,s') \}$. A task-level curriculum is a curriculum $C = (\mathcal{V}, \mathcal{E}, g, \mathcal{T})$ where each vertex is associated with samples from a single task in $\mathcal{T}$. Thus, the mapping function g is defined as $g: \mathcal{C} \rightarrow \{ \mathcal{D}^\mathcal{T}_i | m_i \in \mathcal{T} \}$.
\end{definition}

The simplest form of a curriculum is the sequence curriculum.
In the sequence curriculum, all nodes have an indegree and outdegree of at most 1.
If combined with the task-level curriculum, we end up with the task-level sequence curriculum, which is an ordered list of tasks $[m_1, m_2, \dots m_t]$

\begin{definition}[Sequence Curriculum]\label{def:SequenceCurriculum}
A sequence curriculum is a curriculum $C$ where the indegree and outdegree of each vertex $v$ in the graph $C$ is at most 1, and there is exactly one source node and one sink node.
\end{definition}

\section{Curriculum Learning}
\label{sec:curriculumLearningBackground}
In curriculum learning, we try to find the optimal order in which we feed experience from different source tasks to the agent in order to maximize one of the metrics defined in \cref{sec:TLMetrics}.
The intuition behind curriculum learning is that through generalization over source tasks and knowledge transfer through increasingly more complex tasks, we can increase the sample efficiency or asymptotic performance of our training algorithms.
There are three key elements to curriculum learning:

\begin{itemize}
    \item \textbf{Task Generation}: In order to produce a beneficial curriculum, we need a good set of source tasks.
    The tasks should provide an increasing difficulty.
    The tasks have to be similar to the target task. 
    Otherwise, we might end up with a negative transfer, and using a curriculum over such tasks is hurtful for solving the desired target task.
    Task generation is, therefore, an important part of curriculum learning.
    In a task-level curriculum, the tasks are the nodes of the curriculum graph.
    The task may be generated online during training or pre-specified.
    \item \textbf{Sequencing}: Sequencing is concerned with the ordering of the experience samples.
    According to the graph definition in \cref{sec:curriculumBackground}, sequencing defines the vertices in our curriculum graph.
    These vertices can either be defined online during training or offline before training the agent.
    In \cref{sec:taskSequencing} we go into details of this aspect to curriculum learning.
    \item \textbf{Transfer Learning}:  In order to benefit from a curriculum of tasks, we need a method to transfer knowledge through the curriculum.
    In \cref{sec:transferLearning} we explained multiple methods for transfer knowledge from one or more source tasks to the target task.
    In curriculum learning, we repeatedly transfer knowledge from task to task, whereas we have only one transfer step in standard transfer learning.
\end{itemize}

\begin{figure}[!h]
    \centering
    \includegraphics[width=0.5\linewidth]{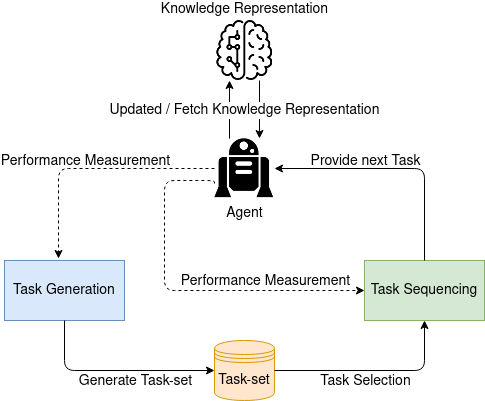}
    \caption{We visualize the interaction between the three key elements of curriculum learning.
    Task generation is concerned with generate a set of tasks for the curriculum. 
    Task sequencing selects a task out of the task set for the agent.
    The agent use knowledge obtained by training on previously selected task through transfer learning.
    After or during training on the new task, the obtained knowledge is stored.
    The task generation and sequencing can be done online and dependent on the student's current performance or offline before training.
    }
    \label{fig:KeyElementsCL}
\end{figure}

We visualized the interaction between the three key elements in \cref{fig:KeyElementsCL}.

When evaluating the curriculum, we use the same metrics as in \cref{sec:TLMetrics}.
Those metrics have to be extended as we have to consider the costs of building the curriculum.
It is not clear how the work for the task sequencing or task generation done by humans should be included into the strong transfer metrics.
In our work, we ignore those costs for the sake of simplicity.

\subsection{Curriculum Learning Categorization}
\label{sec:clCategorization}
We can categorize a curriculum learning approach along six dimensions, organized by attributes (in bold) and their respective values (in italics). This categorization was introduced in \citet{DBLP:journals/jmlr/NarvekarPLSTS20}:

\begin{enumerate}
    \item \textbf{Intermediate task generation}: \textit{target / automatic / domain experts / naive users}. The set of tasks can be either defined offline before training or online during training.
    One can also specify a single task curriculum called \textit{target} where only the target task is used.
    The tasks can be specified by a human, either a \textit{domain expert} \citep{schraner_2020} or a \textit{naive user} with no special domain knowledge.
    There are also methods to \textit{automatically} generate tasks using a set of rules or a generative process as in \citet{10.1145/3321707.3321799}.
    
    \item \textbf{Curriculum representation}: \textit{single / sequence / graph}. The simplest way to represent a curriculum is a \textit{single} task curriculum, where we simply reorder the recorded experience \citep{Schaul2016, NIPS2017_453fadbd}.
    When using multiple tasks we can either represent the curriculum as a simple \textit{sequence} of tasks \citep{schraner_2020} or as directed acyclic \textit{graph} of tasks \citep{Svetlik_Leonetti_Sinapov_Shah_Walker_Stone_2017}. In the task representation, we can allow many-to-one, one-to-many, and many-to-many knowledge transfer.
    
    \item \textbf{Transfer method}: \textit{policies / value function / task model / partial policies / shaping reward / other / no transfer}.
    In \cref{sec:transferLearning} we specified different forms of knowledge transfer.
    The weights of \textit{policy} \citep{pmlr-v123-scheller20a} or \textit{value} \citep{FERNANDEZ2010866, JMLR07-taylor, AAMAS05-transfer} functions can be transferred from task to task.
    One can learn \textit{task models} \citep{FACHANTIDIS201323} and transfer those from task to task, learn an auxiliary \textit{reward function}, or extract options \citep{Sutton99betweenmdps}  and transfer those to the next task. 
    
    \item \textbf{Curriculum sequencer}: \textit{automatic / domain experts / naive users}. 
    The curriculum sequencing is concerned with the task switches during training.
    The switch can happen \textit{automatic} upon specific rules, through a teacher or other methods.
    In \cref{sec:taskSequencing} we go into detail about this aspect of curriculum learning.
    We can also sequence the task manually by \textit{domain experts} or \textit{naive users.}
    
    \item \textbf{Curriculum adaptivity}: \textit{static / adaptive}.
    The adaptivity of a curriculum specifies if the curriculum is completely defined before training or if it is dynamically adapted online during training.
    A \textit{static} curriculum is defined before training, a \textit{adaptive} curriculum is changed during training.
    Adaptive curricula can use the learning progress to estimate, e.g., if a task is easy or hard to learn at the current stage. Static curricula incorporate problem-specific knowledge.
    
    \item \textbf{Evaluation metric}: \textit{time to threshold / asymptotic / jumpstart / total reward}.
    In \cref{sec:TLMetrics} we introduced metrics to quantify the benefits gained from curriculum learning.
    When calculating those metrics, we have to decide if we want to measure the weak or strong transfer.
    
\end{enumerate}

\clearpage

\subsection{Task sequencing}
\label{sec:taskSequencing}
Task sequencing is concerned with how the tasks can be sequenced in order to provide a curriculum.
The tasks need to be sequenced in a way such that the current task at hand is just hard enough to solve.
It might also be of interest to add an already learned task that the agent forgets about to allow live long learning.

In \cref{sec:relatedWork} we already introduced a few methods for task sequencing.
In this section, we detail two sequencing methods that use the teacher-student setup with the MDP formulation.

\citet{ijcai2017-353} formulates curriculum learning as the interaction of two MDPs. 
A student MDP describes the currently selected task, and a teacher MDP models the selection of the next task for the student.
They denote the teacher MDP as a fully observable MDP, where the state space is the set of policies the learning agent can represent.
The state is represented as the parameters of the policy.
The final states are defined as policies where the return on the target task is higher than a specific threshold.
The action space is the set of tasks a student agent can train on.
Taking an action results in the student training on the selected task for a fixed number of steps or until convergence.
The transition function describes how the student agents policy changes as a result of learning a task.
The reward function is defined as the time needed by the student agent to learn a policy that results in a return surpassing a certain threshold on the target task (time to threshold).
The teacher wants to minimize this time to threshold. Therefore the reward is encoded as the negative of the expected time needed to learn the target task starting from a given policy.
A recursive Monte-Carlo algorithm optimizes the teacher agent, and the student is a tabular RL agent with tile coding, trained with Sarsa($\lambda$) and a value function transfer.
In their follow-up paper, they investigated reward shaping as an additional transfer method \citep{AAMAS19-Narvekar}.

\citet{8827566} uses a similar approach, but the state is not fully observable, making the MDP a partially observable Markov decision process (POMDP).
The state and action space of the teacher MDP are the same as in \citet{ijcai2017-353}, but the teacher has no access to the internal parameters of the student agent.
The observation is the reward of the student obtained on the currently selected task.
The reward is the average reward of the student evaluated on all tasks at the end of a teacher step.
One could also optimize for the reward of the target task, but initially, the student might not archive a reward on the target task leaving the teacher without a meaningful signal.

A comprehensive survey of task sequencing methods is provided by \citet{DBLP:journals/jmlr/NarvekarPLSTS20}.
\clearpage
\chapter{Methods}
\label{sec:methods}
In this chapter, we describe the methods used in this thesis.
We characterize our curriculum learning approach along the six dimensions introduced in \cref{sec:curriculumLearningBackground}.
Next, we detail our transfer method and the automated curriculum sequencing approach.
We propose multiple types of observations and reward signals for our curriculum MDP.
Finally, we describe our set of baselines and the evaluation protocol.

\section{Curriculum Learning}
In this section, we characterize our curriculum learning approach.
We use a task-level sequence curriculum. A combination of the task-level curriculum defined in \cref{def:TaskLevelCurriculum} and a sequence curriculum (see \cref{def:SequenceCurriculum}).
Our curriculum learning has the following properties:

\begin{enumerate}
    \item \textbf{Intermediate task generation}: 
    The tasks, represented as nodes in the curriculum graph, are predefined before training.
    We use a subset of the tasks provided by the grid world environment and the Google Research Football environment.
    We initially selected the tasks and kept this selection fixed throughout this thesis.
    The exact task selection is described in \cref{sec:gridworldSetup} and \cref{sec:gfootballSetup}.
    \item \textbf{Curriculum representation}: 
    We represent the curriculum as a task-level sequence.
    We only allow one-to-one knowledge transfer between our source tasks and the target task.
    A source task can be visited multiple times in the sequence.
    Therefore the agent is allowed to retrain on already known tasks.
    \item \textbf{Transfer method}: 
    We experiment with three transfer methods.
    First, we copy the policy and value function weights from task to task to transfer the learned policy and value function.
    Second, the learned value function, obtained by training on the previous task, calculates an additional reward signal that we added to the environment's reward signal.
    Third, we experiment with a combination of the first and second method.
    In \cref{sec:methodsTransferMethod} we go into detail about our knowledge transfer methods.
    \item \textbf{Curriculum sequencer}: 
    In this work, we experiment with automated task sequencers.
    We formulate the task sequencing problem as a curriculum Markov Decision Process where we have full control over the student's MDP.
    In \cref{sec:methodCMDP} we formalize the CMDP and explain our approach in detail.
    \item \textbf{Curriculum adaptivity}: 
    We use an adaptive curriculum.
    The teacher performs task sequencing online to select tasks with a suitable learning potential for the current learning stage.
    Our set of source tasks is defined beforehand and kept fix throughout training.
    \item \textbf{Evaluation metric}: 
    We evaluate our agents using a weak transfer with the asymptotic performance improvement.
    We compare our work against agents trained in \citet{schraner_2020}, which uses manually defined curricula.
\end{enumerate}

\subsection{Transfer Method}
\label{sec:methodsTransferMethod}
We use two types of knowledge transfer methods: policy and value function transfer and reward shaping.
Our student agent starts with a random initialized policy and value function.
After training for a fixed amount of steps at time-step $t$ on a task $m_t$, we switch to a new source task $m_{t+1}$.
The weights of the policy and value function obtained by training on task $m_t$ are used to initialize the agent before training on task $m_{t+1}$. 
Our student uses weight sharing for the policy and value function. Therefore we copy all neural network weights from task to task.

We assume that both the policy and value function transfer from task to task as the dynamics of the environment does not change in our set of tasks $\mathcal{V}$.
Therefore, a state with a true high value in one environment $m_i \in \mathcal{V}$ has also a true high value in another environment $m_j \in \mathcal{V}$.
The same applies to our policy.
The environments in our set of source tasks differ in their initial state distribution, state space and complexity.
All tasks in $\mathcal{V}$ share the same set of actions $\mathcal{A}$, environment dynamics function $p$ and reward function $r$.

It would also be possible to only transfer the weights of the shared embedding together with either the policy head or the value head from task to task.
Like this, either the policy or the value are initialized randomly for each task.
The gradients for the policy and the value function flow through the same model in our network architecture with weight sharing. 
If either the policy or the value function is initialized randomly, this could lead to catastrophic updates to the shared weights.
Therefore, we transfer all the neural network weights from task to task.

Additionally to the policy and value function transfer, we experiment with a shaped reward signal.
We use the value function $v_{m_{t}}$ obtained by training on task $m_t$ as an additional reward signal when training on the next task $m_{t+1}$.
The reward $r_i$ at time-step $i$ is therefore $r_i = r_i + v_{m_{t}}(s_i)$
We expect that the value function can help to overcome the spares reward signal provided by the environment.
For the first task $m_1$ in our curriculum, we do not add a shaped reward signal as we do not have a value function at hand.
Only the value function of the previous task is added as a shaped reward.
One could also use the average of the last $n$ value functions as a shaped reward signal.
We leave this experiment to future work.

The two transfer methods described can be combined, leading to a transfer method with policy and value function transfer and a shaped reward signal.
We also experiment with this combined transfer method.

\section{Teacher Curriculum Markov Decision Process}
\label{sec:methodCMDP}
In this section, we formulate the sequencing problem as a Markov Decision Process.
In this formulation, we define curriculum learning as an interaction between two types of MDPs.
The first MDP is the standard RL MDP modeling the \textit{student} interacting with a task.
The second one is a higher level meta-MDP called curriculum MDP (CMDP).
We use the CMDP to model a teacher selecting tasks for the student.
The CMDP is a 4-tuple $(\mathcal{S}, \mathcal{A}, p, r)$, where $\mathcal{S}$ is the set of all possible states equal to all possible policies the student can represent.
The action space $\mathcal{A}$ is the set of tasks the teacher can assign to the student.
Taking an action in the CMDP equals to an entire training cycle on the selected task in the student MDP for a fixed amount of steps.
The environment's dynamics function $p$ models the transition from one student policy to the next student policy after taking an action in the CMDP.
The dynamics function is unknown.
The reward for the CMDP is defined by the reward function $r$.
In \cref{sec:methodRewardFunction} we detail two reward signals used in this research.

\begin{figure}[htb]
    \centering
    \includegraphics[width=.45\textwidth]{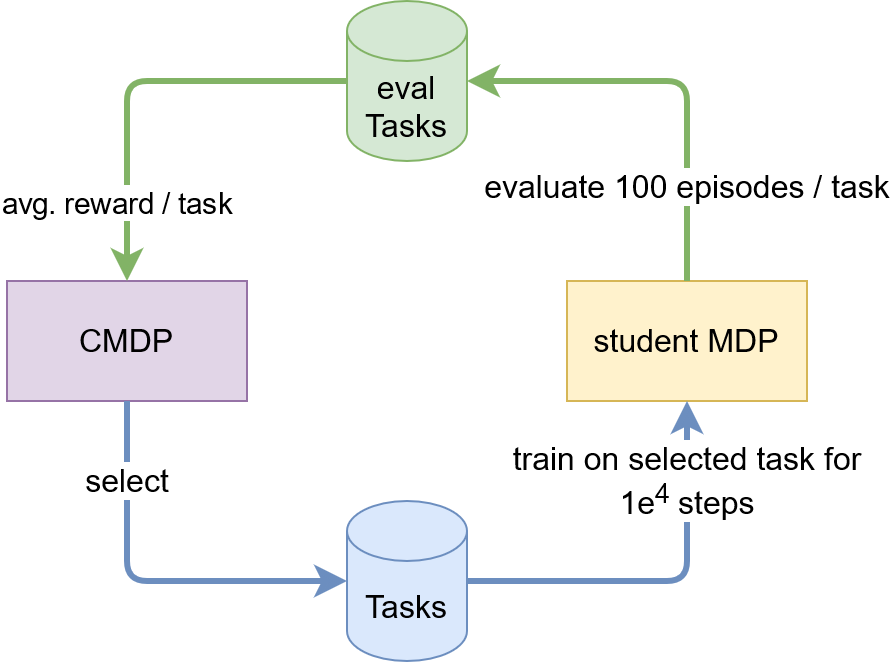}
    \caption{Teacher-student interaction for task sequencing.}
    \label{fig:teacherStudentSetting}
\end{figure}

The interaction between the CMDP and the student MDP is shown in \cref{fig:teacherStudentSetting}.
The teacher selects a task, and the student trains for n steps on the selected task.
After the student training step, we evaluate the student on a set of evaluation tasks and feed the evaluation result to the CMDP.
In our research the set of evaluation tasks $\mathcal{V}_\text{eval}$ is equal to the learning tasks $\mathcal{V}_\text{learn}$.

Now that we have defined the sequencing problem as a CMDP, we can use reinforcement learning to find an optimal policy.
We are using PPO to learn a teacher policy while training the student at the same time.
With this approach, we have to deal with noise from the student training when performing suboptimal task switches, especially at the beginning of our training procedure.
The teacher has to learn how to perform task sequencing while the student has to learn how to solve the selected environment.
Both are acting randomly, which may lead to the teacher selecting too tricky tasks, and the student has a hard time learning from that task.

\subsection{Teacher Observation}
\label{sec:teacherObservation}
We experiment with different types of observations for our teacher agent.
The state of the CMDP is the current policy of the student agent.
We are using a neural network to approximate the optimal student policy $\pi_s^*$.
Therefore the student weights $\theta_s$ are the state of the CMDP, and they fulfill the Markov property.
It is unclear how we should feed this state representation into our teacher agent.
We could flatten the weights $\theta_s$ to a feature vector, but this would leave us with an enormous input vector. 
In our case, we have roughly $400'000$ weights in our student network.
Using such a large input vector is not feasible due to memory limits.
Additionally, we assume that such a feature vector is not easy to interpret for our teacher agent. It is hard to relate $\theta_s$ to the student's performance and the optimal next task to select.

We use principal component analysis (PCA) \citep{doi:10.1080/14786440109462720} to find a reduced representation for $\theta_s$.
We build a training set containing student weights at different training stages and then fit PCA to this dataset. 
This dataset is obtained by saving the weights of the student network for different tasks at different learning stages.
At the end of each student training iteration, we do dimensionality reduction to the first n principal components and use this reduced representation as an input for our teacher.

There are other ways to get a reduced representation for $\theta_s$ like model distillation or auto-encoders, each with its downsides.
We leave the question for an optimal way to bring $\theta_s$ into a meaningful representation to future work.
In this work, we refer to this observation type as \textit{pca input} (PCA).

In our work we also experiment with partial observable curriculum MDPs.
We work with six different types of manual defined observation types:
\textit{reward history} (RH), \textit{previous task reward} (PTR), \textit{learning progress} (LP), \textit{absolute learning progress} (ALP), \textit{exponential moving average} (EMA), and \textit{fast and slow exponential moving average} (FS-EMA).

\paragraph{\textit{Reward History}}:
Instead of using the students weights $\theta_s$ we try to represent the learning potential per task.
We can use a tuple of the student's average reward, obtained in the evaluation cycle at the end of a CMDP step, on each task together with the time-step when this task was last sampled:
\begin{equation}
    \text{RH} = \{(\overline{r^m_\text{eval}}, t^m_\text{last sampled}) | m \in \mathcal{V}\}
\end{equation}
Where $\overline{r^m_\text{eval}}$ is the average reward on all evaluation episodes for task $m$ and $t^m_\text{last sampled}$ is the CMDP time-step when task $m$
last has been sampled.
If the task $m$ never has been sampled $t^m_\text{last sampled}$ equals to 0.

\paragraph{\textit{Previous Task Reward}}:
Instead of using the average rewards of all tasks, we can input the last selected task, one-hot encoded, together with the average reward on that task, obtained in the evaluation cycle at the end of a CMDP step.

\paragraph{\textit{Learning Progress}}:
The learning progress is defined as the difference between the average reward, obtained in the evaluation cycle at the end of a CMDP step, of the current and the previous time step per task:
\begin{equation}\label{eq:lp}
    \text{LP} = \{(\overline{r^m_{t}} - \overline{r^m_{t-1}}) | m \in \mathcal{V} \}
\end{equation}
Where $\overline{r^m_{t}}$ is the average reward on the task $m$ at time-step $t$ and $\overline{r^m_{t-1}}$ is the average reward on task $m$ at the previous time-step $t-1$.

\paragraph{\textit{Absolute Learning Progress}}:
ALP is simply the absolute value of LP:
\begin{equation}
    \text{ALP} = \{(|\overline{r^m_{t}} - \overline{r^m_{t-1}}|) | m \in \mathcal{V}\}
\end{equation}
The intuition behind using the absolute value is that a task at the stage of forgetting, resulting in a negative LP, should be treated similarly to a task with a steep learning curve.
This learning progress representation is inspired by \citet{DBLP:conf/corl/PortelasCHO19}.

\paragraph{\textit{Exponential Moving Average}}:
We can interpret the history of evaluation rewards after every CMDP cycle as a time-series.
We use an exponential moving average over the history of rewards to estimate the next reward.
Depending on the $\alpha \in [0,1]$ value used to calculate the EMA we assign more weight on recent samples than on old ones.
The exponential moving average over a vector $\mathbf{x}$ is defined as:
\begin{equation}
    \text{ema}(\mathbf{x}_t) = 
    \begin{cases}
    \alpha * \mathbf{x}_t + (1 - \alpha) ema(\mathbf{x}_{t-1}), & t > 1 \\
    x_1, & t = 1
    \end{cases}
\end{equation}
Where $\text{ema}(\mathbf{x}_t)$ is the value of the EMA at any time period $t$ and $\mathbf{x}_t$ is the value at a time period $t$.
Therefore the EMA input is:
\begin{equation}
    \text{EMA} = \{[ema(\mathbf{x}^m_t)] | m \in \mathcal{V} \}
\end{equation}
$\mathbf{x}^m$ is the history of the average reward, obtained in the evaluation cycle at the end of a CMDP step, for task $m$ at all CMDP time-steps.
The last time step is denoted as $t$.

\paragraph{\textit{Fast and Slow Exponential Moving Average}}:
\citet{DBLP:journals/corr/abs-2106-14876} introduce a smoothed EMA version, where they calculate a fast and a slow EMA with a high $\alpha$ and a low $\alpha$ respectively.
The fast and slow EMA is then defined as the difference between the two EMAs:
\begin{equation}
    \text{FS-EMA} = \{[ema_\text{fast}(\mathbf{x}^m_t) - ema_\text{slow}(\mathbf{x}^m_t)] | m \in \mathcal{V} \}
\end{equation}
In our work, we wanted to test if this method is superior to a normal EMA with a tuned $\alpha$ value.

\subsection{Reward Signal}
\label{sec:methodRewardFunction}
In reinforcement learning, a meaningful reward signal is crucial for success.
We need to define a reward signal for our CMDP that encodes our intention and is rich enough for the teacher agent to learn fast.
The goal of the teacher agent is to perform task sequencing such that the asymptotic reward on the target task increases.
Therefore we can use the student's average reward on the target task after a CMDP step as a reward signal.
We call this reward signal \textit{target task reward}.
Typically, the target task is hard to solve. Therefore the reward at the beginning of training is usually 0.
This reward signal is not meaningful for the teacher as it does not tell if the student is making progress in easier environments or if the student is completely lost.
Such a sparse reward signal leads to less sample efficiency and as samples in the CMDP are extremely costly, we want our teacher to learn fast.

We assume that the source tasks in $\mathcal{V}$ are related to the target task.
Therefore, if our student achieves a reward on the source tasks, this is a step towards solving the target task.
If this is true, then we can overcome the sparse reward signal by defining a new reward signal \textit{source task reward}:
\begin{equation}
    r_\text{teacher} = \sum_{m \in \mathcal{V}} \overline{r_m}
\end{equation}
Where $\overline{r_m}$ is the average reward on task $m$ obtained in the evaluation cycle at the end of a CMDP step.

\clearpage

\subsection{Action}
The action space in the CMDP contains all tasks in $\mathcal{V}$.
Taking an action equals to selecting a task $m \in \mathcal{V}$, changing the student's environment to the selected task, and then train the student for a predefined amount of steps.
After the training, the student is evaluated on all tasks in $\mathcal{V}$ and the evaluation results are passed to the CMDP agent.
The evaluation is carried out according to the \cref{sec:evalProtocol}.

In our experiment, we train the student for $100'000$ steps.
One could also train the student until convergence, surpassing a threshold, or make the number of steps part of the teacher's action space.
It is unclear when the student will converge, which could lead to very compute-intensive CMDP steps, and if the agent can surpass the threshold, which could lead to an infinite CMDP step.
Therefore, we do not consider these two approaches.
We want to keep the teacher's action space as simple as possible. Therefore we leaf integrating the number of student training steps into the action space for future work.

\subsection{Network Architecture}
\label{sec:teacherArchitecture}
We keep the teacher network architecture simple.
We use a multilayer perceptron (MLP) with three hidden layers with 64, 128, and 64 nodes, respectively, and a ReLU \citep{agarap2018learning} activation function.
We make use of weight sharing as we do it for our student agent.
The policy and value function are represented as individual heads on top of the MLP feature embedding.

We also experiment with LSTMs \citep{hochreiter1997long} in order to provide the teacher a memory.
In this setting, we use the same MLP for the feature embedding and then feed the feature vector into an LSTM with a hidden state of 128.
The hidden state is then used as an input for the policy and value function heads.

Due to computation limits, we were not able to tune the network architecture.
Another network architecture likely yields better results.

\clearpage

\section{Baselines}
We compare our teacher task sequencing approach against four baselines: \textit{uniform}, \textit{LP sampling}, \textit{window} and \textit{thompson sampling}.
The last two are introduced in \citet{8827566}.

\paragraph{\textit{Uniform}}: 
This is the simplest baseline.
We select the next task by uniformly sampling them.

\paragraph{\textit{LP sampling}}:
In this baseline, we select the task with the highest learning progress as defined in \cref{eq:lp}.
If two or more tasks have the same LP, we sampling one of those tasks uniformly.

\paragraph{\textit{Thompson sampling}}:
Similar to LP sampling we sample the next task according to a learning progress metric.
In the Thompson sampling baseline, we sample the task with the highest average reward after the last CMDP step.

\paragraph{\textit{Window}}:
In the window algorithm, we approximate the learning progress with linear regression.
We store the history of average evaluation rewards.
At each CMDP step, we fit a linear regression to the reward history.
The regression coefficient per task is used to estimate the steepness of the learning curve.
The task with the steepest learning curve is selected.

\section{Evaluation Protocol}
\label{sec:evalProtocol}
The goal of the teacher task is to improve the asymptotic performance.
We evaluate our approach using a weak transfer to allow comparison with the results on the Google Research Football environment from our previous work \citep{schraner_2020}.
The asymptotic performance is defined as the increase in the average reward at the end of training.

Additionally, we evaluate how well the trained student agent generalizes across the environments.
The agent is trained overall environments, and it is interesting how well the agent performs on those environments.
To measure the generality, we calculate the total average return, which is the sum of all average returns for all tasks in $\mathcal{V}$, and the percentage of environments solved.
We declare an environment as solved when the average return is greater than zero.

After every CMDP step, we evaluate the student on every task in $\mathcal{V}$ for 100 episodes.
\clearpage
\chapter{Experiments and Results}
\label{sec:practicalWork}
In this chapter we describe the experiments used to evaluate our methods.
We begin with an introduction of the environments used, the experimental setup and then provide results of our baselines and teacher student experiments.
In depth experiments are carried out on a grid world environment, the most promising settings are then evaluated on the Google Research Football environment.
In the following chapters, we use the terms task and environment interchangeably. In our case, the environment is equal to the task to solve, e.g., there is only a single task per environment.
The description of the Google Research Football environment is taken from our previous work \citep{schraner_2020}.

\section{Grid World}
\label{sec:gridworld}
Grid world is a simple, lightweight, and fast environment for reinforcement learning.
In a Grid world environment, the agent must reach a target destination by navigating through a maze.
The difficulty of the environment ranges from very simple empty grids to complex mazes where the agent has to find keys in a specific order to unlock the target destination.
We use the minimalistic grid world (MiniGrid) implementation by \citet{gym_minigrid}. 

In MiniGrid, the world is an NxN grid of tiles.
Each tile in the grid world contains zero or one object, and each object has an associated discrete color and type.
The object types are wall, floor, door, key, ball, and goal.
Doors have a state open, closed, or locked and behave according to this state.
To open a locked door, the agent has to carry a key matching the door's color.
The agent can pick up and carry exactly one object (e.g., ball or key).

The simplicity of the grid world environment allows fast iteration and testing of multiple ideas.
Training a student agent in the CMDP setting in the grid world environment takes around 18 hours.
Training a teacher in the proposed CMDP setting is computationally expensive as we have to train a student agent.
The set of environments with their different levels of complexity is helpful for the curriculum learning setting.

Depending on the selected environment, the maximum number of steps changes.
An epoch in the simplest environment, \textit{Empty-5x5}, ends after 100 steps, the most difficult environment, \textit{KeyCorridor-S6R3}, terminates after 1080 steps.

\subsection{Grid World Environments}
\begin{figure}[htb]
    \centering
    \begin{subfigure}{.2\textwidth}
        \centering
        \includegraphics[width=.99\linewidth]{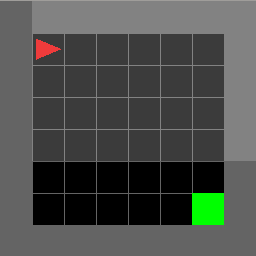}
    \end{subfigure}
    \begin{subfigure}{.2\textwidth}
        \centering
        \includegraphics[width=.99\linewidth]{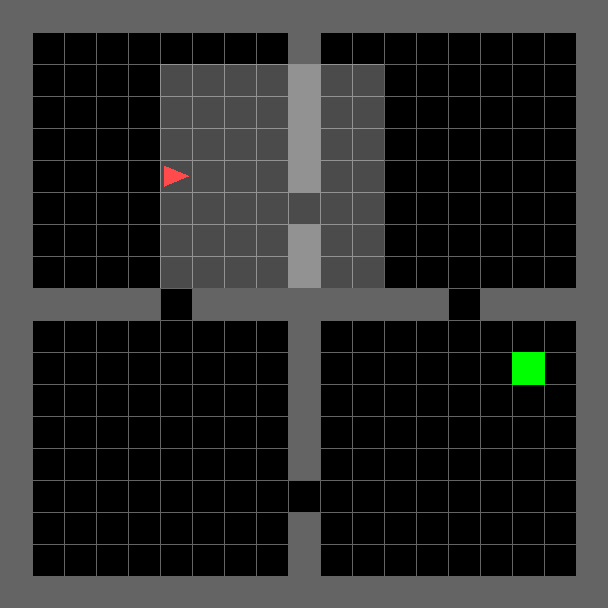}
    \end{subfigure}
    \begin{subfigure}{.2\textwidth}
        \centering
        \includegraphics[width=.99\linewidth]{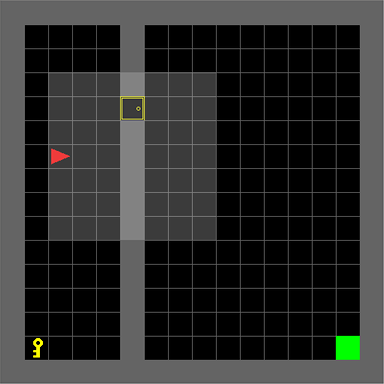}
    \end{subfigure}
    \begin{subfigure}{.2\textwidth}
        \centering
        \centering
        \includegraphics[width=0.99\linewidth]{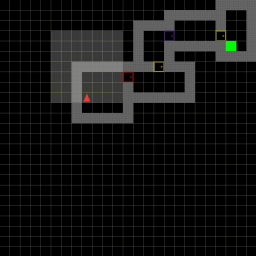}
    \end{subfigure}
    \\
    \begin{subfigure}{.2\textwidth}
        \centering
        \includegraphics[width=0.99\linewidth]{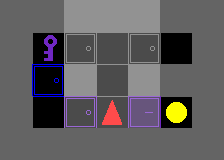}
    \end{subfigure}
    \begin{subfigure}{.2\textwidth}
        \centering
        \includegraphics[width=0.99\linewidth]{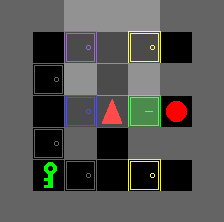}
    \end{subfigure}
    \begin{subfigure}{.2\textwidth}
        \includegraphics[width=0.99\linewidth]{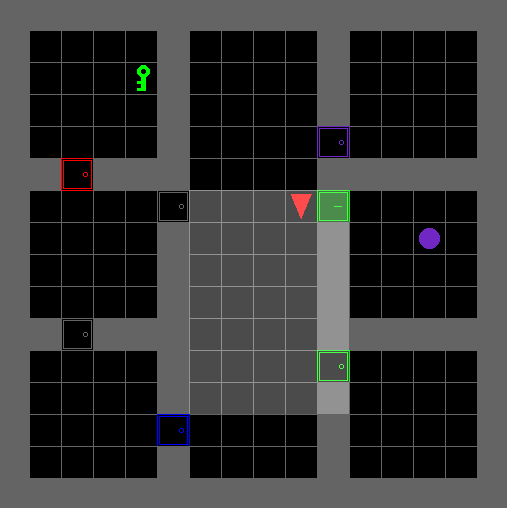}
    \end{subfigure}
    \caption{Visualization of a subset of the MiniGrid environments used in this work.
    The environment names from top left to bottom right: Empty-6x6, FourRooms, DoorKey-16x16, MultiRoom-N4S5, KeyCorridor-S3R2, KeyCorridor-S3R3, and KeyCorridor-S6R3.}
    \label{fig:miniGridEnvironments}
\end{figure}

We describe the MiniGrid environments in \cref{tab:miniGridEnvironments} and illustrate them in \cref{fig:miniGridEnvironments}.
In a grid world environment, the agent has to navigate through a maze and solve some puzzles to reach the green goal square.
In the KeyCorridor environment, the agent has to pick up a hidden ball behind a locked door.
To unlock that door, the agent must find the matching key, which is hidden in another room.
The agent has to explore the environment and move through open doors to find the hidden key.
The bottom row of \cref{fig:miniGridEnvironments} displays two different KeyCorridor environments.

\renewcommand{\arraystretch}{1.5}
\begin{table}[htb]
    \centering
    \begin{tabular}{ p{0.25\textwidth} p{0.65\textwidth} }
         \hline
         \textbf{Name} & \textbf{Description} \\ 
         \hline
         \raggedright\textit{Empty-[X]x[X]} & This environment is an empty room. 
         Upon reaching the green goal square, the student receives a sparse reward. 
         The agent is starting in a random position. 
         The value of X defines the grid size. 
         We use a 5 by 5, 6 by 6, 8 by 8, and 16 by 16 grid.\\ 
         \raggedright\textit{FourRooms} & The agent must navigate in a maze of four rooms. 
         Four gaps in the walls connect the rooms. 
         To obtain a reward, the agent must reach the green goal square. 
         Both the agent and the goal square are placed randomly in any of the four rooms. \\ 
         \raggedright\textit{DoorKey-[X]x[X]} & The agent must pick up a key in order to unlock a door and then reach the green goal square. 
         Due to the sparse reward signal, this environment is difficult to solve. 
         The door, wall, agent and green goal square are placed randomly.
         The value of X defines the grid size. 
         We use a 5 by 5, 6 by 6, 8 by 8, and 16 by 16 grid. \\ 
         \raggedright\textit{MultiRoom-N[X]-S[Y]} & The environment has a series of connected rooms with doors that must be opened in order to get to the next room. 
         The final green goal square is located in the last room. 
         The rooms are all created randomly.
         This environment is challenging to solve using RL alone.
         The value of X defines the number of rooms and X the room size. 
         We use N2-S4, N4-S5, and N6-S10 for our experiments.\\ 
         \raggedright\textit{KeyCorridor-S[X]R[Y]} & The agent has to pick up an object which is behind a locked door. 
         The key is hidden in another room, and the agent has to explore the environment to find it. 
         The key, ball, and agent are placed randomly.
         Due to the exploration required this is a challenging environment.
         The value of X defines the room size and Y number of rows (see \cref{fig:miniGridEnvironments}). 
         We use an S3R1, S3R2, S3R3, S4R3, S5R3, S6R3 grid.\\ 
         \hline
    \end{tabular}
    \caption{Description of the MiniGrid environments used in this thesis.
    All environments impose a penalty for the number of steps taken until reaching the target.}
    \label{tab:miniGridEnvironments}
\end{table}

\subsection{State \& Observations}
MiniGrid supports a variety of observation types.
In our work, we use a fully observable view of the environment. 
This view has a dimension of $\text{N} \times \text{N} \times 3$, where N is the dimension of the grid world.
These values are not pixels, the last channel is encoding the tile at the $(\text{X},\text{Y})$ location.
The tile encoding is a three-dimensional tuple: (OBJECT\_INDEX, COLOR\_INDEX, STATE).
Only doors and agents have a state value other than 0.
The door state 0 represents an open, 1 a closed, and 2 a locked door.
The agents' state indicates the direction of the agent.

We are transferring the policy and value function weights from environment to environment. 
Therefore the input dimension must stay the same for every environment.
We apply padding to have a $25 \times 25$ grid world independent of the selected environment.
As padding values, we use $(1,0,0)$, which is equivalent to a wall.

\clearpage

\subsection{Actions}
The action space in MiniGrid consists of six actions:
Turn left, turn right, move forward, pickup, drop and toggle (open door, interact with objects).

\subsection{Rewards}
\begin{figure}[h!]
    \centering
    \begin{subfigure}{.2\textwidth}
        \centering
        \includegraphics[width=0.99\linewidth]{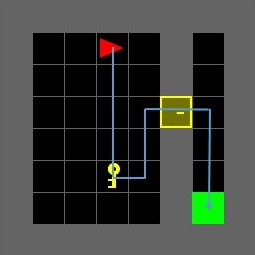}
    \end{subfigure}
    \begin{subfigure}{.2\textwidth}
        \centering
        \includegraphics[width=0.99\linewidth]{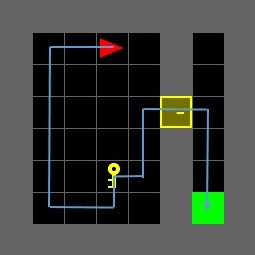}
    \end{subfigure}
    \caption{A blue line illustrates the agent's trajectory in the DoorKey-8x8 environment.
    The maximum number of steps in the DoorKey-8x8 environment is 640.
    If the agent takes the direct path, he takes 18 steps until he reaches the green goal square.
    Therefore the agents reward is $1-0.9*(18/640) = 0.9747$.
    The agent on the right takes 28 steps, leaving him with a reward of $1-0.9*(28/640) = 0.9606$.}
    \label{fig:rewardCalculation}
\end{figure}

The agent receives a reward of 1 for reaching the goal square and 0 otherwise.
A penalty for the number of steps required to reach the target location is imposed.
If the agent takes more steps to reach the target, the reward decreases.
The grid world reward upon success is calculated according to this equation:
\begin{equation}
    r = 1-0.9 * (\text{step\_count} / \text{max\_steps})
\end{equation}
Where step\_count is equal to the number of steps taken to reach the goal square and max\_steps is the maximum number of steps allowed per episode.
In \cref{fig:rewardCalculation} we ilustrated the reward calculation for two different trajectories in the DoorKey-8x8 environment.

\subsection{MDP Statement}
Depending on the type of observation used, the grid world environment does not fulfill the Markov property.
Some observation types only provide a limited field of view.
To ease the problem, we only use a fully observable input for all of our student agents.
Each observation at every time step fully describes the environment's state space, and the following environment state is entirely dependent on the current state.

\section{Grid World Experiments}
\label{sec:gridworldExperiments}

\subsection{Experimental Setup}
\label{sec:gridworldSetup}
\begin{figure}[htb]
    \centering
    \includegraphics[width=0.3\linewidth]{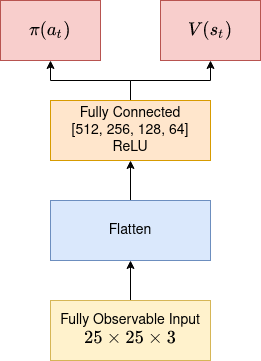}
    \caption{Student neural network architecture for the MiniGrid environments. 
    The input is a $25\times25\times3$ fully observable representation of the environments state.
    After every fully connected layer we use a ReLU activation function.
    The policy head uses a softmax activation function for the action probability distribution.
    The value head does not use an activation function.}
    \label{fig:studentNetwork}
\end{figure}

In this section, we detail the experimental setup applied to all our grid world experiments.
\paragraph{Network architecture} Our student agent uses a fully connected network with ReLU activation functions and separate policy and value function heads.
The network architecture is depicted in \cref{fig:studentNetwork}.

\begin{figure}[htb]
    \centering
    \includegraphics[width=0.9\linewidth]{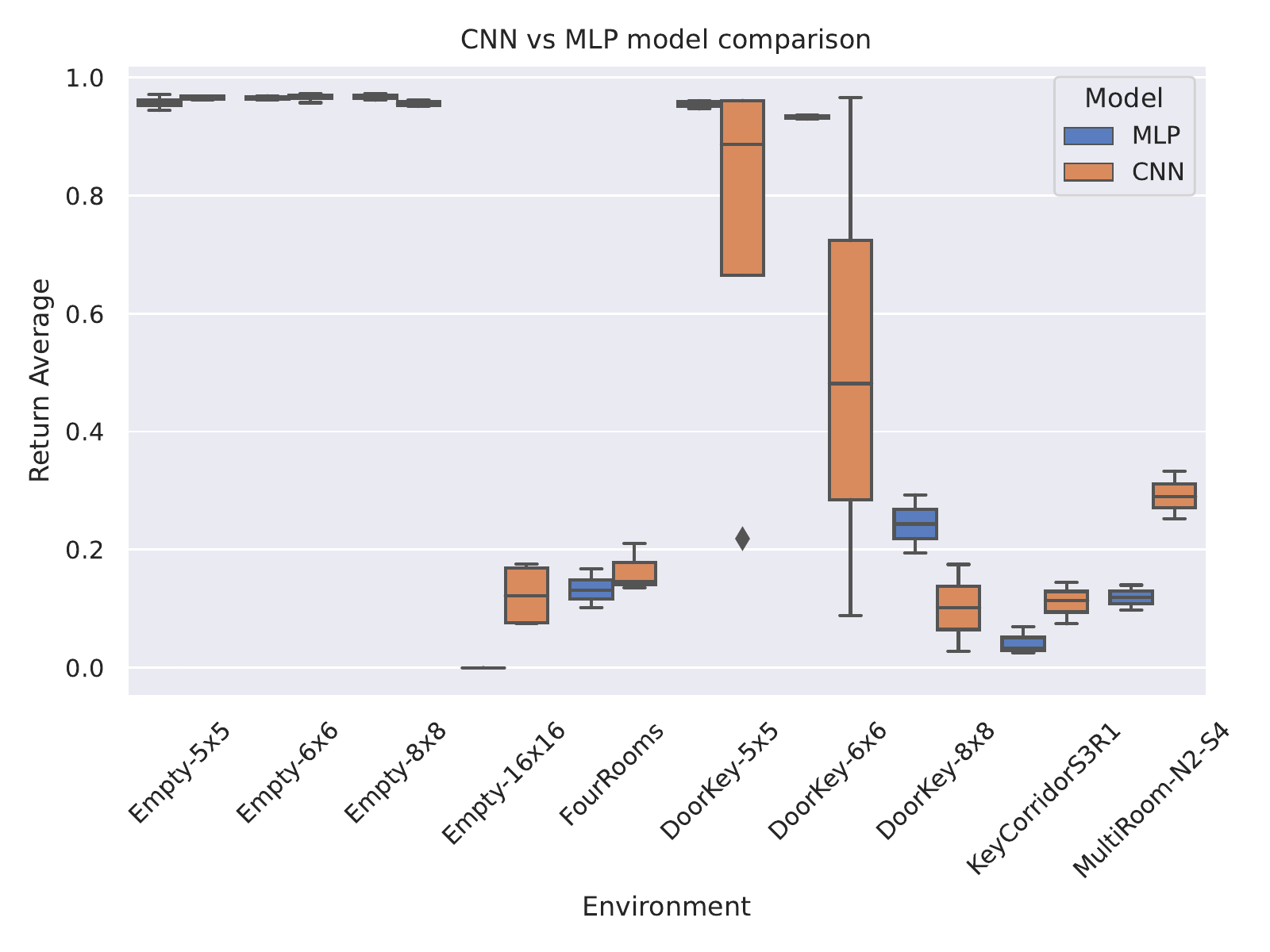}
    \caption{Comparison of the average return over 100 episodes at the end of training between MLP and CNN models.
    The agent is trained with PPO for 10 million steps on a single environment.}
    \label{fig:cnnvsmlp}
\end{figure}

We also evaluated a CNN network architecture.
In the \cref{sec:appendixCNN} we provide details about the CNN architecture.
The $25 \times 25 \times 3$ grid world observation has location-dependent features.
Therefore by intuition we expect that CNN models work better than MLP models.
Flattening the observation into a vector makes it harder to discover patterns and generalize over the state space.
As we can see in \cref{fig:cnnvsmlp}, using a CNN network architecture instead of the MLP architecture improves results on harder environments.
MLP models are less noisy than CNN models, especially in the case of DoorKey experiments.
We value stability over best possible results in our thesis because reward changes in our CMDP setting also lead to a lot of noise, therefore we use MLP models for the rest of our experiments.

\paragraph{No Curriculum Experiments} For all MiniGrid environments, we trained an agent with PPO for 10 million steps and evaluated the agent for 100 episodes at the end of training.
We used these experiments to tune hyperparameters as well as the network architecture.
In \cref{sec:appendixGridworldStudent} we report the hyperparameters.
Additionally, we used the results to select a subset of the MiniGrid environments for our curriculum learning experiments.
We removed the group of MiniGrid environments where our trained agents failed in solving the environment.
In the case of KeyCorridor and MultiRoom, training an agent for the simplest version of those environments succeeded. Therefore we kept all KeyCorridor and MultiRoom environments in our task set $\mathcal{V}$.
All experiments were repeated three times under different random seeds. 
In our results, we report maximum and standard deviations.

\paragraph{Curriculum learning} For both the baseline and the CMDP experiments, we perform $1'000$ teacher steps, where for each teacher step, the student is trained for $10'000$ steps in the selected environment.
Therefore, the student agent is trained for 10 million steps in total.
After each teacher step, the student is evaluated for 100 episodes in each task in $\mathcal{V}$.
We fixed those number of steps to have the same training steps as we used in our no curriculum learning experiments.
We had to balance the number of teacher updates and the number of student updates per CMDP step.
Using fewer teacher steps might not be sufficient for the teacher to learn how to perform task sequencing but would allow our student more time to converge in the selected environment.
Using more teacher steps is beneficial for the teacher, but on the other hand, the student has less time to learn the selected task.
This would provide the teacher with a noisy reward signal.
We selected the teacher and student steps because they seem reasonable. 
In future work, the influence of those values should be evaluated.

The network architecture for the teacher agent is described in \cref{sec:teacherArchitecture}.
We evaluated the MLP and LSTM network architecture in \cref{sec:appendixMLPvsLSTM}.
The MLP architecture is superior to the LSTM architecture.
The hyperparameters for the baselines as well as the CMDP teachers are described in \cref{sec:appendixGridworldStudent} and \cref{sec:appendixTeacherPPO}.

All experiments were repeated three times under different random seeds.
In our results, we report maximum and standard deviations.

\subsection{Results Overview}
\label{sec:gridworldResults}
\Cref{tab:gridworldBestResults} shows the best agent's performance on each MiniGrid environment used in this thesis.
For each environment, we report the average return of 100 evaluation episodes.
The total mean return is defined as the sum over all average returns.
The percentage of environments solved is defined as the number of environments with an average reward greater than zero divided by all environments.
Overall, using no curriculum results in a higher average return on 8 out of the 19 tested tasks.
In contrast, curriculum learning agents outperform standard RL agents in more complex tasks such as Empty-16x16, DoorKey16x16, and KeyCorridor-S6R3.
Comparing the best CMDP agent to the best baseline agents we see, the CMDP agent outperforms the baselines regarding the total mean return and percentage of solved environments.

\begin{table}[!htb]
    \centering
    \begin{adjustbox}{width=\textwidth}
    \begin{tabular}{ccccccc}
        \hline
        \textbf{Environment / Metric} & \textbf{None} & \textbf{Uniform} & \textbf{LP} & \textbf{Thompson} & \textbf{Window} & \textbf{CMDP}\\
        \hline
        \hline
        Total mean return   & - & 1.75 & 2.71 & 2.83 & 2.27 & \textbf{4.44} \\
        \% environments solved & 50\% & 44\% & 39\% & 50\% & 33\% & \textbf{55\%} \\
        Empty-5x5        & \textbf{0.96} & 0.83 & 0.75 & 0.57 & 0.37 & 0.93 \\
        Empty-6x6        & \textbf{0.97} & 0.51 & 0.83 & 0.51 & 0.32 & 0.9 \\
        Empty-8x8        & \textbf{0.96} & 0.0  & 0.0  & 0.62 & 0.0  & 0.93 \\
        Empty-16x16      & 0.0           & 0.0  & \textbf{0.93} & 0.92 & 0.67 & 0.82 \\
        FourRooms        & \textbf{0.17} & 0.07 & 0.09 & 0.09 & 0.0  & 0.09 \\
        DoorKey-5x5      & \textbf{0.96} & 0.17 & 0.02 & 0.02 & 0.0  & 0.13 \\
        DoorKey-6x6      & \textbf{0.94} & 0.04 & 0.0  & 0.04 & 0.81 & 0.13 \\
        DoorKey-8x8      & \textbf{0.29} & 0.07 & 0.0  & 0.05 & 0.0  & 0.14 \\
        DoorKey-16x16    & 0.0  & 0.03 & 0.04 & 0.0  & 0.07 & \textbf{0.17} \\
        MultiRoom-N2-S4  & 0.14 & 0.02 & 0.04 & 0.02 & 0.03 & \textbf{0.15} \\
        MultiRoom-N4-S5  & 0.0  & 0.0  & 0.0  & 0.0  & 0.0  & 0.0  \\
        MultiRoom-N6-S10 & 0.0  & 0.0  & 0.0  & 0.0  & 0.0  & 0.0  \\
        KeyCorridor-S3R1 & \textbf{0.07}& 0.0  & 0.0  & 0.0  & 0.0  & 0.05 \\
        KeyCorridor-S3R2 & 0.0  & 0.0  & 0.0  & 0.0  & 0.0  & 0.0  \\
        KeyCorridor-S3R3 & 0.0  & 0.0  & 0.0  & 0.0  & 0.0  & 0.0  \\
        KeyCorridor-S4R3 & 0.0  & 0.0  & 0.0  & 0.0  & 0.0  & 0.0  \\
        KeyCorridor-S5R3 & 0.0  & 0.0  & 0.0  & 0.0  & 0.0  & 0.0  \\
        KeyCorridor-S6R3 & 0.0  & 0.0  & 0.0  & 0.0  & 0.0  & \textbf{0.01} \\
    \end{tabular}
    \end{adjustbox}
    \caption{Highest average return over 100 episodes for each teacher type at the end of 1000 teacher steps.
    The None results are the average return over 100 episodes obtained when training only the environment of that row.}
    \label{tab:gridworldBestResults}
\end{table}

In the curriculum learning approach, the agent spends less training time in a single environment than with no curriculum.
In easy environments, using no curriculum learning and training on a single task yields better results than the curriculum learning experiments.
Training for 10 Million steps in a single environment is better than learning in a curriculum without specifically targeting those easy environments. 
In the CMDP setting using the \textit{source task reward}, the teacher is encouraged to maximize the student's average return over all environments and not a single environment, as is the case when using no curriculum.

The following sections evaluate different teacher reward signals, transfer methods, observation types, hyperparameters as well as the sample efficiency and generality of trained agents.

\clearpage

\subsection{Transfer Method}
In this section, we evaluate the effect of the transfer methods introduced in \cref{sec:methodsTransferMethod}.
We evaluate policy transfer, reward shaping transfer, and both transfers combined.
All teacher agents use the \textit{source task reward}, as defined in \cref{sec:methodRewardFunction}.
The results for all seven teacher observation types are reported in \cref{tab:gridworldTransferMethod} and the four baselines in \cref{tab:gridworldTransferBaseline}.

\begin{figure}[htb]
    \centering
    \includegraphics[width=0.9\linewidth]{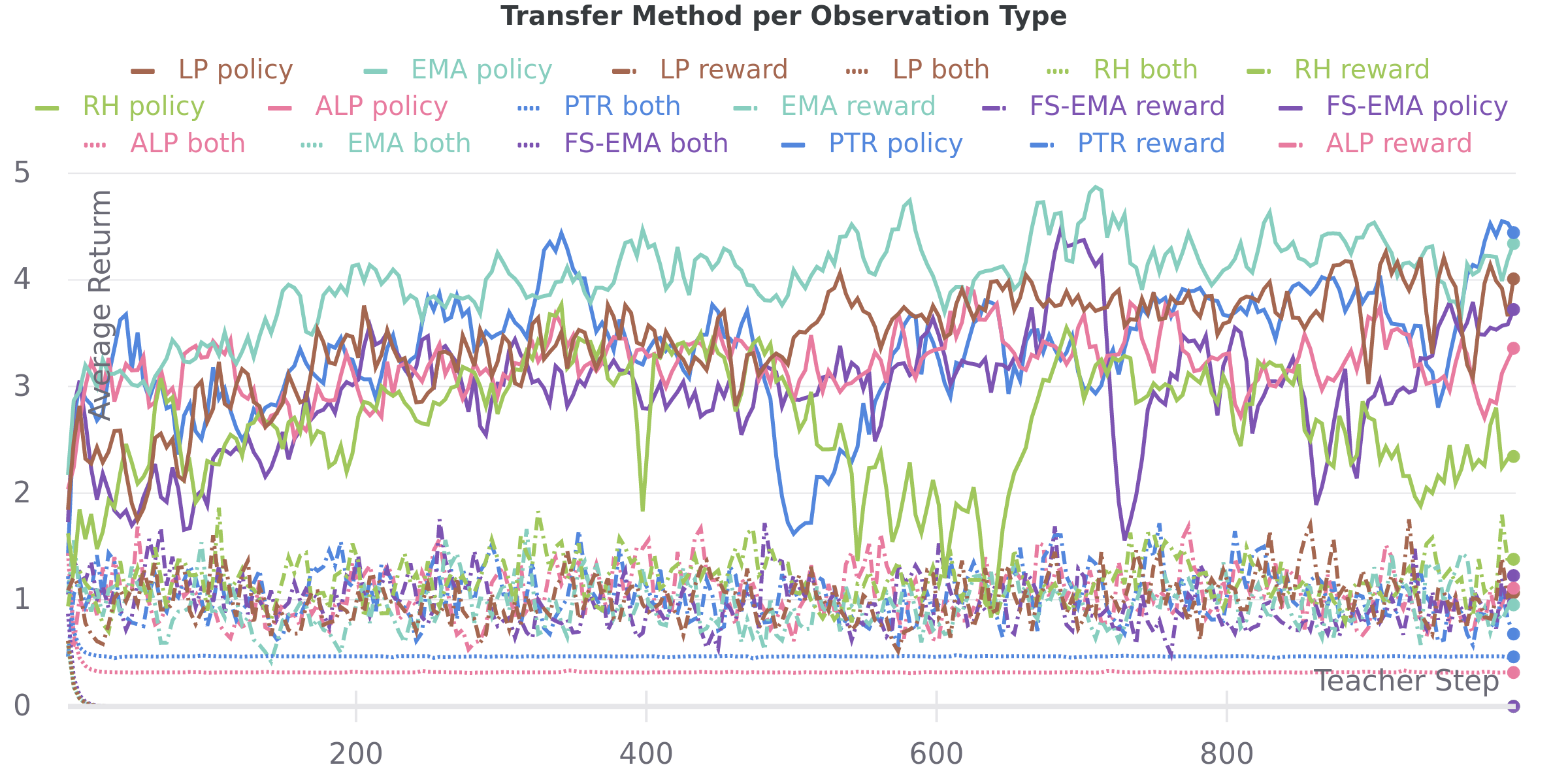}
    \caption{Learning curves for different teacher observations and transfer methods.
    After each teacher step the sum of the students average return over 100 episodes for each environment in $\mathcal{V}$ is plotted.
    We plot the learning curve with the highest total reward at the end of training for every configuration.}
    \label{fig:transferMethodLC}
\end{figure}

\Cref{fig:transferMethodLC} shows the learning curves of the reinforcement learning agents.
All agents with policy transfer surpass agents with reward shaping or the combined knowledge transfer.
The learning curve is noisy in policy transfer due to the environment changes, but the performance is steadily increasing.
The performance of agents with policy transfer did not converge after 1000 teacher steps.
For both reward transfer and the combined transfer, the performance converged during the first 50 to 100 CMDP steps and failed to improve from then on.
The agent's performance with reward transfer scatters around a total mean return of 1 during the whole training.

Approximating the entire student state space by encoding the students' weights with PCA as described in \cref{sec:teacherObservation} fails.
The results of the experiments with PCA observation space are very similar to the uniform baseline results.
We, therefore, discard this method for the rest of our experiments.

The best results are obtained using the PTR, EMA, and LP observations with a policy transfer.
Using the fast-slow EMA as described in \cref{sec:teacherObservation} is not superior to standard EMA.
In \cref{sec:alphaExperiments} we evaluate the best $\alpha$ value for the EMA observation.
Using the reward history (RH) is significantly worse than the other observation types when using policy transfer.
RH resulted in the highest total mean return when using reward transfer but suffers from a high standard deviation.
In general, the results with policy transfer are more stable compared to the results with reward transfer.

\begin{table}[!h]
    \centering
    \begin{adjustbox}{width=\textwidth}
        \begin{tabular}{clccccccc}
            \hline
             & \textbf{Environment / Metric} & \textbf{RH} & \textbf{PTR} & \textbf{LP} & \textbf{ALP} & \textbf{EMA} & \textbf{FS-EMA} & \textbf{PCA} \\
            \hline
            \hline
            \multirow{7}{*}{\begin{sideways}\textbf{Policy}\end{sideways}} & Total mean return      & $2.34 \pm 0.24$ & $\mathbf{4.44 \pm 0.33}$ & $4.17 \pm 0.33$ & $3.35 \pm 0.33$ & $4.35 \pm 0.42$ & $3.72 \pm 0.22$ & $1.64 \pm 0.24$ \\
            & \% environments solved  & 55\% & 55\% & 55\% & 55\% & 55\% &  \textbf{61\%} & 44\%\\
            & Empty-16x16      & $0.58 \pm 0.27$ & $0.07 \pm 0.03$ & $0.9 \pm 0.27$  & $0.08 \pm 0.03$ & $0.82  \pm 0.34$  & $\mathbf{0.91 \pm 0.38}$ & $0.0 \pm 0.0$ \\
            & FourRooms        & $0.11 \pm 0.03$ & $\mathbf{0.13 \pm 0.03}$ & $0.11 \pm 0.02$ & $0.1  \pm 0.04$ & $0.08 \pm 0.01$ & $0.06 \pm 0.03$  & $0.08 \pm 0.02$ \\
            & DoorKey-16x16    & $0.05 \pm 0.01$ & $0.04 \pm 0.02$ & $0.07 \pm 0.03$ & $0.13 \pm 0.05$ & $\mathbf{0.17 \pm 0.06}$ & $0.06 \pm 0.03$  & $0.03 \pm 0.01$\\
            & MultiRoom-N2-S4  & $0.1  \pm 0.02$ & $0.06 \pm 0.02$ & $0.09 \pm 0.01$ & $0.06 \pm 0.03$ & $\mathbf{0.15 \pm 0.05}$ & $0.09 \pm 0.04$  & $0.0 \pm 0.0$\\
            & KeyCorridor-S3R3 & $0.0  \pm 0.0 $ & $0.0  \pm 0.0$  & $0.0 \pm 0.0$   & $0.0  \pm 0.0 $ & $0.0  \pm 0.0$  & $0.0  \pm 0.0$  & $0.0 \pm 0.0$\\
            \hline
            \multirow{7}{*}{\begin{sideways}\textbf{Reward}\end{sideways}} & Total mean return & $\mathbf{1.38 \pm 0.85}$ & $0.68 \pm 0.3$ & $1.01 \pm 0.32$ & $1.11 \pm 0.63$ & $0.95 \pm 0.1$ & $1.22 \pm 0.58$  & $0.63 \pm 0.21$\\
            & \% environments solved    & 50\% & 44\% & \textbf{61\%} & 50\% & 11\% & 50\% & 34\%\\
            & Empty-16x16      & $\mathbf{0.56 \pm 0.28}$ & $0.0 \pm 0.0 $    & $0.46 \pm 0.23$ & $0.0 \pm 0.0$   & $0.0 \pm 0.0$ & $0.0 \pm 0.0$  & $0.0 \pm 0.0$\\
            & FourRooms        & $0.04 \pm 0.002$& $0.03 \pm 0.01$   & $\mathbf{0.05 \pm 0.03}$ & $0.03 \pm 0.01$ & $0.0 \pm 0.0$ & $0.03 \pm 0.01$  & $0.0 \pm 0.0$\\
            & DoorKey-16x16    & $0.0  \pm 0.0 $ & $0.0 \pm 0.0 $    & $\mathbf{0.03 \pm 0.01}$ & $0.02 \pm 0.01$ & $0.0 \pm 0.0$ & $0.0 \pm 0.0$  & $0.0 \pm 0.0$\\
            & MultiRoom-N2-S4  & $0.0  \pm 0.0 $ & $0.0 \pm 0.0 $    & $\mathbf{0.06 \pm 0.03}$ & $0.0 \pm 0.0$   & $0.0 \pm 0.0$ & $0.02 \pm 0.001$  & $0.0 \pm 0.0$\\
            & KeyCorridor-S3R3 & $0.003\pm0.001$ & $0.003 \pm 0.002$ & $0.0  \pm 0.0 $ & $0.0 \pm 0.0$   & $0.0 \pm 0.0$ & $\mathbf{0.004 \pm 0.001}$  & $0.0 \pm 0.0$\\
            \hline
            \multirow{7}{*}{\begin{sideways}\textbf{Both}\end{sideways}} & Total mean return & $0.0 \pm 0.0$ & $\mathbf{0.47 \pm 0.24}$ & $0.0 \pm 0.0$ & $0.32 \pm 0.16$ & $0.0 \pm 0.0$ & $0.0 \pm 0.0$  & $0.0 \pm 0.0$\\
            & \% environments solved   & 0\% & \textbf{17\%} & 0\% & \textbf{17\%} & 0\% & 0\% & 0\%\\
            & Empty-16x16      & $0.0 \pm 0.0$ & $0.0 \pm 0.0$ & $0.0 \pm 0.0$ & $0.0 \pm 0.0$  & $0.0 \pm 0.0$ & $0.0 \pm 0.0$  & $0.0 \pm 0.0$\\
            & FourRooms        & $0.0 \pm 0.0$ & $0.0 \pm 0.0$ & $0.0 \pm 0.0$ & $\mathbf{0.02 \pm 0.01}$& $0.0 \pm 0.0$ & $0.0 \pm 0.0$  & $0.0 \pm 0.0$\\
            & DoorKey-16x16    & $0.0 \pm 0.0$ & $0.0 \pm 0.0$ & $0.0 \pm 0.0$ & $0.0 \pm 0.0$  & $0.0 \pm 0.0$ & $0.0 \pm 0.0$  & $0.0 \pm 0.0$\\
            & MultiRoom-N2-S4  & $0.0 \pm 0.0$ & $0.0 \pm 0.0$ & $0.0 \pm 0.0$ & $0.0 \pm 0.0$  & $0.0 \pm 0.0$ & $0.0 \pm 0.0$  & $0.0 \pm 0.0$\\
            & KeyCorridor-S3R3 & $0.0 \pm 0.0$ & $0.0 \pm 0.0$ & $0.0 \pm 0.0$ & $0.0 \pm 0.0$  & $0.0 \pm 0.0$ & $0.0 \pm 0.0$  & $0.0 \pm 0.0$\\
        \end{tabular}
    \end{adjustbox}
    \caption{We compare the different knowledge transfer methods for each teacher observation type.
    Each configuration is run with three different random seeds.
    The maximum return over 100 episodes at the end of 1'000 teacher steps is reported.
    Additionally, we provide the standard deviation between the three runs.}
    \label{tab:gridworldTransferMethod}
\end{table}

\begin{figure}[!h]
    \centering
    \begin{subfigure}{.4\textwidth}
        \centering
        \includegraphics[width=.99\linewidth]{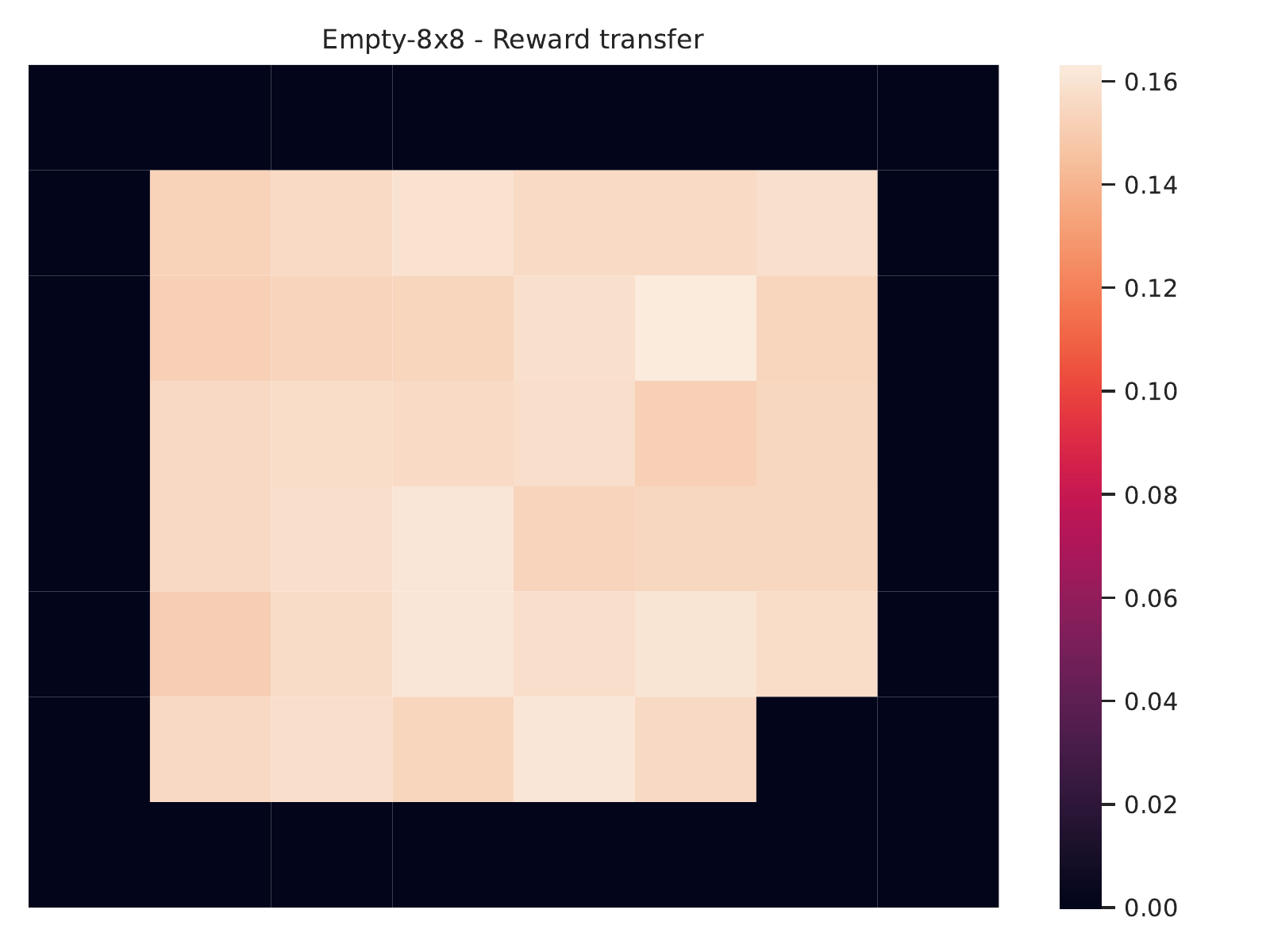}
    \end{subfigure}
    \begin{subfigure}{.4\textwidth}
        \centering
        \includegraphics[width=.99\linewidth]{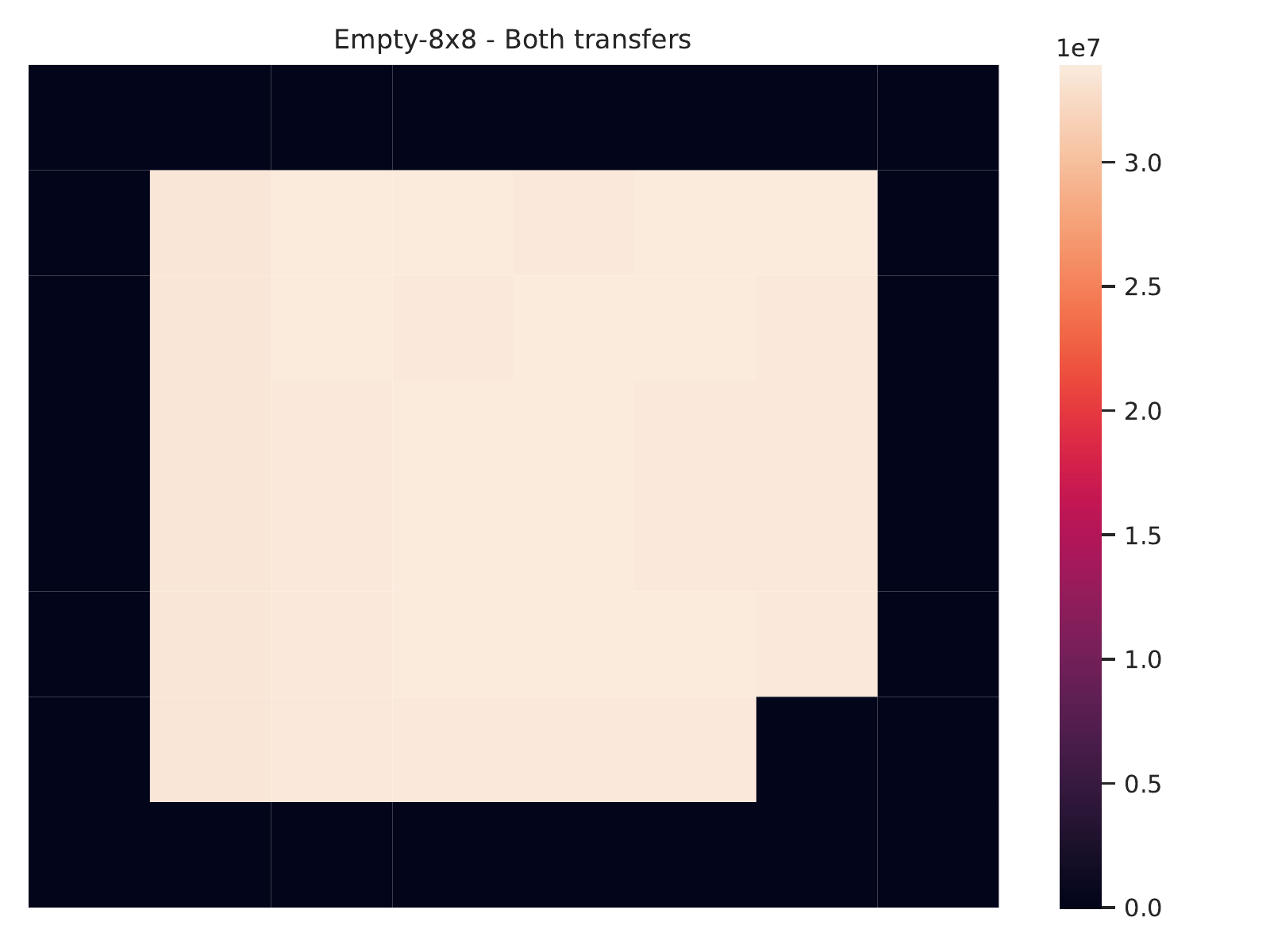}
    \end{subfigure}
    \begin{subfigure}{.4\textwidth}
        \centering
        \includegraphics[width=.99\linewidth]{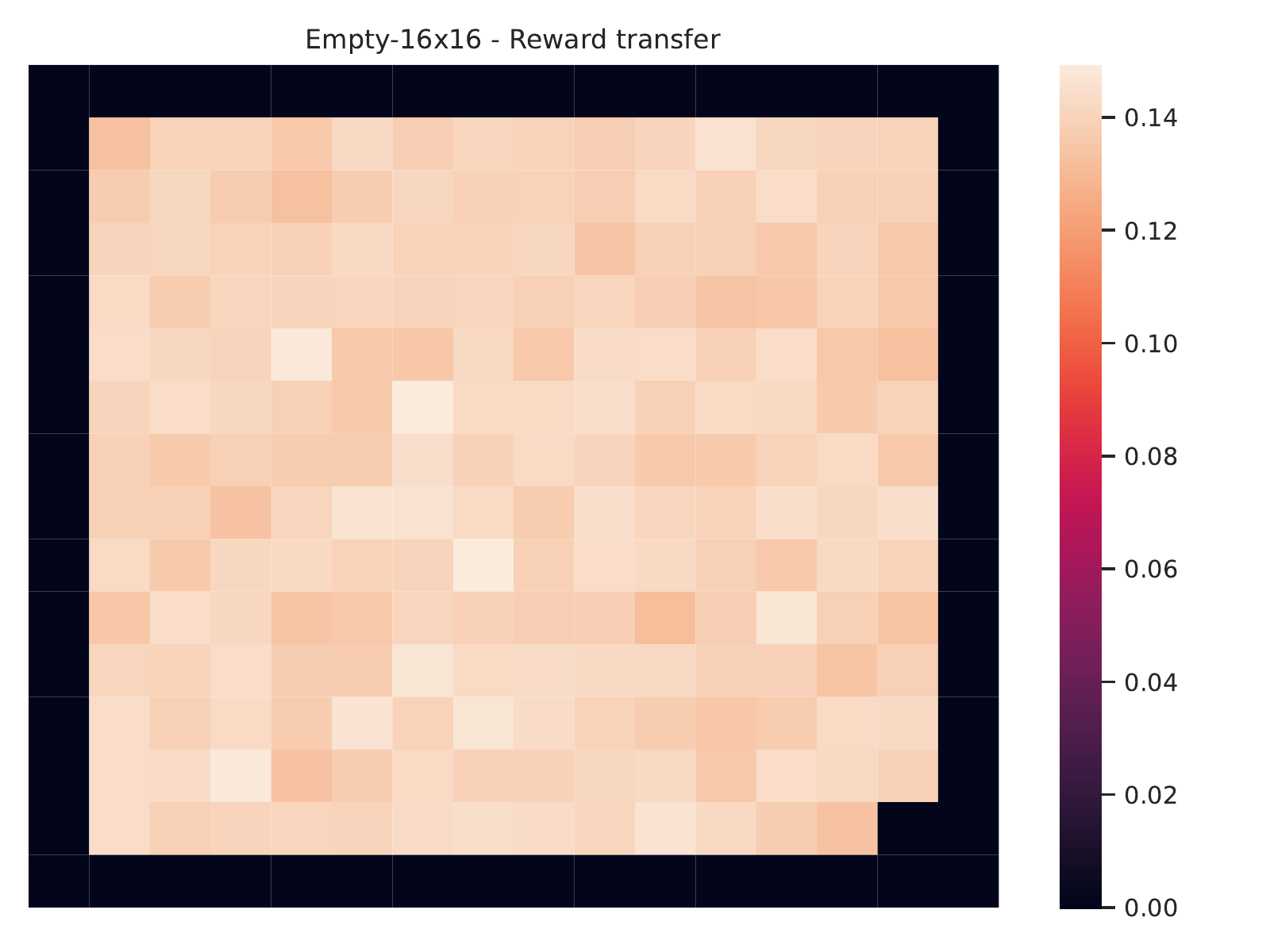}
    \end{subfigure}
    \begin{subfigure}{.4\textwidth}
        \centering
        \includegraphics[width=.99\linewidth]{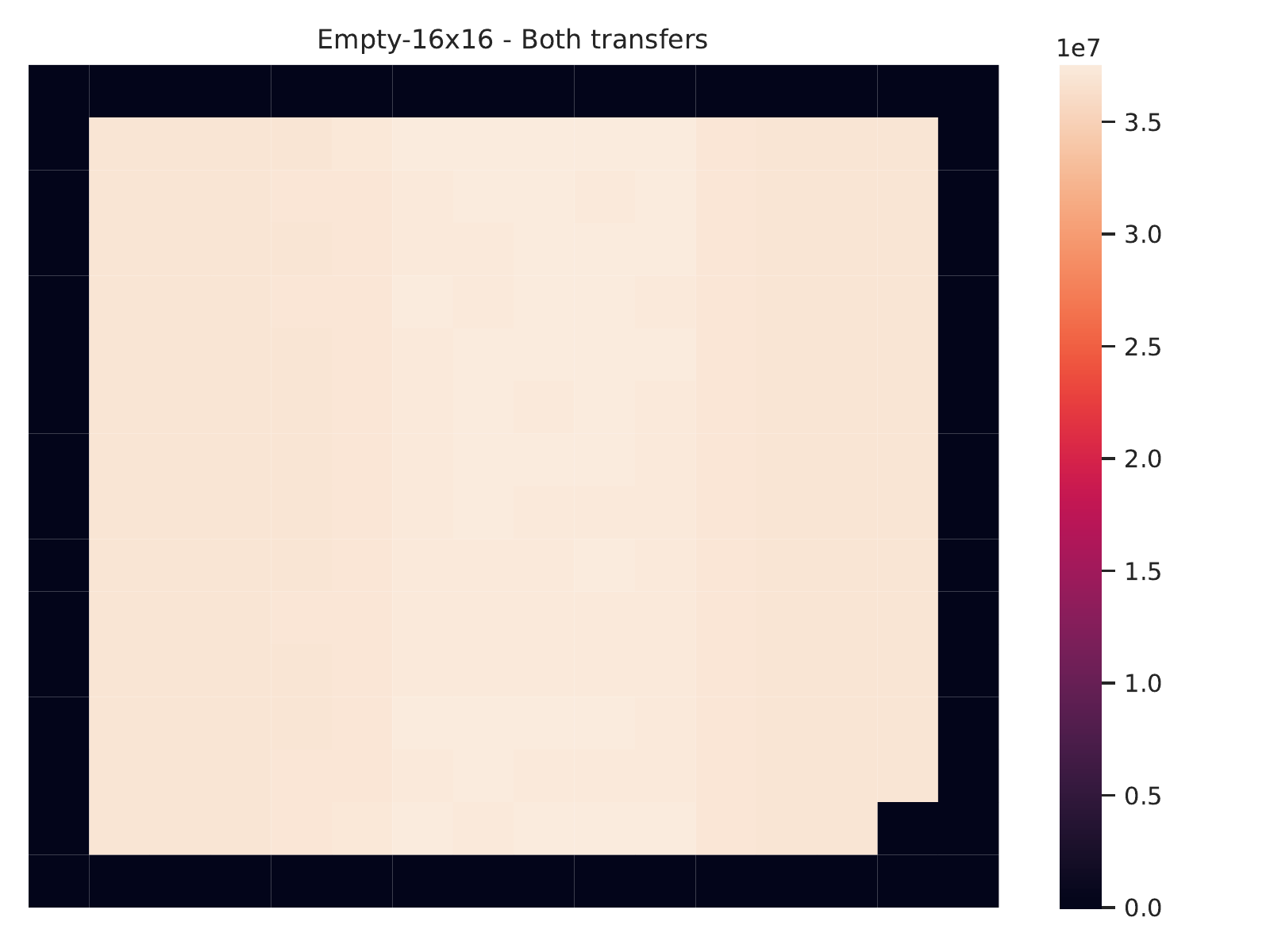}
    \end{subfigure}
    \caption{We visualize the added reward, when using reward transfer (left column) and combined transfer (right column).
    For every state in the Empty-8x8 (top row) and Empty-16x16 (bottom row) grid world environment we plot the added reward.
    A light color encodes a high added reward, black equals to zero added reward.
    Dark tiles at the end of the grid are walls, the black tile in the bottom right corner is the green goal square.}
    \label{fig:valuaFunctionViz}
\end{figure}

Agents with the combined knowledge transfer are failing to learn at all.
In \cref{fig:transferMethodLC} we see that the average return drops from around 0.7 to 0 during the first ten teacher steps and fails to recover from there.
We visualized the synthetic reward signal added by the reward transfer in \cref{fig:valuaFunctionViz} for the Empty-8x8 and Empty-16x16 environment.
Each square in the image equals a square in the grid world environment.
The black squares are wall objects, and the element in the bottom right is the green goal square.
High state values estimated by the value function obtained in experiments with only reward transfer (images on the left) are scattered randomly around the grid.
The estimated state value is between 0.12 and 0.16.
In the value function with a combined knowledge transfer, the state values are almost identical for all states and have an estimated value of 10 Million.
We believe that the agent starts to optimize and estimate its previous value function and this leads to an ever increasing value estimate.
This extremely high additional reward overshadows the actual reward signal of the environment, making it impossible for the agent to learn how to solve the environment successfully.

\begin{figure}[htb]
    \centering
    \begin{subfigure}{0.49\textwidth}
        \includegraphics[width=.99\linewidth]{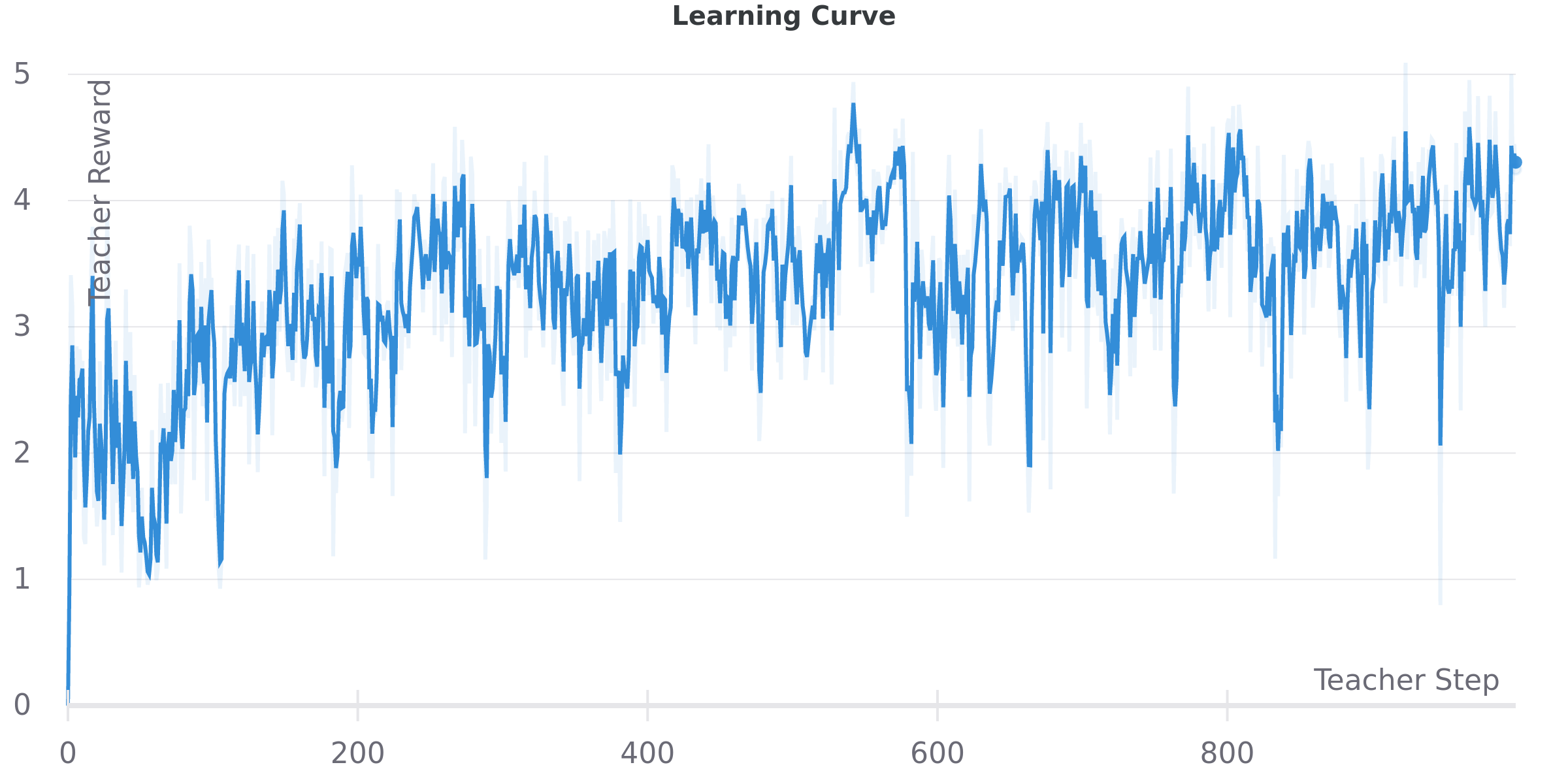}
    \end{subfigure}
    \begin{subfigure}{0.49\textwidth}
        \includegraphics[width=.99\linewidth]{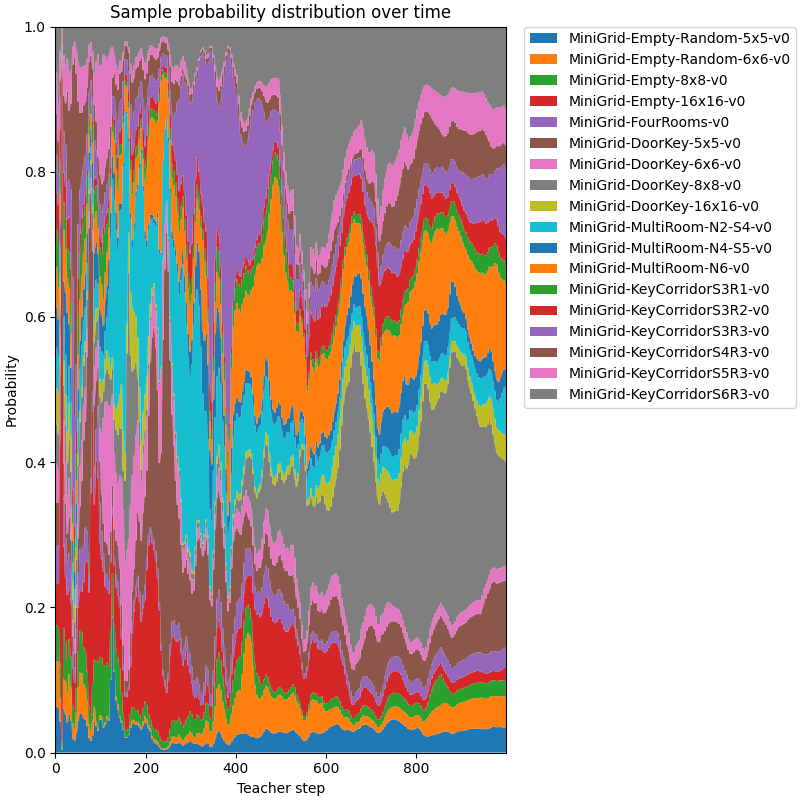}
    \end{subfigure}
    \caption{We visualize the sample probability distribution and learning curve of one teacher agent over its training time.
    The colors in the legends are ordered upside down to the order in the plot.}
    \label{fig:sampleProbability}
\end{figure}

In \cref{fig:sampleProbability} we plot the teacher's sampling probability distribution over the teacher timesteps next to the teacher's learning curve.
During the first 100 episodes, the teacher samples tasks similar to a uniform distribution.
The teacher favors the family of Empty grid world environments during the first 200 episodes.
From then on, the teacher selects more challenging environments such as KeyCorridor and DoorKey environments.
After roughly 400 teacher steps, the sample distribution does not show the noisy changes in the sample probabilities as at the beginning of the training.
We interpret this as the teacher getting more confident in performing task sequencing.
The teacher's learning curve is noisy but steadily increasing.
There are some drastic drops in performance, probably due to selecting too challenging environments.
The student always manages to recover after those harmful teacher actions.

\Cref{tab:gridworldTransferBaseline} shows the same properties as described above.
The Thompson sampling experiment with reward transfer is interesting.
This is the only experiment where the knowledge transfer with a reward signal surpassed agents with a policy transfer.
Therefore, this approach to transfer learning is feasible but very noisy, which is also indicated by the high standard deviation for the Thompson sampling experiments with reward transfer.
While uniformly sampling environments is a strong baseline, it still shows the worst performance across the baselines.
In general, sampling environments with the highest learning progress estimate yields the strongest baseline results.

\begin{table}[!htb]
    \centering
    \begin{adjustbox}{width=\textwidth}
    \begin{tabular}{clcccc}
        \hline
         & \textbf{Environment / Metric} & \textbf{Uniform} & \textbf{LP} & \textbf{Thompson} & \textbf{Window} \\
        \hline
        \hline
        \multirow{7}{*}{\begin{sideways}\textbf{Policy}\end{sideways}} & Total mean return      & $1.75 \pm 0.24$ & $\mathbf{2.71 \pm 0.33}$ & $1.94 \pm 0.21$  & $2.28 \pm 0.32$ \\
        & \% environments solved  & \textbf{44\%} & 39\% & 22\% & 34\% \\
        & Empty-16x16      & $0.0 \pm 0.0$  & $\mathbf{0.92 \pm 0.23}$  & $0.0 \pm 0.0$  & $0.67 \pm 0.14$ \\
        & FourRooms        & $0.07 \pm 0.02$ & $\mathbf{0.09 \pm 0.02}$ & $0.0 \pm 0.0$  & $0.0 \pm 0.0$ \\
        & DoorKey-16x16    & $0.03 \pm 0.01$ & $0.04 \pm 0.01$  & $0.0 \pm 0.1$ & $\mathbf{0.07 \pm 0.03}$ \\
        & MultiRoom-N2-S4  & $0.02 \pm 0.01$ & $\mathbf{0.04 \pm 0.01}$ & $0.03 \pm 0.01$ & $0.03 \pm 0.01$ \\
        & KeyCorridor-S3R3 & $0.0 \pm 0.0$  & $0.0 \pm 0.0$  & $0.0 \pm 0.0$  & $0.0 \pm 0.0$ \\
        \hline
        \multirow{7}{*}{\begin{sideways}\textbf{Reward}\end{sideways}} & Total mean return & $0.8 \pm 0.34$ & $0.51 \pm 0.23$ & $\mathbf{2.83 \pm 1.1}$ & $0.45 \pm 0.12$ \\
        & \% environments solved    & 44\% & 11\% & \textbf{50\%} & 22\% \\
        & Empty-16x16      & $0.0 \pm 0.0$ & $0.0 \pm 0.0$ & $\mathbf{0.92 \pm 0.36}$ & $0.0 \pm 0.0$ \\
        & FourRooms        & $0.03 \pm 0.01$ & $0.0 \pm 0.0$ & $\mathbf{0.09 \pm 0.05}$ & $0.0 \pm 0.0$ \\
        & DoorKey-16x16    & $0.0 \pm 0.0$ & $0.0 \pm 0.0$ & $0.0 \pm 0.0$  & $0.0 \pm 0.0$ \\
        & MultiRoom-N2-S4  & $\mathbf{0.06 \pm 0.02}$ & $0.0 \pm 0.0$ & $0.02 \pm 0.01$ & $0.0 \pm 0.0$ \\
        & KeyCorridor-S3R3 & $\mathbf{0.01 \pm 0.01}$ & $0.0 \pm 0.0$ & $0.0 \pm 0.0$  & $0.0 \pm 0.0$ \\
        \hline
        \multirow{7}{*}{\begin{sideways}\textbf{Both}\end{sideways}} & Total mean return & $0.47 \pm 0.21$ & $0.0 \pm 0.0$ & $0.0 \pm 0.0$ & $\mathbf{0.52 \pm 0.22}$ \\
        & \% environments solved    & \textbf{17\%} & 0\% & 0\% & \textbf{17\%} \\
        & Empty-16x16      & $0.0 \pm 0.0$  & $0.0 \pm 0.0$ & $0.0 \pm 0.0$ & $0.0 \pm 0.0$ \\
        & FourRooms        & $\mathbf{0.05 \pm 0.01}$ & $0.0 \pm 0.0$ & $0.0 \pm 0.0$ & $0.02 \pm 0.01$ \\
        & DoorKey-16x16    & $0.0 \pm 0.0$  & $0.0 \pm 0.0$ & $0.0 \pm 0.0$ & $0.0 \pm 0.0$ \\
        & MultiRoom-N2-S4  & $0.0 \pm 0.0$  & $0.0 \pm 0.0$ & $0.0 \pm 0.0$ & $0.0 \pm 0.0$ \\
        & KeyCorridor-S3R3 & $0.0 \pm 0.0$  & $0.0 \pm 0.0$ & $0.0 \pm 0.0$ & $0.0 \pm 0.0$ \\
    \end{tabular}
    \end{adjustbox}
    \caption{We compare the different knowledge transfer methods for each baseline.
    Each configuration is run with three different random seeds.
    The maximum the average students return of 100 episodes at the end of 1'000 teacher steps is reported.
    Additionally, we provide the standard deviation between the three runs.}
    \label{tab:gridworldTransferBaseline}
\end{table}

\clearpage

\subsection{Teacher Reward Signal}
In this section we evaluate the \textit{source task reward} and the \textit{target reward}.
Both reward signals are defined in \cref{sec:methodRewardFunction}.
We use the KeyCorridor-S3R3 environment as the target task.
The agents use a policy knowledge transfer between tasks.
In \cref{tab:gridworldTeacherReward} we report experiments for both reward signals with six observation types.

For all observation types, the reported results are better when using the \textit{source task reward} except for the RH experiments.
The reported total mean return has a lower standard deviation over three different random seeds when using the \textit{source task reward} instead of the \textit{target reward}.

\begin{table}[!htb]
    \centering
    \begin{adjustbox}{width=\textwidth}
        \begin{tabular}{clcccccc}
            \hline
             & \textbf{Environment / Metric} & \textbf{RH} & \textbf{PTR} & \textbf{LP} & \textbf{ALP} & \textbf{EMA} & \textbf{FS-EMA} \\
            \hline
            \hline
            \multirow{7}{*}{\begin{sideways}\textbf{Total Eval. Reward}\end{sideways}} & Total mean return      & $2.34 \pm 0.24$ & $\mathbf{4.44 \pm 0.33}$ & $4.17 \pm 0.33$ & $3.35 \pm 0.33$ & $4.35 \pm 0.42$ & $3.72 \pm 0.22$\\
            & \% environments solved  & 55\% & 55\% & 55\% & 55\% & 55\% &  \textbf{61\%}\\
            & Empty-16x16      & $0.58 \pm 0.27$ & $0.07 \pm 0.03$ & $0.9 \pm 0.27$  & $0.08 \pm 0.03$ & $0.82  \pm 0.34$  & $\mathbf{0.91 \pm 0.38}$ \\
            & FourRooms        & $0.11 \pm 0.03$ & $\mathbf{0.13 \pm 0.03}$ & $0.11 \pm 0.02$ & $0.1  \pm 0.04$ & $0.08 \pm 0.01$ & $0.06 \pm 0.03$ \\
            & DoorKey-16x16    & $0.05 \pm 0.01$ & $0.04 \pm 0.02$ & $0.07 \pm 0.03$ & $0.13 \pm 0.05$ & $\mathbf{0.17 \pm 0.06}$ & $0.06 \pm 0.03$ \\
            & MultiRoom-N2-S4  & $0.1  \pm 0.02$ & $0.06 \pm 0.02$ & $0.09 \pm 0.01$ & $0.06 \pm 0.03$ & $\mathbf{0.15 \pm 0.05}$ & $0.09 \pm 0.04$ \\
            & KeyCorridor-S3R3 & $0.0  \pm 0.0 $ & $0.0  \pm 0.0$  & $0.0 \pm 0.0$   & $0.0  \pm 0.0 $ & $0.0  \pm 0.0$  & $0.0  \pm 0.0$ \\
            \hline
            \multirow{7}{*}{\begin{sideways}\textbf{Target Reward}\end{sideways}} & Total mean return & $\mathbf{3.6 \pm 0.49}$ & $\mathbf{3.6 \pm 0.36}$ & $3.38 \pm 0.62$ & $2.35 \pm 0.66$ & $3.41 \pm 0.51$ & $3.38 \pm 0.42$ \\
            & \% environments solved    & 55\% & 55\% & 55\% & 55\% & 55\% & \textbf{61\%} \\
            & Empty-16x16      & $0.58 \pm 0.27$ & $0.07 \pm 0.03$ & $0.24 \pm 0.08$ & $0.5  \pm 0.1$  & $\mathbf{0.72 \pm 0.21}$ & $0.63 \pm 0.08$ \\
            & FourRooms        & $0.11 \pm 0.03$ & $\mathbf{0.13 \pm 0.03}$ & $0.11 \pm 0.04$ & $0.1  \pm 0.04$ & $0.05 \pm 0.01$ & $0.06 \pm 0.02$ \\
            & DoorKey-16x16    & $0.05 \pm 0.004$& $0.02 \pm 0.01$ & $0.0  \pm 0.0 $ & $0.08 \pm 0.02$ & $\mathbf{0.09 \pm 0.03}$ & $0.04 \pm 0.01$ \\
            & MultiRoom-N2-S4  & $0.07 \pm 0.01$ & $\mathbf{0.11 \pm 0.04}$ & $0.01 \pm 0.003$& $0.06 \pm 0.03$ & $0.02 \pm 0.01$ & $0.02 \pm 0.01$ \\
            & KeyCorridor-S3R3 & $0.0 \pm 0.0$   & $0.0 \pm 0.0$   & $0.0  \pm 0.0$  & $0.0  \pm 0.0 $ & $0.0  \pm 0.0$  & $0.0  \pm 0.0 $ \\
        \end{tabular}
    \end{adjustbox}
    \caption{We compare the different teacher reward signals for each teacher observation type.
    Each configuration is run with three different random seeds.
    The maximum return over 100 episodes at the end of 1'000 teacher steps is reported.
    Additionally, we provide the standard deviation between the three runs.}
    \label{tab:gridworldTeacherReward}
\end{table}

\begin{figure}[htb]
    \centering
    \includegraphics[width=0.7\linewidth]{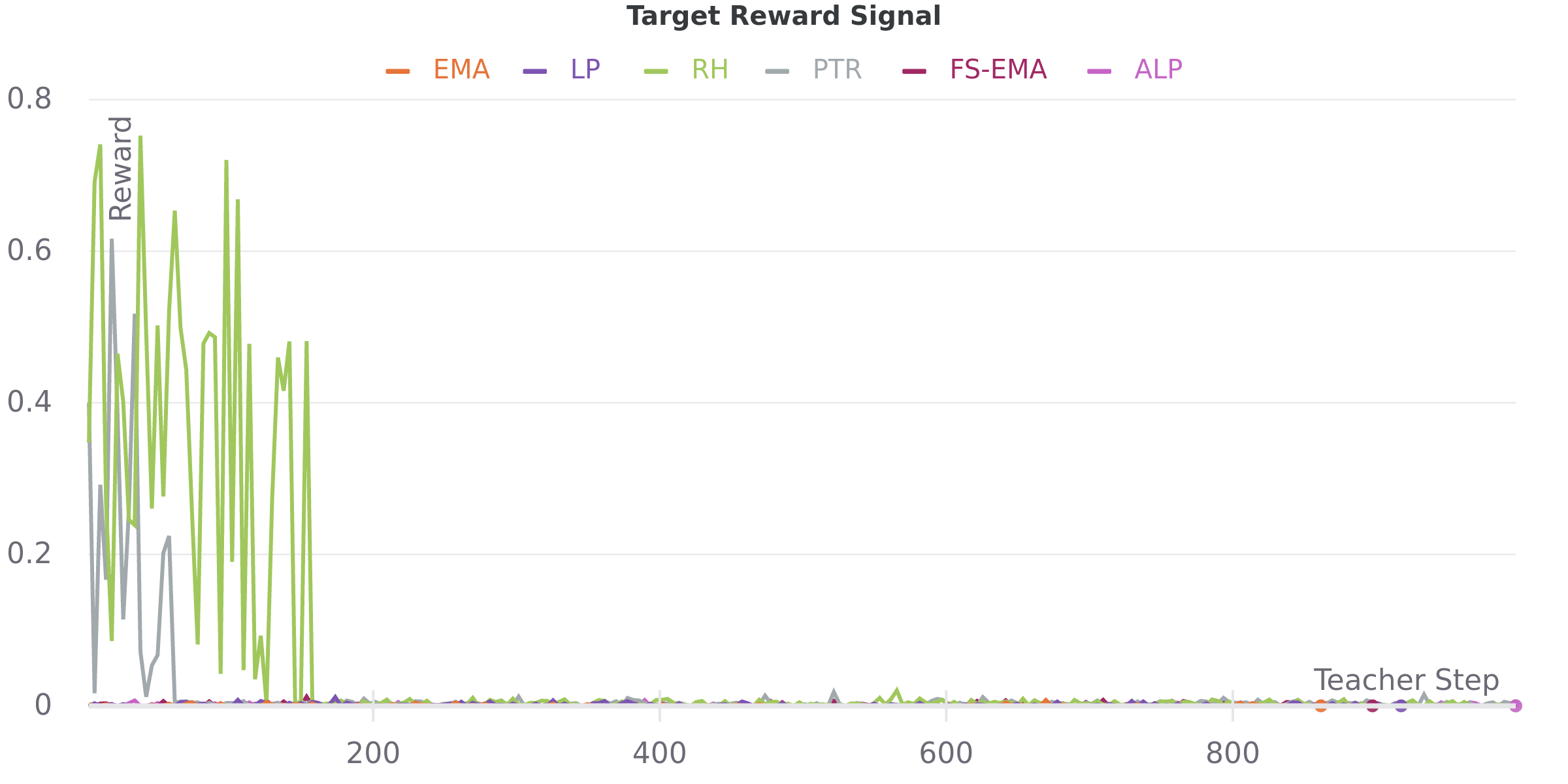}
    \caption{Visualization of the \textit{target reward} signal for each teacher observation type over the teachers training cycle.}
    \label{fig:rewardSignal}
\end{figure}

In \cref{fig:rewardSignal} we plot the \textit{target reward} signal along the steps in the CMDP.
The reward signal for the EMA, FS-EMA, LP, and ALP experiments are very close to zero during training, therefore not providing the teacher agent with a helpful signal.
For RH and PTR, the reward for the first 200 teacher steps scatters around 0.01 and 0.8 and then degenerates to the same reward signal as for the other observation types.
Surprisingly, this reward signal provided enough information to the teacher to better perform task sequencing than the baselines.

The \textit{target reward} pushes the teacher agent towards solving the KeyCorridor-S3R3 environment, but at the end of training the agent fails to solve this environment with both reward signals.
With both approaches, we did not manage to solve the challenging KeyCorridor-S3R3 environment.

\subsection{Choosing Alpha for Exponential Moving Average}
\label{sec:alphaExperiments}
In this section, we optimize the $\alpha$ value for the exponential moving average.
An $\alpha$ value close to one favors current values over old values in a time series.
We experimented with $\alpha \in [0.1, 0.3, 0.5, 0.7, 0.9]$ and reported the results in \cref{tab:gridworldAlpha}.
The teacher agents use a policy transfer and the \textit{source task reward}.

By looking at the results in \cref{tab:gridworldAlpha} it is hard to settle for an $\alpha$ value.
While an $\alpha$ of 0.5 or 0.9 have the highest total mean returns, the results for challenging environments are the best when selecting an alpha value of 0.3.
The experiments with a value of 0.7 and 0.9 suffer from a high standard deviation in the reported total mean return.

\begin{table}[!htb]
    \centering
    \begin{adjustbox}{width=\textwidth}\begin{tabular}{cccccc}
        \hline
        \textbf{Environment / Metric} & \textbf{0.1} & \textbf{0.3} & \textbf{0.5} & \textbf{0.7} & \textbf{0.9}\\
        \hline
        \hline
        Total mean return & $3.49 \pm 0.43$ & $3.01 \pm 0.39$ & $\mathbf{4.35 \pm 0.42}$ & $3.89 \pm 0.58$ & $4.23 \pm 0.95$ \\
        \% environments solved & 50\% & \textbf{61\%} & 55\% & 55\% & \textbf{61\%} \\
        Empty-16x16      & $0.32 \pm 0.06$ & $0.58 \pm 0.27$          & $0.82 \pm 0.34$ & $\mathbf{0.97 \pm 0.4}$ & $0.94 \pm 0.28$ \\
        FourRooms        & $0.08 \pm 0.01$ & $\mathbf{0.13 \pm 0.03}$ & $0.08 \pm 0.01$ & $0.09 \pm 0.04$ &  $0.06 \pm 0.01$\\
        DoorKey-16x16    & $0.0 \pm 0.0$   & $\mathbf{0.05 \pm 0.02}$ & $0.17 \pm 0.06$ & $0.0 \pm 0.0$   &  $0.02 \pm 0.01$\\
        MultiRoom-N2-S4  & $0.15 \pm 0.06$ & $0.08 \pm 0.03$          & $0.15 \pm 0.05$ & $0.13 \pm 0.02$ & $\mathbf{0.29 \pm 0.12}$ \\
        KeyCorridor-S3R3 & $0.0 \pm 0.0$   & $\mathbf{0.01 \pm 0.004}$& $0.0 \pm 0.0$   & $ 0.0 \pm 0.0 $ &  $0.004 \pm 0.002$ \\
    \end{tabular}\end{adjustbox}
    \caption{We compare the different $\alpha$ values for the EMA observation type.
    Each configuration is run with three different random seeds.
    The maximum return of 100 episodes at the end of 1'000 teacher steps is reported.
    Additionally, we provide the standard deviation between the three runs.}  
    \label{tab:gridworldAlpha}
\end{table}

\begin{figure}[ht]
    \centering
    \includegraphics[width=0.7\linewidth]{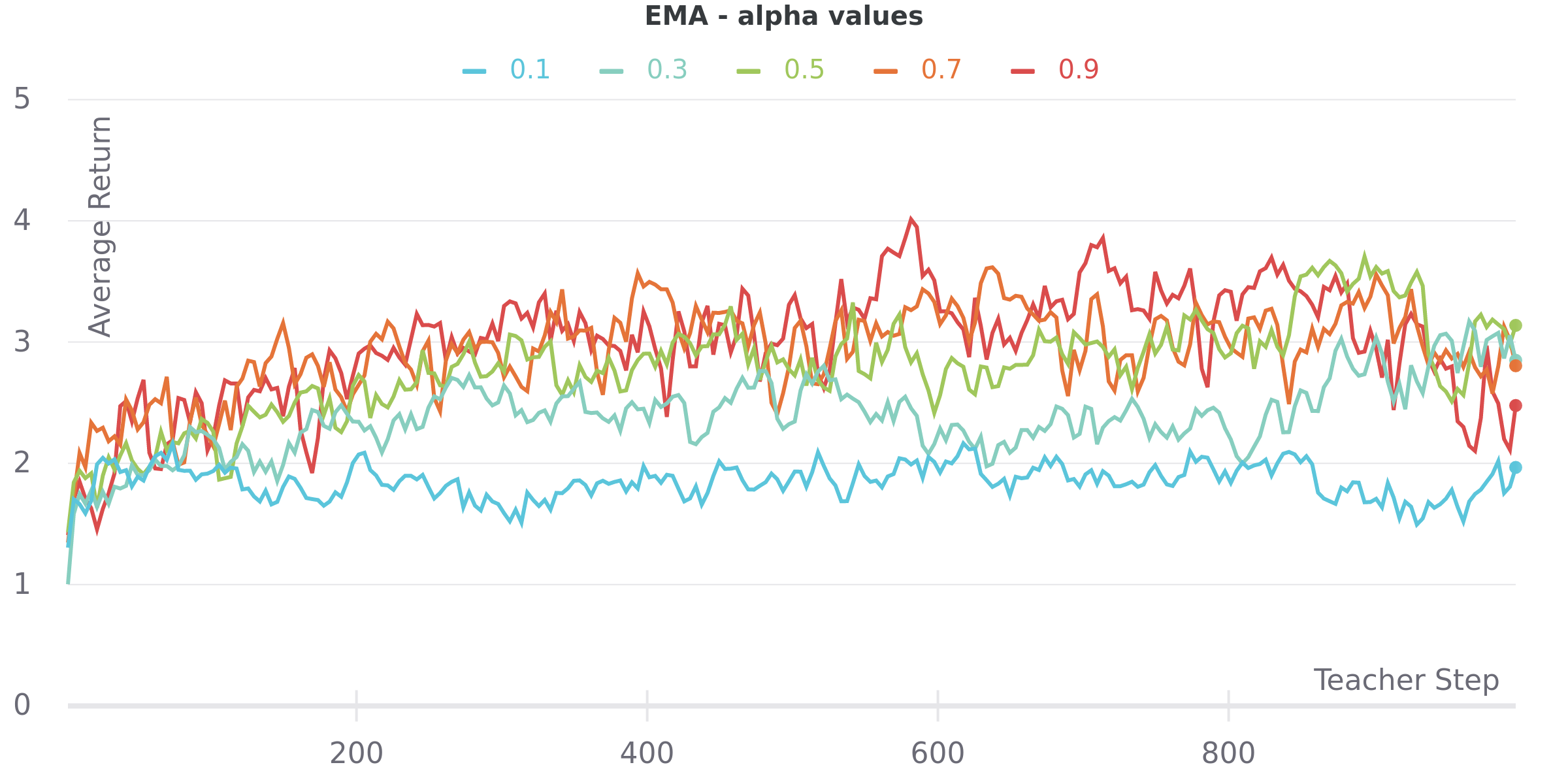}
    \caption{The average learning curve of three runs with three different random seeds for each $\alpha$ value.}
    \label{fig:alphaLC}
\end{figure}

In \cref{fig:alphaLC} we plot the average learning curve over three different random seeds for all $\alpha$ values.
An $\alpha$ value of 0.1 yields the worst results.
The learning curve of a value with 0.3 is below higher $\alpha$ values but surpassed most of the other experiments in the last 20 CMDP steps.
The learning curves for an $\alpha$ value of 0.5, 0.7, and 0.9 follow each other closely.
Our experiments do not allow us to select a clear winner.
Because of the highest total mean return combined with a relatively small standard deviation, we propose to use an $alpha$ value of 0.5.

\clearpage

\subsection{Sample Efficiency}
\label{sec:sampleEfficiency}
\begin{figure}[htb]
    \centering
    \begin{subfigure}{0.49\textwidth}
        \centering
        \includegraphics[width=0.99\linewidth]{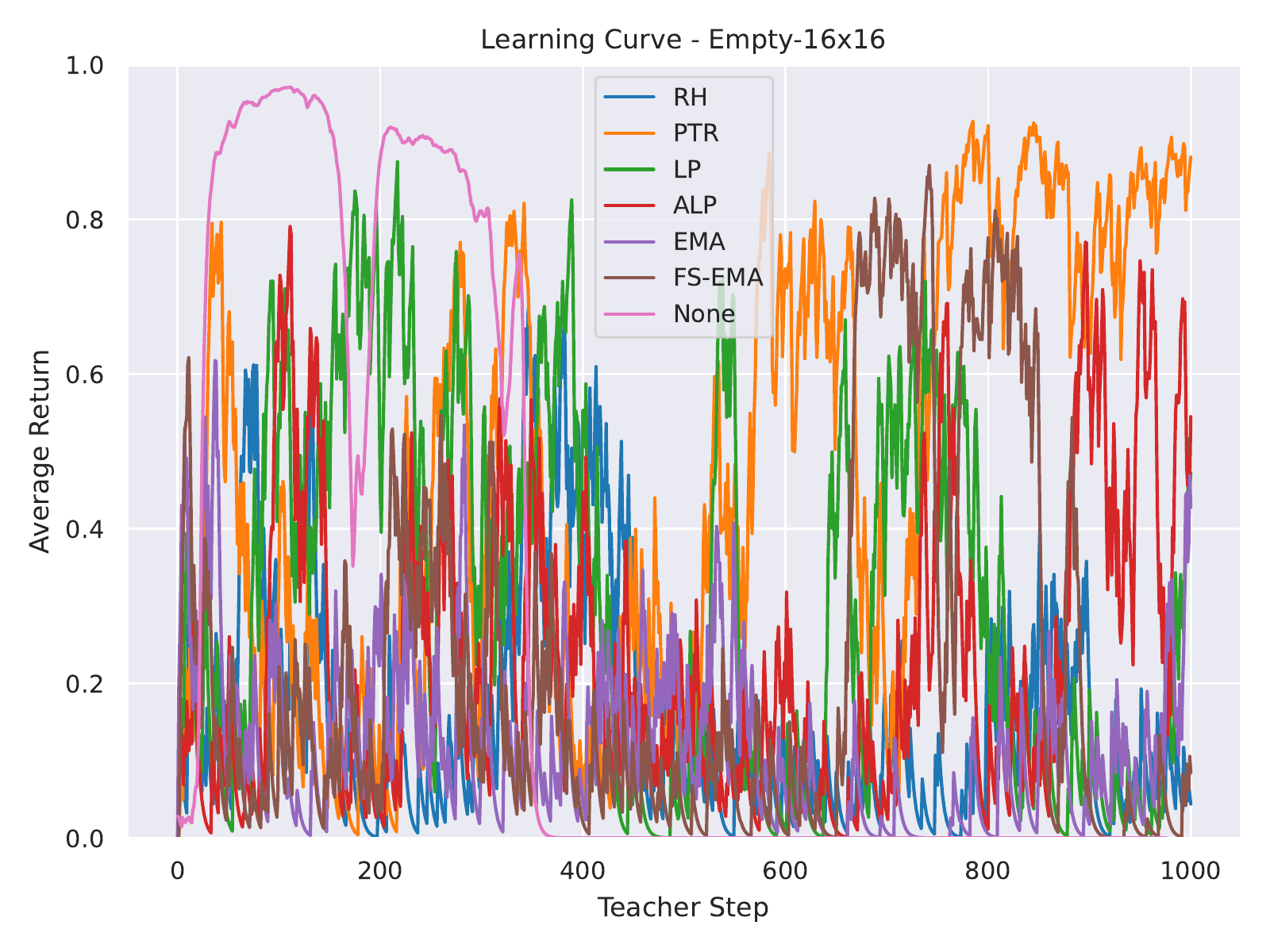}
    \end{subfigure}
    \begin{subfigure}{0.49\textwidth}
        \centering
        \includegraphics[width=0.99\linewidth]{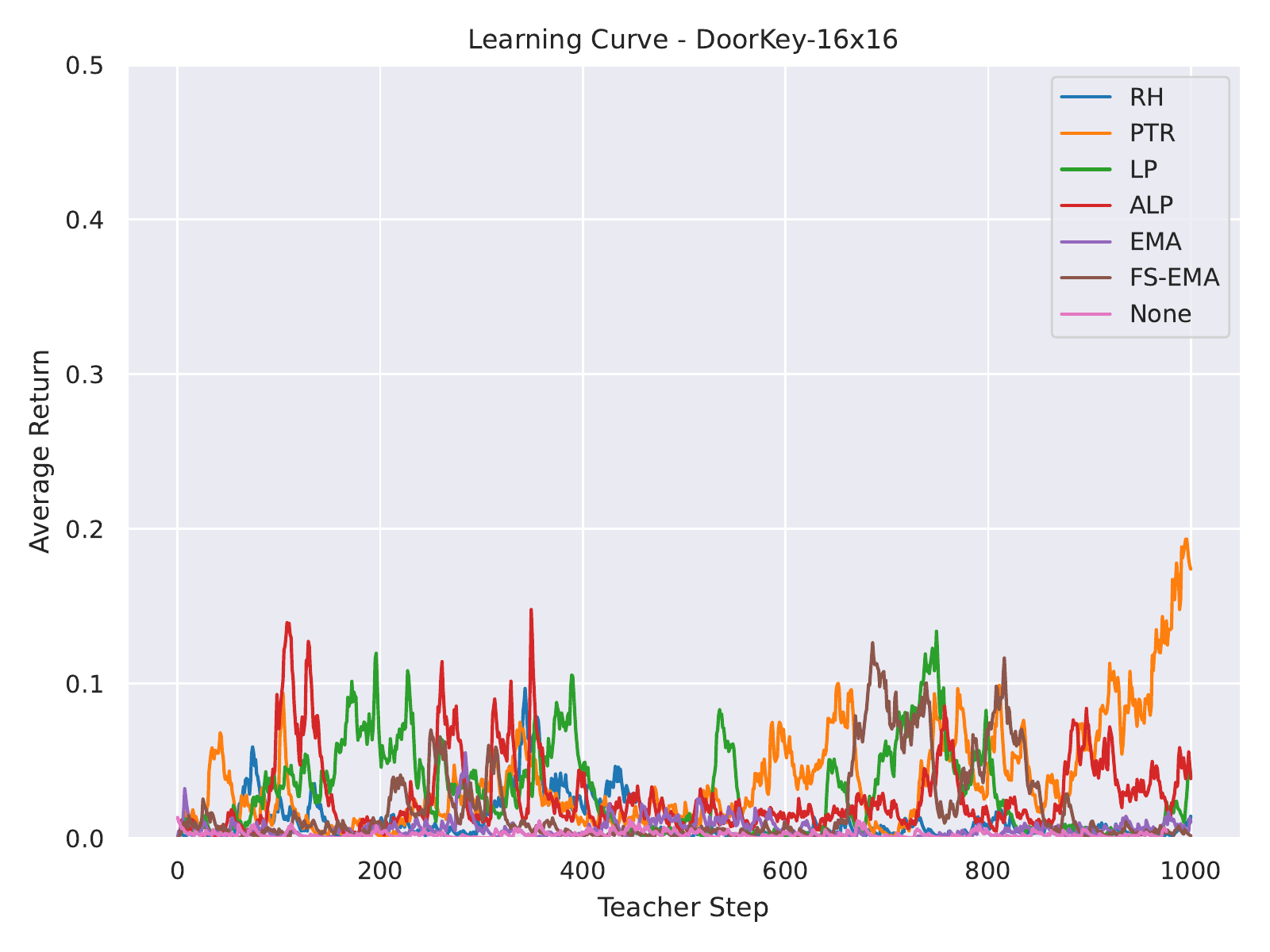}
    \end{subfigure}
    \begin{subfigure}{0.49\textwidth}
        \centering
        \includegraphics[width=0.99\linewidth]{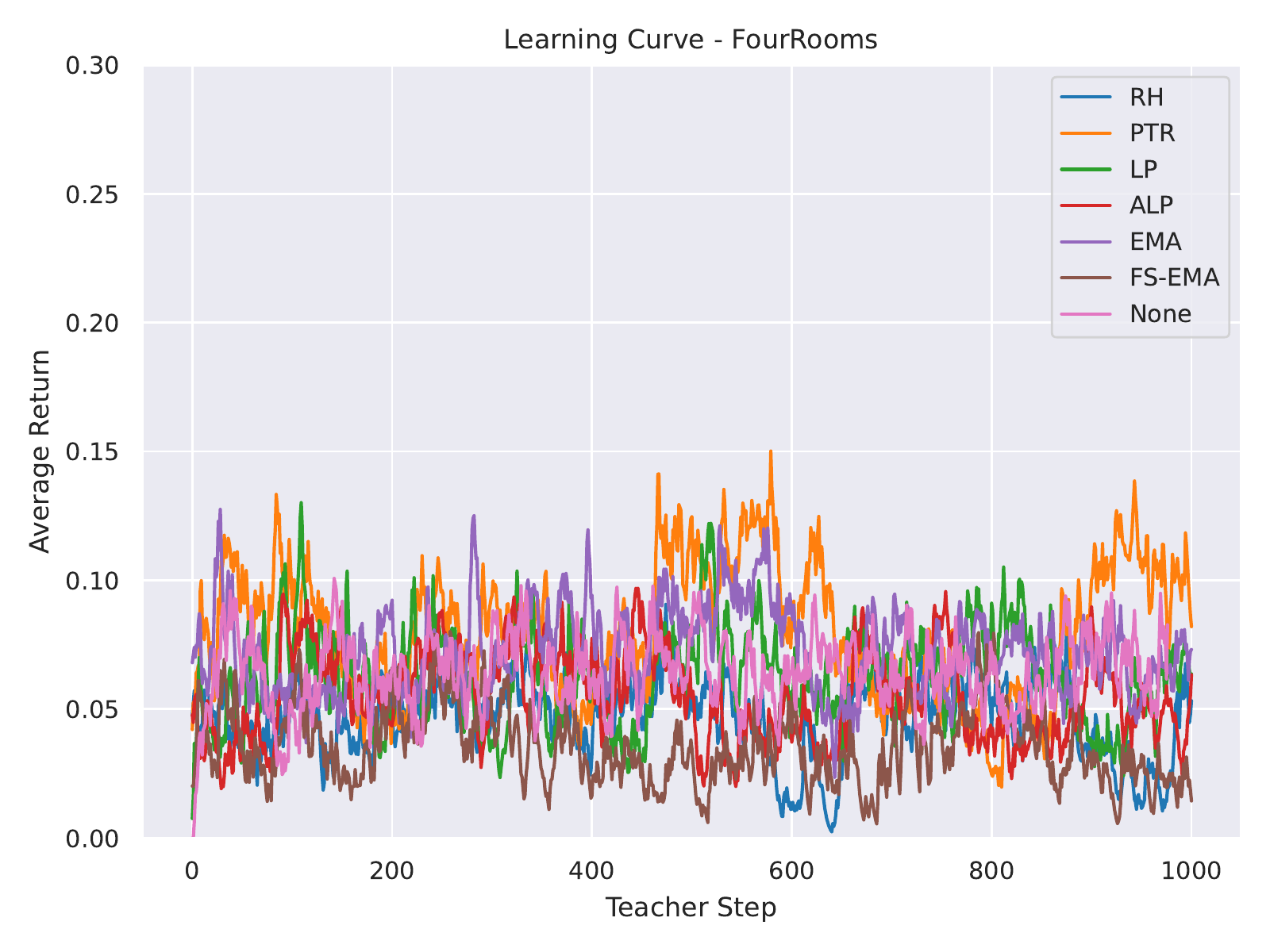}
    \end{subfigure}
    \caption{We visualize the learning curves of our agent when trained in the teacher-student setup and directly on the target task for the Empty-16x16, DoorKey16x16, and FourRooms environment.
    We use a teacher with policy transfer, \textit{source task reward} signal, and the six partially observable inputs. 
    The learning curve of the agent trained directly on the task is labeled with "None".}
    \label{fig:sampleEfficiency}
\end{figure}

In this section, we analyze the sample efficiency of an agent when trained in the teacher-student setting compared to trained directly on the target task.
Learning curves compare the agent's experience with their performance.
We use the teacher step on the x-axis to measure the experience. One teacher step equals 10'000 steps in an environment.
The performance is measured by the average return on 100 evaluation episodes during the training.
In \cref{fig:sampleEfficiency} we plot three learning curves, one for each of the following environments: Empty-16x16, DoorKey16x16, and FourRooms.
The sample efficiency of an agent can be measured as the area under the curve with transfer minus the area under the curve without transfer.
We do the evaluation manually by looking at the curves.

In the Empty16x16 environment, the sample efficiency of the agents with curriculum learning is negative for the first 400 teacher steps.
After 400 teacher steps, the agent trained directly in the Empty16x16 environment suffers from a catastrophic update and fails to recover from then on.
It seems like the agent got stuck in a bad local optimum.
This is probably due to a bad choice in the hyperparameters.
We note that agents trained directly on the Empty16x16 environment with three random seeds showed the same behavior.

In the DoorKey16x16 environment, all curriculum learning agents have a positive sample efficiency.
The learning curve is very noisy due to the training in different environments.
Although we notice an improvement in the sample efficiency, the agent's performance is not monotonically increasing and dropping to the same performance level as when trained directly in the DoorKey16x16 environment.
One could perform early stopping when crossing a reward threshold to solve this issue, but defining such a threshold is not straightforward.

There is no improvement in the sample efficiency in the FourRooms environment.
Curriculum learning with the PTR teacher shows the best results with a slight improvement in sample efficiency during some learning stages. 

\clearpage

\subsection{Generality of Agents}
\label{sec:generality}
We compare the generality of agents trained with curriculum learning with agents trained in a single environment.
To measure the generality, we evaluate the agent at the end of training on five different random seeds for 100 episodes on each environment in the task-set $\mathcal{V}$.
After the evaluation, we calculate the total mean return and the percentage of environments solved with five random seeds and report the median of both measurements in \cref{tab:generality}.

Agents trained with curriculum learning are more general than agents trained in a single environment.
With every teacher, we could solve more environments than training in one environment.
The agent trained solely in Empty-6x6 scored a total mean return of 3.25. Most of the obtained reward is received in the easy Empty environment family.
Some environments are especially useful to generalize to other environments.
The FourRooms and MultiRoom-N2-S4 environment solve a high percentage of environments.

\begin{table}[htb]
    \centering
    \begin{tabular}{lcc}
        \hline
        Environment trained on & Total mean return & \% environments solved \\
        \hline
        \hline
        Empty-5x5        & 2.1  & 33\% \\
        Empty-6x6        & 3.25 & 28\% \\
        Empty-8x8        & 0.32 & 22\% \\
        Empty-16x16      & 0.02 & 6\% \\
        FourRooms        & 2.41 & 56\%\\
        DoorKey-5x5      & 1.89 & 28\% \\
        DoorKey-6x6      & 2.58 & 39\% \\
        DoorKey-8x8      & 1.0  & 44\% \\
        DoorKey-16x16    & 0.6  & 39\% \\
        MultiRoom-N2-S4  & 1.48 & 56\% \\
        MultiRoom-N4-S5  & 0.66 & 39\% \\
        MultiRoom-N6-S10 & 0.99 & 33\% \\
        KeyCorridor-S3R1 & 0.47 & 28\% \\
        KeyCorridor-S3R2 & 0.53 & 28\% \\
        KeyCorridor-S3R3 & 0.32 & 22\% \\
        KeyCorridor-S4R3 & 0.21 & 11\% \\
        KeyCorridor-S5R3 & 0.11 & 11\% \\
        KeyCorridor-S6R3 & 0.72 & 28\% \\
        Curriculum RH    & 2.64 & 56\% \\
        Curriculum PTR   & \textbf{4.14} & 61\% \\
        Curriculum LP    & 3.9  & 61\% \\
        Curriculum ALP   & 3.06 & \textbf{67\%} \\
        Curriculum EMA   & 3.93 & 61\% \\
        Curriculum FS-EMA& 3.59 & 61\% \\
    \end{tabular}
    \caption{We report the median measurements obtained by evaluating each agent with five random seed for 100 episodes on every environment in $\mathcal{V}$.}
    \label{tab:generality}
\end{table}

\clearpage

\section{Google Football Environment}
\label{sec:GFenvironment}
Google Research Football environment \citep{kurach2019google} is a novel 3D RL environment to provide a highly optimized, stochastic, and open-source simulation.
The environment provides single-agent RL, where the agent controls all players of his team, and multi-agent RL, where a separate agent controls each player.
It is also possible to research the effect of self-play, where the agent plays against different versions of itself.
The environment provides a comprehensive set of progressively more demanding and diverse scenarios with the \textit{Football Academy}.
These scenarios enable us to analyze our algorithm on a range of tasks requiring different levels of abstractions and different tactics.

\begin{figure}[h]
    \centering
    \includegraphics[width=12cm]{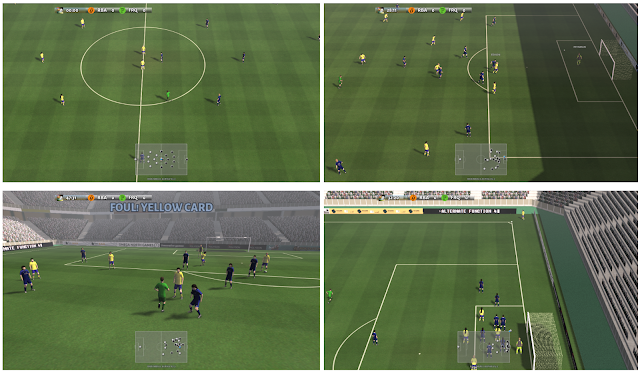}
    \caption{The Google Football Engine is a football simulation which supports the major football rules like kickoffs (top left), goals (top right), fouls, cards (bottom left), corner and penalty kicks (bottom right), and offside. \citep{kurach2019google}}
    \label{fig:environment}
\end{figure}

The engine implements a full 11 vs. 11 football game with the standard rules including goal kicks, corner kicks, yellow and red cards, offsides, handballs, and penalty kicks as shown in \cref{fig:environment}. 
This full 11 vs. 11 football game, consisting of 3000 frames, is called \textit{Football Benchmark}.
 
Players have different characteristics like speed or accuracy, but both teams have the same set of players. 
Further, players are getting tired over time, which influences their behavior and skills.

\subsection{State \& Observations}
There are three ways to represent the environment state at the current time step to the reinforcement learning agent (called observation).

\textit{pixel}. The representation is a 1280 x 720 x 3 tensor corresponding to the rendered screen. The scoreboard and a mini-map at the bottom of the image are present in this representation. The mini-map tells the location of the ball and the players of both teams.

\textit{Super Mini Map} (SMM). The SMM is a 72 x 96 x 4 tensor encoding information about both teams, the ball, and the currently active player.
The encoding is binary and indicates whether there is a player or the ball at the given coordination or not.

\textit{Floats}. This representation uses a 115-dimensional vector to capture the game state, player coordinates, ball coordinates, possession, and the active player.

\subsection{Actions}
The agent can execute one of 20 actions per time step.
The currently active player executes all of those actions, except the keeper rush action.
The active player moves by selecting one of eight dedicated moving actions.
There are four different ways to kick the ball (short pass, high pass, long pass, shot).
The move, sprint, and dribble actions are sticky and have to be ended explicitly by their respective stop action.
Finally, there are sliding and do-nothing actions. 

\subsection{Rewards}
The environment provides two different reward functions to choose from.
It is also possible to define custom reward functions to look into reward shaping.
In this work, we used the out-of-the-box reward functions.

\texttt{SCORING} rewards the agent when scoring a goal with a +1 reward and a -1 reward when conceding one.
This reward signal is sparse and can lead to no signal during the early stages of learning where the agent does not know how to overcome the opponent's defense.

\texttt{CHECKPOINT} is an additional shaped reward designed to overcome the sparsity of the \texttt{SCORING} reward signal.
Once per episode, the agent receives a +0.1 reward for getting closer to the opponent's goal measured by the Euclidean distance.
The opponent's field is divided into ten checkpoint regions, and the agent receives the +0.1 reward once for every region.
The agent also receives all non-collected checkpoint rewards when scoring to avoid penalizing agents that do not go through all the checkpoints before scoring.

\subsection{Football Academy Scenarios}
\label{sec:academyScenarios}
In addition to the full 11 vs. 11 football game, the environment allows agents to train on a set of 11 progressively more complex scenarios.
This set of tasks is called \textit{Football Academy} where it is possible to define custom scenarios to train agents for a particular situation.
In \cref{tab:footballAcademy} the available scenarios are described.

\renewcommand{\arraystretch}{1.5}
\begin{table}[htb]
    \centering
    \begin{tabular}{ p{0.25\textwidth} p{0.65\textwidth} }
         \hline
         \textbf{Name} & \textbf{Description} \\ 
         \hline
         \raggedright\textit{Empty Goal Close} & Our player starts inside the box with the ball and needs to score against an empty goal. \\ 
         \raggedright\textit{Empty Goal} & Our player starts in the middle of the field with the ball and needs to score against an empty goal. \\ 
         \raggedright\textit{Run to Score} & Our player starts in the middle of the field with the ball and needs to score against an empty goal. Five opponent players chase ours from behind. \\ 
         \raggedright\textit{Pass and Shoot with Keeper} & Two of our players try to score from the edge of the box. One is on the side with the ball and next to a defender. The other is at the center, unmarked and facing the opponent keeper. \\ 
         \raggedright\textit{Run, Pass and Shoot with Keeper} & Two of our players try to score from the edge of the box. One is on the side with the ball and unmarked. The other is at the center, next to a defender, and facing the opponent keeper. \\ 
         \raggedright\textit{Easy Counter-Attack} & 4 versus 1 counter-attack with keeper; all the remaining players of both teams run back towards the ball. \\ 
         \raggedright\textit{Hard Counter-Attack} & 4 versus 2 counter-attack with keeper; all the remaining players of both teams run back towards the ball. \\ 
         \raggedright\textit{11 versus 11 with Lazy Opponents} & Full 11 versus 11 game, where the opponents cannot move but intercept the ball if it is close enough to them. Our center-back defender has the ball at first. The maximum duration of the episode is 3000 frames instead of 400 frames. \\ 
         \raggedright\textit{11 versus 11 easy} & Full 11 versus 11 game, with a duration of 3000 frames per episode.
         The game starts with a kick-off, the team starting with the kick-off is assigned randomly.
         The agent receives a reward of -1 when receiving a goal and a reward of +1 when scoring one.
         The opponents difficulty is set to easy.\\ 
         \raggedright\textit{11 versus 11 medium} & Same as 11 versus 11 easy, but the opponents difficulty is set to medium. \\ 
         \raggedright\textit{11 versus 11 hard} & Same as 11 versus 11 easy, but the opponents difficulty is set to hard. \\ 
         \hline
    \end{tabular}
    \caption{Description of the Google Football environments taken directly from \citet{kurach2019google}.
    All scenarios end after 400 frames, or if the ball is lost, a team scores or the game stops (e.g., if the ball leaves the pitch or a free kick is awarded).}
    \label{tab:footballAcademy}
\end{table}

\subsection{MDP Statement}
\label{sec:mdp}

\paragraph{The Google research Football} environment does not fulfill the Markov property if we only consider the pixel input at time step $t$ as observation $o_t$.
We do not know in which direction the ball or the players are moving from a single observation.
It is possible to have two identical observations, but in one observation, the ball moves to the left of the pitch, and in the other, it moves to the right.
The direction and the velocity of the ball can not be inferred based on a single image.
The agent only partially observes its environment's state, making it a partially observable Markov decision process (POMDP).

There are a few possibilities to overcome this issue:
\begin{itemize}
    \item We can use another state representation than pixels.
    This representation would have to encode the direction and velocity of the ball and the players.
    \item We can use a long short-term memory (LSTM) \citep{hochreiter1997long} inside the agents function approximator.
    The observation $o_t$ together with the hidden state of the LSTM $h_t$ are enough to overcome the uncertainty about the current state $s_t$ of the environment.
    \item We can use the last $k$ observations to approximate the true state $s_t$.
    This is the solution proposed by \citet{Mnih2015}.
\end{itemize}

\begin{figure}[htb]
    \centering
    \includegraphics[width=10cm]{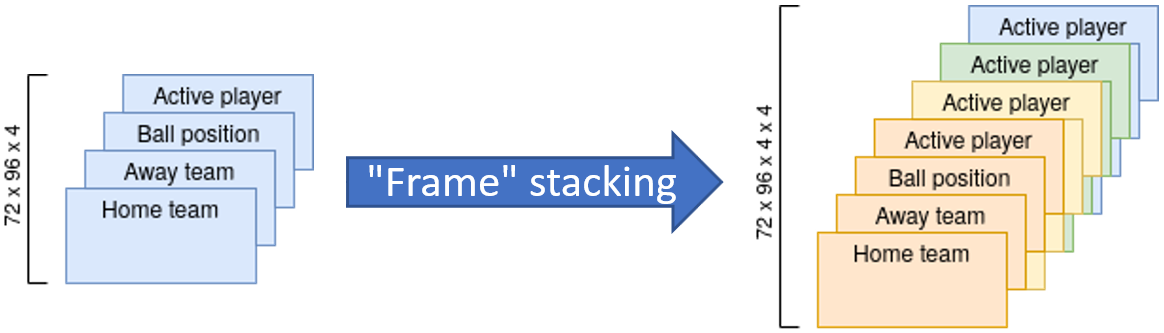}
    \caption{To approximate the Markov property we use a frame stack of the last 4 frames as an input for our neural networks.}
    \label{fig:frameStack}
\end{figure}

In our work, we used the last four observations stacked on top of each other, illustrated in \cref{fig:frameStack}, as input to approximate the true Markov state.

\clearpage
\section{Google Football Experiments}
\label{sec:gfootballExperiments}

\subsection{Experimental Setup}
\label{sec:gfootballSetup}

The experimental setup for the Google Football environment is similar to the grid world environment.
As in grid world, we train the teacher agent for $1'000$ steps, and for each teacher step, we train the student agent for $10'000$ steps in the selected environment.
The student agent is therefore trained for 10 million steps in total.
After each step, we evaluate the student on all environments in $\mathcal{V}$ for 100 episodes.
In our previous work \citep{schraner_2020}, we train the student agent for 50 million steps.
Due to a limit in computational resources, we limit the training in this thesis to 10 million student steps and only one random seed per experiment.
Executing one experiment with the described amount of steps takes around five days.
We fixed the knowledge transfer to policy transfer and used the \textit{source task reward} signal.
The hyperparameters are reported in \cref{sec:appendixGFootballStudentPPO}.
We use all Google Football environments listed in \cref{tab:footballAcademy}.
The student agent receives the \texttt{SCORING} reward signal.

\begin{figure}[h]
    \centering
    \includegraphics[width=0.43\textwidth]{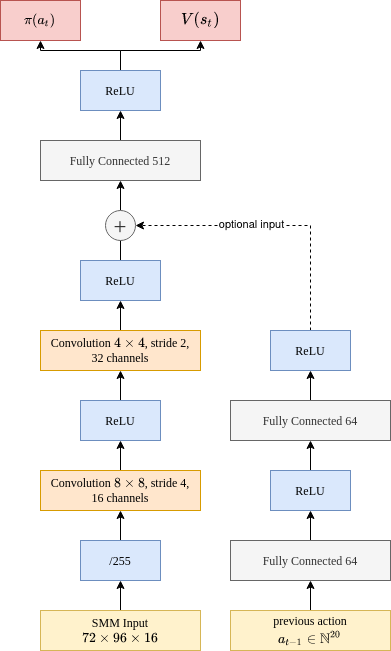}
    \caption{CNN network architecture with the stacked \textit{Super Mini Map} observation as input.
    The one-hot encoded previous action $a_{t-1}$ is used as an additional input.
    The policy and value function are two heads sharing the network torso.}
    \label{fig:GfootballStudentArchitecture}
\end{figure}

\paragraph{Network architecture} The network architecture for the teacher agent is described in \cref{sec:teacherArchitecture}.
The student's network architecture is the same as in our previous work \citep{schraner_2020} and depicted in \cref{fig:GfootballStudentArchitecture}.

Actor-critic algorithms can share the network torso and represent the policy and value function as individual heads.
This allows weight sharing between the two function approximators.
Sharing weights can improve the training time and forces the shared part to be more general.
This may prevent over-fitting.
As we share network weights, we get different gradients from different loss functions.
In our work, we made use of sharing weights.

Each network uses the SMM observation, with a shape of $72 \times 96 \times 4$.
The football field has a dimension of $72 \times 96$, and there is a separate channel for each team, the ball, and the currently active player.
To approximate the Markov property, we use a frame stack of the last four frames, as shown in \cref{fig:frameStack}.

\clearpage

\subsection{Results}
\label{sec:resultsFootball}
We transfer the best CMDP teacher settings from the MiniGrid environment to the Google Research Football environment.
The four baseline methods and our results in our previous work are used to compare the results in \cref{tab:GfootballResults}.
We report the average return of 100 episodes per environment after training on one random seed. 
Therefore the results should be interpreted with care.

The LP and Thompson baseline are by far the worst.
In the case of Thompson sampling, only one out of eleven environments were solved.
The LP teacher fails for the 11 vs. 11 full game.
Sampling environments uniformly works well on easy environments but fails for more challenging ones such as the counter-attack and 11 vs. 11 environments.
The window baseline is the strongest of all baselines.
With our proposed teacher-student setting we surpass our baselines on all environments.
The LP based teacher had the best results on empty goal close, empty goal and run to score, but the improvement over the other experiments is marginal.

Compared to the results in our previous work \citep{schraner_2020} reported in \cref{tab:prevwork_results2} we do not match the results.
Training optimized PPO for 50 million steps leads to a return of -1.4 in the 11 vs. 11 hard environment.
The highest score we obtained in this environment was -1.45 with the window baseline and LP teacher.
Using a manually defined curriculum, we archive a return between -2.08 and 1, depending on the curriculum.

\begin{table}[htb]
    \centering
    \begin{adjustbox}{width=\textwidth}
        \begin{tabular}{p{3cm}cccccccccc}
            \hline
             \raggedright{\textbf{Environment / Metric}} & \textbf{Uniform} & \textbf{LP} & \textbf{Thompson} & \textbf{Window} & \textbf{RH} & \textbf{PTR} & \textbf{LP} & \textbf{ALP} & \textbf{EMA} & \textbf{FS-EMA} \\
            \hline
            \hline
            \raggedright{Total mean return}            & -3.26 & -11.62& -2.67 & -0.81 & -0.3  & -0.28 & -0.5  & \textbf{1.84}  & -1.36 & -1.51 \\
            \raggedright{\% environments solved}          & 45\%  & 27\%  & 9\%   & 55\%  & 55\%  & \textbf{64\%}  & 45\%  & 45\%  & 45\%  & 45\%  \\
            \raggedright{Empty Goal Close}                & 0.99  & \textbf{1}     & -0.05 & \textbf{1}     & 0.96  & 0.98  & \textbf{1}     & 0.94  & 0.79  & \textbf{1}     \\
            \raggedright{Empty Goal}                      & 0.84  & \textbf{1}     & 0.64  & 0.93  & 0.85  & 0.67  & 0.98  & 0.93  & 0.6   & 0.76  \\
            \raggedright{Run to Score}                    & 0.88  & \textbf{0.98}  & 0.0   & 0.76  & 0.95  & 0.7   & 0.05  & \textbf{0.98}  & 0.67  & 0.91  \\
            \raggedright{Pass and Shoot with Keeper}      & 0.12  & 0.0   & -0.02 & 0.0   & 0.33  & 0.03  & 0.0   & \textbf{0.51}  & 0.0   & -0.07 \\
            \raggedright{Run, Pass and Shoot with Keeper} & 0.02  & 0.0   & 0.0   & 0.0   & -0.11 & -0.02 & 0.0   & \textbf{0.05}  & 0.0   & -0.17 \\
            \raggedright{Easy Counter-Attack}             & 0.0   & -0.02 & 0.0   & 0.23  & \textbf{0.33}  & 0.17  & 0.0   & 0.0   & 0.02  & 0.0   \\
            \raggedright{Hard Counter-Attack}             & 0.0   & 0.0   & 0.0   & 0.22  & \textbf{0.18}  & 0.05  & 0.03  & 0.0   & -0.05 & 0.02  \\
            \raggedright{11 vs. 11 lazy}                  & 0.0   & -0.1  & 0.0   & 0.05  & 0.0   & 0.02  & \textbf{0.16}  & 0.0   & 0.12  & 0.08  \\
            \raggedright{11 vs. 11 easy}                  & -1.55 & -3.8  & -0.55 & -0.53 & -0.42 & \textbf{0.0}   & \textbf{0.0}   & -0.23 & -0.85 & -0.55 \\
            \raggedright{11 vs. 11 medium}                & -1.98 & -5.28 & -1.48 & -1.23 & -1.75 & -1.2  & -1.08 & \textbf{-0.65} & -1.58 & -1.78 \\
            \raggedright{11 vs. 11 hard}                  & -2.63 & -4.98 & -1.5  &\textbf{-1.45}  & -1.88 & -1.85 & \textbf{-1.45} & -1.7  & -1.9  & -1.7  \\
        \end{tabular}
    \end{adjustbox}
    \caption{Google football results obtained on one random seed after training for 1'000 CMDP steps, equal to 10 million student steps. The average return over 100 episodes is reported. We use policy transfer between the source tasks and the \textit{source task reward} signal.}
    \label{tab:GfootballResults}
\end{table}

Using the ALP, RH, PTR, and LP observation types leads to the best results.
Whereas using ALP gives the highest total average and PTR the highest number of solved environments.
The EMA-based teachers fall behind, probably due to a not-tuned $\alpha$ value for the Google Football environments.

\begin{table}[htb]
    \centering
    \begin{tabular}{cc}
    \hline
        \textbf{Experiment} & \textbf{Return} \\
        \hline
        \hline
        PPO & -1.4 \\
        Best \textit{scenarios curriculum} & -0.85 \\
        Best \textit{11 vs 11 curriculum} & -2.08 \\
        Best \textit{increasing curriculum} & -0.48 \\
        Best \textit{smooth increasing curriculum} & 1 \\
        Prioritized level replay & -1.05 \\
    \end{tabular}
    \caption{Comparison of our different curriculum learning approaches of our previous work \citet{schraner_2020}, the average return over 100 episodes on the 11 vs 11 hard environment is reported.}
    \label{tab:prevwork_results2}
\end{table}

\clearpage

In \cref{tab:ScenariosVsBaselineScor} we report our results obtained in our previous work \citep{schraner_2019} by training an agent with IMPALA in a single Google Football environments.
With our curriculum learning approach, we can match and surpass the results of our previous work by only using a fifth of the samples.
Compared to the baselines published with the Google Football environment \citep{kurach2019google} we archived better results on the easy environments (Empty Goal, Run to Score, Pass and Shoot with Keeper) but slightly worse on the difficult environments (Run, Pass and Shoot with Keeper, Easy Counter-Attack, Hard Counter-Attack, and 11 vs. 11 lazy).

\begin{table}[htb]
    \centering
    \begin{tabular}{ccc}
        \hline
         \textbf{Scenario} & \textbf{Ours@50M} & \textbf{Google Baseline@50M} \\
         \hline
         \hline
         Empty Goal Close & 0.99 & 1.0 \\
         Empty Goal & 0.84 & 0.86 \\
         Run to Score & 0.88 & 0.88 \\
         Pass and Shoot with Keeper & 0.0 & 0.66 \\
         Run, Pass and Shoot with Keeper & -0.05 & 0.18 \\
         Easy Counter-Attack & 0.0 & 0.5 \\
         Hard Counter-Attack & -0.02 & 0.2 \\
         11 vs. 11 lazy & 0.01 & 0.2 \\
         11 vs. 11 easy & -1.59 & -0.35 \\
         11 vs. 11 medium & -1.6 & -0.79 \\
         11 vs. 11 hard & -2.78 & -1.16
    \end{tabular}
    \caption{Results on the \textit{Football Academy} scenarios obtained in our previous work \citep{schraner_2019} and the Google baseline \citep{kurach2019google} with the IMPALA algorithm.
    The reported results is the average reward of 100 episodes after a training on 50 million frames.}
    \label{tab:ScenariosVsBaselineScor}
\end{table}
\clearpage
\chapter{Discussion}
\label{sec:discussion}

Recent work in curriculum learning for rl has shown promising results in improving sample efficiency and asymptotic performance in challenging environments.
Most of these works focused on automatic task generation e.g., by using procedural generated environments \citep{10.1145/3321707.3321799}, changing the initial state distribution to easy starts \citep{pmlr-v78-florensa17a}, creating additional synthetic goals to guide the student \citep{DBLP:conf/iclr/RacaniereLSRFL20} or generating auxiliary tasks \citep{pmlr-v80-riedmiller18a}.
Using self-play to create a curriculum was a core idea in AlphaStar \citep{Vinyals2019}.
Methods to automatically select tasks in a teacher-student setting are also popular.
Most of those methods use heuristics or sampling-based approaches to select the tasks \citep{8827566, 8850690, ijcai2019-320, ijcai2017-353, AAMAS19-Narvekar, DBLP:journals/corr/abs-2106-14876}.
In our thesis, we propose a new approach to formulate the task sequencing problem as an MDP and train a teacher rl agent to perform the task sequencing while training the student simultaneously.

Our experiments show that performing task sequencing with an rl teacher agent is superior to heuristic-based task sequencing.
On the grid world environment, we were able to outperform all baselines significantly.
Our approach failed to increase the asymptotic performance on most of the MiniGrid when compared to tablua-rasa rl.
Training an agent directly in easy grid world environments leads to better performance compared to curriculum learning approaches.
Our Google Football experiments improved the asymptotic performance on easy environments such as Empty Goal, Run to Score, and Pass and Shoot with Keeper.
We believe that the root cause of failing on challenging environments lies in the teacher's reward signal.
Curriculum learning with the \textit{source task reward} signal is in some cases beneficial for more challenging environments such as DoorKey-16x16, but it is difficult to target the teacher agent towards specific environments.
Using a targeted reward signal to guide the teacher agent towards more challenging environments is often insufficient, as the teacher faces a sparse reward signal in that setting.
When providing the teacher with a sparse reward signal, our teacher-student settings collapses towards a uniform sampling approach.

The teacher agent's learning curves did not converge after 1'000 steps.
Training both agents beyond 1'000 CMDP steps might increase the asymptotic performance and help the student agent solve more challenging environments.
This would increase the asymptotic performance at the cost of sample efficiency.
We fixed the CMDP and student steps to match the number of training steps used in our no curriculum learning experiments to carry out a fair comparison.
When specifying the number of teacher and student steps, we have to balance the teacher's and student's performance.
Using fewer teacher steps might not be sufficient for the teacher to learn how to perform task sequencing but would give the student more time to converge in the selected environment.
Using more teacher steps is beneficial for the teacher, but on the other hand, the student has less time to learn the selected task.
Using fewer student steps might lead to a noisy reward signal for the teacher agent.
In future work, it would be interesting to analyze the importance and effect of those values.
One could also train the student until convergence or make the number of student steps part of the teacher's action space.

When comparing the three transfer methods policy, reward and combined, it turns out that the policy transfer leads to the best results.
Reward transfer is volatile for both the rl teacher agents and the baselines.
Combining reward transfer with policy transfer fails.
After a few CMDP steps, the added reward bonus covers the environment's reward, forcing the student agent to maximize its reward signal.

We compared six partial observable state representations and one approximately fully observable state representation (PCA) for the CMDP state.
The PCA representation type failed utterly.
There are other ways to represent neural network weights in a compact embeddings, such as autoencoders or network distillations.
In future work, we can investigate other methods to represent the student weights.
The results of the six partial observable representations did not differ too much from each other.
Overall the PTR, LP, and EMA representation resulted in the highest asymptotic rewards and the most general student agents for the MiniGrid environments.
In our Google Football environment RH, PTR, LP and ALP achived the highest asymptotic reward and percentage of solved environments.
In that environment tuning the $\alpha$ value for the EMA representation might improve the results.

The proposed curriculum learning approach is beneficial for multi-task reinforcement learning.
The reward across all tasks and the overall percentage of solved environments are significantly higher compared to our baselines or when training an agent on a single environment.

We investigated the sample efficiency of our teacher-student setup in three grid world environments.
The sample efficiency improves in the Empty-16x16 and DoorKey16x16 environment but remains similar in the FourRooms environment.
We discovered that training in the proposed curriculum setup is noisy regarding the student's performance throughout training.
The sample efficiency on the Google Football environments increased considerably.
By only using one-fifth of the samples, we almost matched or surpassed the performance on 8 out of 11 environments compared to directly training in those environments.
Training agents for the 11 vs. 11 football game remains difficult with our proposed curriculum learning approach.
The asymptotic performance remained the same compared to training directly in an environment.
Using a careful, manually defined curriculum improves the asymptotic performance compared to our teacher-student setup.
Our previous work concluded that defining such a curriculum by hand requires a lot of tweaking and domain knowledge.

In this work, we analyzed different knowledge transfer methods, teacher observation types, and reward signals.
Using policy transfer combined with the PTR, LP, or EMA representation and the \textit{source task reward} reward signal, we observe the following:
\begin{itemize}
    \item The generality of our agent improves.
    \item The asymptotic performance in some environments increases compared to using no curriculum.
    \item Training an rl agent to perform task sequencing is superior over our heuristic-based baselines.
    \item The sample efficiency compared to no curriculum does increase in one out of three analyzed grid world environments.
    \item The sample efficiency compared to no curriculum does increase by a factor of five on 8 out of 11 Google Football environments.
\end{itemize}

The improvement in the generality and sample efficiency with curriculum has to be investigated more carefully in future research.
Another weakness of our work is the limited hyperparameter search.
The effect of training teachers or students for more steps remains unclear.
Other types of rl algorithms such as DQN or traditional tabular methods for the teacher agent have to be tested.
The teacher network architecture should be tuned further.
Our agents trained without curriculum learning on the grid world environments results in zero rewards for hard environments.
To perform more meaningful analysis, we propose to tune those baseline agents and then transfer them to the teacher-student setting.
It remains unclear if other PPO hyperparameters for the teacher training would improve performance and how well these parameters transfer across all methods.
\clearpage
\chapter{Conclusion}
\label{sec:conclusion}

In this thesis, we successfully layout a teacher-student curriculum learning approach with automated online task sequencing to improve the generality and sample efficiency of our rl agents.
Our CMDP framework allows training a teacher and a student agent simultaneously, where the teacher selects tasks for the student.
The student's asymptotic performance does not increase for most grid world and Google Football environments compared to non-cl agents.
We evaluated multiple transfer methods, teacher reward signals, and CMDP observations on the MiniGrid and the Google Research Football environment.
This work is a step towards successful multi-task reinforcement learning agents.

We found that knowledge transfer through policy transfer works well and is robust across a variety of teachers.
In future work, it might be interesting to experiment with additional knowledge transfer methods such as macro actions \citep{10.5555/3157382.3157488} and task models \citep{8351991}.
The choice of teacher observation is not as important as the transfer method.
PTR, LP, ALP, and EMA proved to provide useful observations.
The teacher's reward signal is essential for training the teacher fast.
The \textit{source task reward} provides the teacher agent with a rich reward signal and encourages to find a general student agent.
With this reward signal, it is difficult to target specific environments to solve, but using the \textit{target reward} signal is too sparse.
One could use a weighted form of the \textit{source task reward} to lay focus on a subset of the tasks while still providing a rich reward signal.
Another option would be a combination of both reward signals, where the \textit{source task reward} signal is used at the beginning of training and slowly exchanged with the \textit{target reward}.

For future work, it would be interesting to experiment with different rl algorithms to train the teacher.
One could use traditional rl methods such as Q learning or off-policy algorithms with experience replay such as algorithms from the DQN family.
More work must be dedicated to the hyperparameter tuning of the rl algorithms, teacher-student step settings, and network architectures.

Although our work did not increase the asymptotic performance on most of the MiniGrid and Google Football environments, we increased the overall performance across the tasks and the sample efficiency.
Future work should experiment with other domains such as robotics and further tune the hyperparameters to better understand our findings.
\clearpage
% ...
%

\bibliographystyle{bibliography/IEEEtranN}
\bibliography{bibliography/references}

\begin{thebibliography}{60}
\providecommand{\natexlab}[1]{#1}
\providecommand{\url}[1]{#1}
\csname url@samestyle\endcsname
\providecommand{\newblock}{\relax}
\providecommand{\bibinfo}[2]{#2}
\providecommand{\BIBentrySTDinterwordspacing}{\spaceskip=0pt\relax}
\providecommand{\BIBentryALTinterwordstretchfactor}{4}
\providecommand{\BIBentryALTinterwordspacing}{\spaceskip=\fontdimen2\font plus
\BIBentryALTinterwordstretchfactor\fontdimen3\font minus
  \fontdimen4\font\relax}
\providecommand{\BIBforeignlanguage}[2]{{%
\expandafter\ifx\csname l@#1\endcsname\relax
\typeout{** WARNING: IEEEtranN.bst: No hyphenation pattern has been}%
\typeout{** loaded for the language `#1'. Using the pattern for}%
\typeout{** the default language instead.}%
\else
\language=\csname l@#1\endcsname
\fi
#2}}
\providecommand{\BIBdecl}{\relax}
\BIBdecl

\bibitem[Silver et~al.(2016)Silver, Huang, Maddison, Guez, Sifre, van~den
  Driessche, Schrittwieser, Antonoglou, Panneershelvam, Lanctot, Dieleman,
  Grewe, Nham, Kalchbrenner, Sutskever, Lillicrap, Leach, Kavukcuoglu, Graepel,
  and Hassabis]{Silver2016}
\BIBentryALTinterwordspacing
D.~Silver, A.~Huang, C.~J. Maddison, A.~Guez, L.~Sifre, G.~van~den Driessche,
  J.~Schrittwieser, I.~Antonoglou, V.~Panneershelvam, M.~Lanctot, S.~Dieleman,
  D.~Grewe, J.~Nham, N.~Kalchbrenner, I.~Sutskever, T.~Lillicrap, M.~Leach,
  K.~Kavukcuoglu, T.~Graepel, and D.~Hassabis, ``Mastering the game of go with
  deep neural networks and tree search,'' \emph{Nature}, vol. 529, no. 7587,
  pp. 484--489, 2016.
\BIBentrySTDinterwordspacing

\bibitem[Silver et~al.(2017)Silver, Schrittwieser, Simonyan, Antonoglou, Huang,
  Guez, Hubert, Baker, Lai, Bolton, Chen, Lillicrap, Hui, Sifre, van~den
  Driessche, Graepel, and Hassabis]{Silver2017}
\BIBentryALTinterwordspacing
D.~Silver, J.~Schrittwieser, K.~Simonyan, I.~Antonoglou, A.~Huang, A.~Guez,
  T.~Hubert, L.~Baker, M.~Lai, A.~Bolton, Y.~Chen, T.~Lillicrap, F.~Hui,
  L.~Sifre, G.~van~den Driessche, T.~Graepel, and D.~Hassabis, ``Mastering the
  game of go without human knowledge,'' \emph{Nature}, vol. 550, no. 7676, pp.
  354--359, 2017.
\BIBentrySTDinterwordspacing

\bibitem[Silver et~al.(2018)Silver, Hubert, Schrittwieser, Antonoglou, Lai,
  Guez, Lanctot, Sifre, Kumaran, Graepel, Lillicrap, Simonyan, and
  Hassabis]{Silver1140}
\BIBentryALTinterwordspacing
D.~Silver, T.~Hubert, J.~Schrittwieser, I.~Antonoglou, M.~Lai, A.~Guez,
  M.~Lanctot, L.~Sifre, D.~Kumaran, T.~Graepel, T.~Lillicrap, K.~Simonyan, and
  D.~Hassabis, ``A general reinforcement learning algorithm that masters chess,
  shogi, and go through self-play,'' \emph{Science}, vol. 362, no. 6419, pp.
  1140--1144, 2018.
\BIBentrySTDinterwordspacing

\bibitem[Mnih et~al.(2015)Mnih, Kavukcuoglu, Silver, Rusu, Veness, Bellemare,
  Graves, Riedmiller, Fidjeland, Ostrovski, Petersen, Beattie, Sadik,
  Antonoglou, King, Kumaran, Wierstra, Legg, and Hassabis]{Mnih2015}
\BIBentryALTinterwordspacing
V.~Mnih, K.~Kavukcuoglu, D.~Silver, A.~A. Rusu, J.~Veness, M.~G. Bellemare,
  A.~Graves, M.~Riedmiller, A.~K. Fidjeland, G.~Ostrovski, S.~Petersen,
  C.~Beattie, A.~Sadik, I.~Antonoglou, H.~King, D.~Kumaran, D.~Wierstra,
  S.~Legg, and D.~Hassabis, ``Human-level control through deep reinforcement
  learning,'' \emph{Nature}, vol. 518, no. 7540, pp. 529--533, 2015.
\BIBentrySTDinterwordspacing

\bibitem[OpenAI et~al.(2019{\natexlab{a}})OpenAI, :, Berner, Brockman, Chan,
  Cheung, Debiak, Dennison, Farhi, Fischer, Hashme, Hesse, J{\'o}zefowicz,
  Gray, Olsson, Pachocki, Petrov, de~Oliveira~Pinto, Raiman, Salimans,
  Schlatter, Schneider, Sidor, Sutskever, Tang, Wolski, and
  Zhang]{openai2019dota}
OpenAI, :, C.~Berner, G.~Brockman, B.~Chan, V.~Cheung, P.~Debiak, C.~Dennison,
  D.~Farhi, Q.~Fischer, S.~Hashme, C.~Hesse, R.~J{\'o}zefowicz, S.~Gray,
  C.~Olsson, J.~Pachocki, M.~Petrov, H.~P. de~Oliveira~Pinto, J.~Raiman,
  T.~Salimans, J.~Schlatter, J.~Schneider, S.~Sidor, I.~Sutskever, J.~Tang,
  F.~Wolski, and S.~Zhang, ``Dota 2 with large scale deep reinforcement
  learning,'' 2019.

\bibitem[Vinyals et~al.(2019)Vinyals, Babuschkin, Czarnecki, Mathieu, Dudzik,
  Chung, Choi, Powell, Ewalds, Georgiev, Oh, Horgan, Kroiss, Danihelka, Huang,
  Sifre, Cai, Agapiou, Jaderberg, Vezhnevets, Leblond, Pohlen, Dalibard,
  Budden, Sulsky, Molloy, Paine, Gulcehre, Wang, Pfaff, Wu, Ring, Yogatama,
  W{\"u}nsch, McKinney, Smith, Schaul, Lillicrap, Kavukcuoglu, Hassabis, Apps,
  and Silver]{Vinyals2019}
\BIBentryALTinterwordspacing
O.~Vinyals, I.~Babuschkin, W.~M. Czarnecki, M.~Mathieu, A.~Dudzik, J.~Chung,
  D.~H. Choi, R.~Powell, T.~Ewalds, P.~Georgiev, J.~Oh, D.~Horgan, M.~Kroiss,
  I.~Danihelka, A.~Huang, L.~Sifre, T.~Cai, J.~P. Agapiou, M.~Jaderberg, A.~S.
  Vezhnevets, R.~Leblond, T.~Pohlen, V.~Dalibard, D.~Budden, Y.~Sulsky,
  J.~Molloy, T.~L. Paine, C.~Gulcehre, Z.~Wang, T.~Pfaff, Y.~Wu, R.~Ring,
  D.~Yogatama, D.~W{\"u}nsch, K.~McKinney, O.~Smith, T.~Schaul, T.~Lillicrap,
  K.~Kavukcuoglu, D.~Hassabis, C.~Apps, and D.~Silver, ``Grandmaster level in
  starcraft ii using multi-agent reinforcement learning,'' \emph{Nature}, vol.
  575, no. 7782, pp. 350--354, 2019.
\BIBentrySTDinterwordspacing

\bibitem[OpenAI et~al.(2019{\natexlab{b}})OpenAI, Akkaya, Andrychowicz,
  Chociej, Litwin, McGrew, Petron, Paino, Plappert, Powell, Ribas, Schneider,
  Tezak, Tworek, Welinder, Weng, Yuan, Zaremba, and Zhang]{openai2019solving}
OpenAI, I.~Akkaya, M.~Andrychowicz, M.~Chociej, M.~Litwin, B.~McGrew,
  A.~Petron, A.~Paino, M.~Plappert, G.~Powell, R.~Ribas, J.~Schneider,
  N.~Tezak, J.~Tworek, P.~Welinder, L.~Weng, Q.~Yuan, W.~Zaremba, and L.~Zhang,
  ``Solving rubik's cube with a robot hand,'' 2019.

\bibitem[Badia et~al.(2020)Badia, Piot, Kapturowski, Sprechmann, Vitvitskyi,
  Guo, and Blundell]{badia2020agent57}
A.~P. Badia, B.~Piot, S.~Kapturowski, P.~Sprechmann, A.~Vitvitskyi, D.~Guo, and
  C.~Blundell, ``Agent57: Outperforming the atari human benchmark,'' 2020.

\bibitem[Pomerleau(1989)]{10.5555/89851.89891}
D.~A. Pomerleau, \emph{ALVINN: An Autonomous Land Vehicle in a Neural
  Network}.\hskip 1em plus 0.5em minus 0.4em\relax San Francisco, CA, USA:
  Morgan Kaufmann Publishers Inc., 1989, p. 305–313.

\bibitem[Ross et~al.(2011)Ross, Gordon, and Bagnell]{pmlr-v15-ross11a}
\BIBentryALTinterwordspacing
S.~Ross, G.~Gordon, and D.~Bagnell, ``A reduction of imitation learning and
  structured prediction to no-regret online learning,'' ser. Proceedings of
  Machine Learning Research, G.~Gordon, D.~Dunson, and M.~Dudík, Eds.,
  vol.~15.\hskip 1em plus 0.5em minus 0.4em\relax Fort Lauderdale, FL, USA:
  JMLR Workshop and Conference Proceedings, 11--13 Apr 2011, pp. 627--635.
\BIBentrySTDinterwordspacing

\bibitem[Scheller et~al.(2020)Scheller, Schraner, and
  Vogel]{pmlr-v123-scheller20a}
\BIBentryALTinterwordspacing
C.~Scheller, Y.~Schraner, and M.~Vogel, ``Sample efficient reinforcement
  learning through learning from demonstrations in minecraft,'' ser.
  Proceedings of Machine Learning Research, H.~J. Escalante and R.~Hadsell,
  Eds., vol. 123.\hskip 1em plus 0.5em minus 0.4em\relax Vancouver, CA: PMLR,
  08--14 Dec 2020, pp. 67--76.
\BIBentrySTDinterwordspacing

\bibitem[Sutton et~al.(1998)Sutton, Precup, and Singh]{10.5555/645527.657453}
R.~S. Sutton, D.~Precup, and S.~P. Singh, ``Intra-option learning about
  temporally abstract actions,'' in \emph{Proceedings of the Fifteenth
  International Conference on Machine Learning}, ser. ICML '98.\hskip 1em plus
  0.5em minus 0.4em\relax San Francisco, CA, USA: Morgan Kaufmann Publishers
  Inc., 1998, p. 556–564.

\bibitem[Dayan and Hinton(1993)]{NIPS1992_714}
\BIBentryALTinterwordspacing
P.~Dayan and G.~E. Hinton, ``Feudal reinforcement learning,'' in \emph{Advances
  in Neural Information Processing Systems 5}, S.~J. Hanson, J.~D. Cowan, and
  C.~L. Giles, Eds.\hskip 1em plus 0.5em minus 0.4em\relax Morgan-Kaufmann,
  1993, pp. 271--278.
\BIBentrySTDinterwordspacing

\bibitem[Sutton et~al.(1999{\natexlab{a}})Sutton, Precup, and
  Singh]{10.1016/S0004-3702(99)00052-1}
\BIBentryALTinterwordspacing
R.~S. Sutton, D.~Precup, and S.~Singh, ``Between mdps and semi-mdps: A
  framework for temporal abstraction in reinforcement learning,'' \emph{Artif.
  Intell.}, vol. 112, no. 1–2, p. 181–211, Aug. 1999.
\BIBentrySTDinterwordspacing

\bibitem[Elman(1993)]{Elman1993-ELMLAD}
J.~L. Elman, ``Learning and development in neural networks: The importance of
  starting small,'' \emph{Cognition}, vol.~48, no.~1, pp. 71--99, 1993.

\bibitem[Lazaric et~al.(2008)Lazaric, Restelli, and
  Bonarini]{10.1145/1390156.1390225}
\BIBentryALTinterwordspacing
A.~Lazaric, M.~Restelli, and A.~Bonarini, ``Transfer of samples in batch
  reinforcement learning,'' in \emph{Proceedings of the 25th International
  Conference on Machine Learning}, ser. ICML '08.\hskip 1em plus 0.5em minus
  0.4em\relax New York, NY, USA: Association for Computing Machinery, 2008, p.
  544–551.
\BIBentrySTDinterwordspacing

\bibitem[Lazaric and Restelli(2011)]{NIPS2011_fe7ee8fc}
\BIBentryALTinterwordspacing
A.~Lazaric and M.~Restelli, ``Transfer from multiple mdps,'' in \emph{Advances
  in Neural Information Processing Systems}, J.~Shawe-Taylor, R.~Zemel,
  P.~Bartlett, F.~Pereira, and K.~Q. Weinberger, Eds., vol.~24.\hskip 1em plus
  0.5em minus 0.4em\relax Curran Associates, Inc., 2011.
\BIBentrySTDinterwordspacing

\bibitem[Sutton et~al.(1999{\natexlab{b}})Sutton, Precup, and
  Singh]{Sutton99betweenmdps}
R.~Sutton, D.~Precup, and S.~Singh, ``Between mdps and semi-mdps: A framework
  for temporal abstraction in reinforcement learning,'' \emph{Artificial
  Intelligence}, vol. 112, pp. 181--211, 1999.

\bibitem[Soni and Singh(2006)]{DBLP:conf/aaai/SoniS06}
\BIBentryALTinterwordspacing
V.~Soni and S.~P. Singh, ``Using homomorphisms to transfer options across
  continuous reinforcement learning domains,'' in \emph{Proceedings, The
  Twenty-First National Conference on Artificial Intelligence and the
  Eighteenth Innovative Applications of Artificial Intelligence Conference,
  July 16-20, 2006, Boston, Massachusetts, {USA}}.\hskip 1em plus 0.5em minus
  0.4em\relax {AAAI} Press, 2006, pp. 494--499.
\BIBentrySTDinterwordspacing

\bibitem[Vezhnevets et~al.(2016)Vezhnevets, Mnih, Agapiou, Osindero, Graves,
  Vinyals, and Kavukcuoglu]{10.5555/3157382.3157488}
A.~S. Vezhnevets, V.~Mnih, J.~Agapiou, S.~Osindero, A.~Graves, O.~Vinyals, and
  K.~Kavukcuoglu, ``Strategic attentive writer for learning macro-actions,'' in
  \emph{Proceedings of the 30th International Conference on Neural Information
  Processing Systems}, ser. NIPS'16.\hskip 1em plus 0.5em minus 0.4em\relax Red
  Hook, NY, USA: Curran Associates Inc., 2016, p. 3494–3502.

\bibitem[Fachantidis et~al.(2013)Fachantidis, Partalas, Tsoumakas, and
  Vlahavas]{FACHANTIDIS201323}
\BIBentryALTinterwordspacing
A.~Fachantidis, I.~Partalas, G.~Tsoumakas, and I.~Vlahavas, ``Transferring task
  models in reinforcement learning agents,'' \emph{Neurocomputing}, vol. 107,
  pp. 23--32, 2013, timely Neural Networks Applications in Engineering.
\BIBentrySTDinterwordspacing

\bibitem[Fernández et~al.(2010)Fernández, García, and
  Veloso]{FERNANDEZ2010866}
\BIBentryALTinterwordspacing
F.~Fernández, J.~García, and M.~Veloso, ``Probabilistic policy reuse for
  inter-task transfer learning,'' \emph{Robotics and Autonomous Systems},
  vol.~58, no.~7, pp. 866--871, 2010, advances in Autonomous Robots for Service
  and Entertainment.
\BIBentrySTDinterwordspacing

\bibitem[Taylor et~al.(2007)Taylor, Stone, and Liu]{JMLR07-taylor}
M.~E. Taylor, P.~Stone, and Y.~Liu, ``Transfer learning via inter-task mappings
  for temporal difference learning,'' \emph{Journal of Machine Learning
  Research}, vol.~8, no.~1, pp. 2125--2167, 2007.

\bibitem[Taylor and Stone(2005)]{AAMAS05-transfer}
M.~E. Taylor and P.~Stone, ``Behavior transfer for value-function-based
  reinforcement learning,'' in \emph{The Fourth International Joint Conference
  on Autonomous Agents and Multiagent Systems}, F.~Dignum, V.~Dignum,
  S.~Koenig, S.~Kraus, M.~P. Singh, and M.~Wooldridge, Eds.\hskip 1em plus
  0.5em minus 0.4em\relax New York, NY: {ACM Press}, July 2005, pp. 53--59.

\bibitem[Taylor and Stone(2009)]{JMLR:v10:taylor09a}
\BIBentryALTinterwordspacing
------, ``Transfer learning for reinforcement learning domains: A survey,''
  \emph{Journal of Machine Learning Research}, vol.~10, no.~56, pp. 1633--1685,
  2009.
\BIBentrySTDinterwordspacing

\bibitem[Lazaric(2012)]{Lazaric2012}
\BIBentryALTinterwordspacing
A.~Lazaric, ``Transfer in reinforcement learning: A framework and a survey,''
  in \emph{Adaptation, Learning, and Optimization}.\hskip 1em plus 0.5em minus
  0.4em\relax Springer Berlin Heidelberg, 2012, pp. 143--173.
\BIBentrySTDinterwordspacing

\bibitem[Bengio et~al.(2009)Bengio, Louradour, Collobert, and
  Weston]{bengio:2009}
Y.~Bengio, J.~Louradour, R.~Collobert, and J.~Weston, ``Curriculum learning,''
  in \emph{International Conference on Machine Learning, {ICML}}, 2009.

\bibitem[Schaul et~al.(2016)Schaul, Quan, Antonoglou, and Silver]{Schaul2016}
T.~Schaul, J.~Quan, I.~Antonoglou, and D.~Silver, ``Prioritized experience
  replay,'' in \emph{International Conference on Learning Representations},
  Puerto Rico, 2016.

\bibitem[Andrychowicz et~al.(2017)Andrychowicz, Wolski, Ray, Schneider, Fong,
  Welinder, McGrew, Tobin, Pieter~Abbeel, and Zaremba]{NIPS2017_453fadbd}
\BIBentryALTinterwordspacing
M.~Andrychowicz, F.~Wolski, A.~Ray, J.~Schneider, R.~Fong, P.~Welinder,
  B.~McGrew, J.~Tobin, O.~Pieter~Abbeel, and W.~Zaremba, ``Hindsight experience
  replay,'' in \emph{Advances in Neural Information Processing Systems},
  I.~Guyon, U.~V. Luxburg, S.~Bengio, H.~Wallach, R.~Fergus, S.~Vishwanathan,
  and R.~Garnett, Eds., vol.~30.\hskip 1em plus 0.5em minus 0.4em\relax Curran
  Associates, Inc., 2017.
\BIBentrySTDinterwordspacing

\bibitem[Ren et~al.(2018)Ren, Dong, Li, and Chen]{8278851}
Z.~Ren, D.~Dong, H.~Li, and C.~Chen, ``Self-paced prioritized curriculum
  learning with coverage penalty in deep reinforcement learning,'' \emph{IEEE
  Transactions on Neural Networks and Learning Systems}, vol.~29, no.~6, pp.
  2216--2226, 2018.

\bibitem[Kim and Choi(2018)]{DBLP:journals/corr/abs-1801-00904}
\BIBentryALTinterwordspacing
T.~Kim and J.~Choi, ``Screenernet: Learning curriculum for neural networks,''
  \emph{CoRR}, vol. abs/1801.00904, 2018.
\BIBentrySTDinterwordspacing

\bibitem[Florensa et~al.(2017)Florensa, Held, Wulfmeier, Zhang, and
  Abbeel]{pmlr-v78-florensa17a}
\BIBentryALTinterwordspacing
C.~Florensa, D.~Held, M.~Wulfmeier, M.~Zhang, and P.~Abbeel, ``Reverse
  curriculum generation for reinforcement learning,'' in \emph{Proceedings of
  the 1st Annual Conference on Robot Learning}, ser. Proceedings of Machine
  Learning Research, S.~Levine, V.~Vanhoucke, and K.~Goldberg, Eds.,
  vol.~78.\hskip 1em plus 0.5em minus 0.4em\relax PMLR, 13--15 Nov 2017, pp.
  482--495.
\BIBentrySTDinterwordspacing

\bibitem[Foglino et~al.(2019{\natexlab{a}})Foglino, Christakou, and
  Leonetti]{8850690}
F.~Foglino, C.~C. Christakou, and M.~Leonetti, ``An optimization framework for
  task sequencing in curriculum learning,'' in \emph{2019 Joint IEEE 9th
  International Conference on Development and Learning and Epigenetic Robotics
  (ICDL-EpiRob)}, 2019, pp. 207--214.

\bibitem[Foglino et~al.(2019{\natexlab{b}})Foglino, Coletto~Christakou,
  Luna~Gutierrez, and Leonetti]{ijcai2019-320}
\BIBentryALTinterwordspacing
F.~Foglino, C.~Coletto~Christakou, R.~Luna~Gutierrez, and M.~Leonetti,
  ``Curriculum learning for cumulative return maximization,'' in
  \emph{Proceedings of the Twenty-Eighth International Joint Conference on
  Artificial Intelligence, {IJCAI-19}}.\hskip 1em plus 0.5em minus 0.4em\relax
  International Joint Conferences on Artificial Intelligence Organization, 7
  2019, pp. 2308--2314.
\BIBentrySTDinterwordspacing

\bibitem[Narvekar et~al.(2017)Narvekar, Sinapov, and Stone]{ijcai2017-353}
\BIBentryALTinterwordspacing
S.~Narvekar, J.~Sinapov, and P.~Stone, ``Autonomous task sequencing for
  customized curriculum design in reinforcement learning,'' in
  \emph{Proceedings of the Twenty-Sixth International Joint Conference on
  Artificial Intelligence, {IJCAI-17}}, 2017, pp. 2536--2542.
\BIBentrySTDinterwordspacing

\bibitem[{Matiisen} et~al.(2019){Matiisen}, {Oliver}, {Cohen}, and
  {Schulman}]{8827566}
T.~{Matiisen}, A.~{Oliver}, T.~{Cohen}, and J.~{Schulman}, ``Teacher-student
  curriculum learning,'' \emph{IEEE Transactions on Neural Networks and
  Learning Systems}, pp. 1--9, 2019.

\bibitem[Narvekar and Stone(2019)]{AAMAS19-Narvekar}
S.~Narvekar and P.~Stone, ``Learning curriculum policies for reinforcement
  learning,'' in \emph{Proceedings of the 18th International Conference on
  Autonomous Agents and Multiagent Systems (AAMAS)}, May 2019.

\bibitem[Narvekar et~al.(2016)Narvekar, Sinapov, Leonetti, and
  Stone]{AAMAS16-Narvekar}
\BIBentryALTinterwordspacing
S.~Narvekar, J.~Sinapov, M.~Leonetti, and P.~Stone, ``Source task creation for
  curriculum learning,'' in \emph{Proceedings of the 15th International
  Conference on Autonomous Agents and Multiagent Systems (AAMAS 2016)},
  Singapore, May 2016.
\BIBentrySTDinterwordspacing

\bibitem[Schmidhuber(2013)]{10.3389/fpsyg.2013.00313}
\BIBentryALTinterwordspacing
J.~Schmidhuber, ``Powerplay: Training an increasingly general problem solver by
  continually searching for the simplest still unsolvable problem,''
  \emph{Frontiers in Psychology}, vol.~4, p. 313, 2013.
\BIBentrySTDinterwordspacing

\bibitem[Jiang et~al.(2020)Jiang, Grefenstette, and
  Rocktäschel]{jiang2020prioritized}
M.~Jiang, E.~Grefenstette, and T.~Rocktäschel, ``Prioritized level replay,''
  2020.

\bibitem[Kurach et~al.(2019)Kurach, Raichuk, Stanczyk, Zajac, Bachem, Espeholt,
  Riquelme, Vincent, Michalski, Bousquet, and Gelly]{kurach2019google}
K.~Kurach, A.~Raichuk, P.~Stanczyk, M.~Zajac, O.~Bachem, L.~Espeholt,
  C.~Riquelme, D.~Vincent, M.~Michalski, O.~Bousquet, and S.~Gelly, ``Google
  research football: A novel reinforcement learning environment,'' 2019.

\bibitem[Flet-Berliac et~al.(2020)Flet-Berliac, Ouhamma, Maillard, and
  Preux]{fletberliac2020standard}
Y.~Flet-Berliac, R.~Ouhamma, O.-A. Maillard, and P.~Preux, ``Is standard
  deviation the new standard? revisiting the critic in deep policy gradients,''
  2020.

\bibitem[Schraner(2020)]{schraner_2020}
Y.~Schraner, \emph{ReinforcementLearning on the GoogleFootball environment},
  2020.

\bibitem[Sutton and Barto(2018)]{Sutton1998}
\BIBentryALTinterwordspacing
R.~S. Sutton and A.~G. Barto, \emph{Reinforcement Learning: An Introduction},
  2nd~ed.\hskip 1em plus 0.5em minus 0.4em\relax The MIT Press, 2018.
\BIBentrySTDinterwordspacing

\bibitem[Schulman et~al.(2017)Schulman, Wolski, Dhariwal, Radford, and
  Klimov]{DBLP:journals/corr/SchulmanWDRK17}
\BIBentryALTinterwordspacing
J.~Schulman, F.~Wolski, P.~Dhariwal, A.~Radford, and O.~Klimov, ``Proximal
  policy optimization algorithms,'' \emph{CoRR}, vol. abs/1707.06347, 2017.
\BIBentrySTDinterwordspacing

\bibitem[Schulman et~al.(2015)Schulman, Levine, Abbeel, Jordan, and
  Moritz]{pmlr-v37-schulman15}
\BIBentryALTinterwordspacing
J.~Schulman, S.~Levine, P.~Abbeel, M.~Jordan, and P.~Moritz, ``Trust region
  policy optimization,'' ser. Proceedings of Machine Learning Research, F.~Bach
  and D.~Blei, Eds., vol.~37.\hskip 1em plus 0.5em minus 0.4em\relax Lille,
  France: PMLR, 07--09 Jul 2015, pp. 1889--1897.
\BIBentrySTDinterwordspacing

\bibitem[Hsu et~al.(2020)Hsu, Mendler{-}D{\"{u}}nner, and
  Hardt]{DBLP:journals/corr/abs-2009-10897}
\BIBentryALTinterwordspacing
C.~C. Hsu, C.~Mendler{-}D{\"{u}}nner, and M.~Hardt, ``Revisiting design choices
  in proximal policy optimization,'' \emph{CoRR}, vol. abs/2009.10897, 2020.
\BIBentrySTDinterwordspacing

\bibitem[Narvekar et~al.(2020)Narvekar, Peng, Leonetti, Sinapov, Taylor, and
  Stone]{DBLP:journals/jmlr/NarvekarPLSTS20}
\BIBentryALTinterwordspacing
S.~Narvekar, B.~Peng, M.~Leonetti, J.~Sinapov, M.~E. Taylor, and P.~Stone,
  ``Curriculum learning for reinforcement learning domains: {A} framework and
  survey,'' \emph{J. Mach. Learn. Res.}, vol.~21, pp. 181:1--181:50, 2020.
\BIBentrySTDinterwordspacing

\bibitem[Svetlik et~al.(2017)Svetlik, Leonetti, Sinapov, Shah, Walker, and
  Stone]{Svetlik_Leonetti_Sinapov_Shah_Walker_Stone_2017}
\BIBentryALTinterwordspacing
M.~Svetlik, M.~Leonetti, J.~Sinapov, R.~Shah, N.~Walker, and P.~Stone,
  ``Automatic curriculum graph generation for reinforcement learning agents,''
  \emph{Proceedings of the AAAI Conference on Artificial Intelligence},
  vol.~31, no.~1, Feb. 2017.
\BIBentrySTDinterwordspacing

\bibitem[Wang et~al.(2019)Wang, Lehman, Clune, and
  Stanley]{10.1145/3321707.3321799}
\BIBentryALTinterwordspacing
R.~Wang, J.~Lehman, J.~Clune, and K.~O. Stanley, ``Poet: Open-ended coevolution
  of environments and their optimized solutions,'' in \emph{Proceedings of the
  Genetic and Evolutionary Computation Conference}, ser. GECCO '19.\hskip 1em
  plus 0.5em minus 0.4em\relax New York, NY, USA: Association for Computing
  Machinery, 2019, p. 142–151.
\BIBentrySTDinterwordspacing

\bibitem[F.R.S.(1901)]{doi:10.1080/14786440109462720}
K.~P. F.R.S., ``Liii. on lines and planes of closest fit to systems of points
  in space,'' \emph{The London, Edinburgh, and Dublin Philosophical Magazine
  and Journal of Science}, vol.~2, no.~11, pp. 559--572, 1901.

\bibitem[Portelas et~al.(2019)Portelas, Colas, Hofmann, and
  Oudeyer]{DBLP:conf/corl/PortelasCHO19}
\BIBentryALTinterwordspacing
R.~Portelas, C.~Colas, K.~Hofmann, and P.~Oudeyer, ``Teacher algorithms for
  curriculum learning of deep {RL} in continuously parameterized
  environments,'' in \emph{3rd Annual Conference on Robot Learning, CoRL 2019,
  Osaka, Japan, October 30 - November 1, 2019, Proceedings}, ser. Proceedings
  of Machine Learning Research, L.~P. Kaelbling, D.~Kragic, and K.~Sugiura,
  Eds., vol. 100.\hskip 1em plus 0.5em minus 0.4em\relax {PMLR}, 2019, pp.
  835--853.
\BIBentrySTDinterwordspacing

\bibitem[Kanitscheider et~al.(2021)Kanitscheider, Huizinga, Farhi, Guss,
  Houghton, Sampedro, Zhokhov, Baker, Ecoffet, Tang, Klimov, and
  Clune]{DBLP:journals/corr/abs-2106-14876}
\BIBentryALTinterwordspacing
I.~Kanitscheider, J.~Huizinga, D.~Farhi, W.~H. Guss, B.~Houghton, R.~Sampedro,
  P.~Zhokhov, B.~Baker, A.~Ecoffet, J.~Tang, O.~Klimov, and J.~Clune,
  ``Multi-task curriculum learning in a complex, visual, hard-exploration
  domain: Minecraft,'' \emph{CoRR}, vol. abs/2106.14876, 2021.
\BIBentrySTDinterwordspacing

\bibitem[Agarap(2018)]{agarap2018learning}
\BIBentryALTinterwordspacing
A.~F. Agarap, ``Deep learning using rectified linear units (relu),'' 2018, cite
  arxiv:1803.08375Comment: 7 pages, 11 figures, 9 tables.
\BIBentrySTDinterwordspacing

\bibitem[Hochreiter and Schmidhuber(1997)]{hochreiter1997long}
S.~Hochreiter and J.~Schmidhuber, ``Long short-term memory,'' \emph{Neural
  computation}, vol.~9, no.~8, pp. 1735--1780, 1997.

\bibitem[Chevalier-Boisvert et~al.(2018)Chevalier-Boisvert, Willems, and
  Pal]{gym_minigrid}
M.~Chevalier-Boisvert, L.~Willems, and S.~Pal, ``Minimalistic gridworld
  environment for openai gym,'' \url{https://github.com/maximecb/gym-minigrid},
  2018.

\bibitem[Schraner(2019)]{schraner_2019}
Y.~Schraner, \emph{IP7 \!- Reinforcement Learning on the Google Football
  environment}, 2019.

\bibitem[Racani{\`{e}}re et~al.(2020)Racani{\`{e}}re, Lampinen, Santoro,
  Reichert, Firoiu, and Lillicrap]{DBLP:conf/iclr/RacaniereLSRFL20}
\BIBentryALTinterwordspacing
S.~Racani{\`{e}}re, A.~K. Lampinen, A.~Santoro, D.~P. Reichert, V.~Firoiu, and
  T.~P. Lillicrap, ``Automated curriculum generation through setter-solver
  interactions,'' in \emph{8th International Conference on Learning
  Representations, {ICLR} 2020, Addis Ababa, Ethiopia, April 26-30,
  2020}.\hskip 1em plus 0.5em minus 0.4em\relax OpenReview.net, 2020.
\BIBentrySTDinterwordspacing

\bibitem[Riedmiller et~al.(2018)Riedmiller, Hafner, Lampe, Neunert, Degrave,
  van~de Wiele, Mnih, Heess, and Springenberg]{pmlr-v80-riedmiller18a}
\BIBentryALTinterwordspacing
M.~Riedmiller, R.~Hafner, T.~Lampe, M.~Neunert, J.~Degrave, T.~van~de Wiele,
  V.~Mnih, N.~Heess, and J.~T. Springenberg, ``Learning by playing solving
  sparse reward tasks from scratch,'' in \emph{Proceedings of the 35th
  International Conference on Machine Learning}, ser. Proceedings of Machine
  Learning Research, J.~Dy and A.~Krause, Eds., vol.~80.\hskip 1em plus 0.5em
  minus 0.4em\relax PMLR, 10--15 Jul 2018, pp. 4344--4353.
\BIBentrySTDinterwordspacing

\bibitem[Shao et~al.(2019)Shao, Zhu, and Zhao]{8351991}
K.~Shao, Y.~Zhu, and D.~Zhao, ``Starcraft micromanagement with reinforcement
  learning and curriculum transfer learning,'' \emph{IEEE Transactions on
  Emerging Topics in Computational Intelligence}, vol.~3, no.~1, pp. 73--84,
  2019.

\end{thebibliography}
\addcontentsline{toc}{chapter}{Bibliography}
\clearpage

\listoffigures
\clearpage

\listoftables
\clearpage

\appendix
\chapter{Appendix}
\label{sec:appendix}

\section{Grid World CNN Architecture}
\label{sec:appendixCNN}
\begin{figure}[htb]
    \centering
    \includegraphics[width=0.25\textwidth]{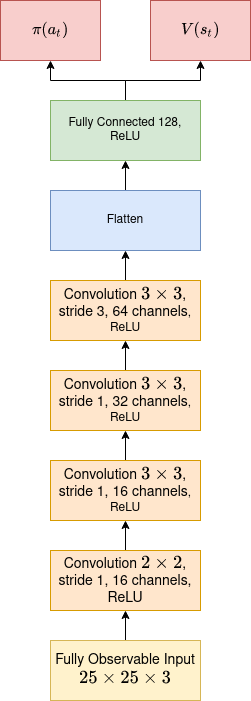}
    \caption{Student CNN neural network architecture for the MiniGrid environments.}
    \label{fig:CNNGridworldStudent}
\end{figure}

In \cref{fig:CNNGridworldStudent} we depict the CNN network architecture for grid world environments.
The network input is a $25\times25\times3$ fully observable representation of the environment's state.
We use four convolution blocks with $[16, 16, 32, 64]$ channels, a $[2\times2, 3\times3, 3\times3, 3\times3]$ kernel and a stride of $[1,1,1,3]$.
A ReLU activation function follows each convolution block.
After the convolution blocks, we use a fully connected layer with 128 units.
The policy and value function are separate heads on top of the fully connected layer.
The policy head uses a softmax activation function.

\clearpage

\section{Grid World Student Hyperparameters}
\label{sec:appendixGridworldStudent}
We optimized a limited number of hyperparameters for our students in the grid world environment.
The hyperparameters were tuned by training an rl agent with PPO on a single environment for 10 million steps.
We evaluate all hyperparameters on all environments in $\mathcal{V}$.
The parameters are reported in \cref{tab:studentGWPPO}.
Parameters with only a single "range" value in \cref{tab:studentGWPPO} were set to common values used for PPO.

\begin{table}[htb]
    \centering
    \begin{tabular}{ccc}
        \hline
        \textbf{Parameter} & \textbf{Range} & \textbf{Best} \\
        \hline
        \hline
        Learning rate & [0.01, 0.03, 0.001, 0.003, 0.0001, 0.0003] & 0.0003 \\
        Discount & [0.97, 0.99, 0.997, 0.999] & 0.999 \\
        Entropy loss coefficient & 0.01 & 0.01 \\
        Value loss coefficient & 0.5 & 0.5 \\
        Gradient norm clip & 0.5 & 0.5 \\
        GAE $\lambda$ & 0.95 & 0.95 \\
        Clipping range & 0.2 & 0.2 \\
        Normalize advantage & True & True \\
        Minibatches & 1 & 1 \\
        Epochs & [1,2,4] & 2 \\
        Optimizer & adam & adam \\
        Training steps & $10'000$ & $10'000$ \\
        Unroll length & [256, 512] & 256 \\
        Number of actors & 40 & 40
    \end{tabular}
    \caption{Hyperparameters used for training a student in the CMDP setting on the grid world environments.}
    \label{tab:studentGWPPO}
\end{table}

\clearpage

\section{Google Football Student Hyperparameters}
\label{sec:appendixGFootballStudentPPO}
In \cref{tab:FixedHyperparameters} we report the hyperparameters used in our Google Football CMDP experiments.
These values were tuned in our previous work \citep{schraner_2020}. Therefore we do not tune any of those parameters in this thesis.

\begin{table}[htb]
    \centering
    \begin{tabular}{cc}
        \hline
         \textbf{Parameter} & \textbf{Value} \\
         \hline
         \hline
         Learning rate & 0.0011879 \\
         Discount & $0.997$ \\
         Entropy loss coefficient & $0.00155$ \\
         Gradient norm clip & $0.76$ \\
         Value loss coefficient & $0.5$ \\
         GAE $\lambda$ & $0.95$ \\
         Clipping range & 0.115 \\
         Normalize advantage & True \\
         Minibatches & 4 \\
         Epochs & 2 \\
         Optimizer & adam \\
         Reward & Scoring \\
         Observation & SMM \\
         Training steps & $10'000$ \\
         Unroll length & 512 \\
         Number of actors & 40 \\
    \end{tabular}
    \caption{Important fixed student hyperparameters used in our Google Football CMDP experiments.}
    \label{tab:FixedHyperparameters}
\end{table}

\clearpage

\section{Teacher Hyperparameters}
\label{sec:appendixTeacherPPO}
We optimized a limited number of hyperparameters for our teacher on the grid world environment.
The hyperparameters were tuned by training a teacher agent for 1000 steps with PPO, using policy transfer, \textit{total average return} and the RH observation.
The parameters are reported in \cref{tab:teacherPPO} and used for both the grid world and Google Football experiments.
Parameters with only a single "range" value in \cref{tab:teacherPPO} were set to common values used for PPO.

\begin{table}[htb]
    \centering
    \begin{tabular}{ccc}
        \hline
        \textbf{Parameter} & \textbf{Range} & \textbf{Best} \\
        \hline
        \hline
        Learning rate & [0.1, 0.3, 0.01, 0.03] & 0.03 \\
        Linear learning rate schedule & True & True \\
        Discount & 1 & 1 \\
        Entropy loss coefficient & 0.01 & 0.01 \\
        Value loss coefficient & 0.5 & 0.5 \\
        Gradient norm clip & 0.5 & 0.5 \\
        GAE $\lambda$ & 0.95 & 0.95 \\
        Clipping range & 0.2 & 0.2 \\
        Normalize advantage & True & True \\
        Minibatches & 1 & 1 \\
        Epochs & 4 & 4 \\
        Optimizer & adam & adam \\
        Training steps & 1000 & 1000 \\
        Unroll length & 4 & 4 \\
    \end{tabular}
    \caption{Hyperparameters used for training a teacher in the CMDP setting used in both the grid world and Google Football experiments.}
    \label{tab:teacherPPO}
\end{table}

\clearpage

\section{MLP vs. LSTM Teacher Model}
\label{sec:appendixMLPvsLSTM}
We evaluated teachers with an MLP network architecture and an LSTM architecture as described in \cref{sec:teacherArchitecture}.
Executing experiments with the LSTM architecture takes around 1.5 times as long as using the MLP architecture.
In \cref{tab:teacherLSTMvsMLP} we compare both network architectures.
We report the total average return. Each teacher is trained for 1'000 CMDP steps, where for each step, the student is trained for 10'000 frames in the selected environment.
The \textit{total average return} reward signal and policy transfer are used in all experiments.
We do not report results for LP and EMA for the LSTM model, as those observation types were developed after this evaluation has been carried out.
The MLP architecture is superior to the LSTM architecture for all experiments except the RH observation type.
Therefore, we use the MLP architecture for the experiments reported in this thesis.

\begin{table}[htb]
    \centering
    \begin{tabular}{ccccccc}
        \hline
        \textbf{Model} & \textbf{RH} & \textbf{PTR} & \textbf{LP} & \textbf{ALP} & \textbf{EMA} & \textbf{FS-EMA} \\
        \hline
        \hline
        MLP  & 2.34 & \textbf{4.44} & \textbf{4.17} & \textbf{3.35} & \textbf{4.35} & \textbf{3.72} \\
        LSTM & \textbf{3.12} & 2.97 & -    & 2.95 & -    & 2.76
    \end{tabular}
    \caption{Comparison of the MLP and LSTM teacher architecture.
    We report the total average return over 100 episodes on all environments in $\mathcal{V}$ at the end of training.}
    \label{tab:teacherLSTMvsMLP}
\end{table}
\clearpage

\end{document}